\documentclass{article}

\usepackage[table]{xcolor}        
\usepackage[final,nonatbib]{neurips_2025}
\usepackage{mathrsfs}
\usepackage[most]{tcolorbox}
\usepackage{enumitem}
\usepackage{array}
\usepackage[utf8]{inputenc} 
\usepackage[T1]{fontenc}    
\usepackage{hyperref}       
\usepackage{url}      
\usepackage{xurl}           
\usepackage{booktabs}       
\usepackage{amsfonts}       
\usepackage{nicefrac}       
\usepackage{microtype}      
\usepackage{subfigure}
\usepackage{makecell}
\usepackage{amsmath}
\usepackage{algpseudocode}
\usepackage[ruled,vlined]{algorithm2e}
\usepackage{threeparttable}

\usepackage{amssymb}
\usepackage{pifont}
\usepackage{graphicx}
\usepackage{multirow}
\usepackage{caption}
\usepackage{float}
\usepackage{setspace}

\title{OneIG-Bench: Omni-dimensional Nuanced Evaluation for Image Generation}

\author{
  Jingjing Chang$^{1,2}$ 
  \quad
  Yixiao Fang$^{2,\dag}$
  \quad
  Peng Xing$^{2}$ 
  \quad
  Shuhan Wu$^{2}$ 
  \quad
  Wei Cheng$^{2}$ 
  \\
  \textbf{Rui Wang}$^{2}$ 
  \quad
  \textbf{Xianfang Zeng}$^{2}$ 
  \quad
  \textbf{Gang Yu}$^{2,\ddag}$
  \quad
  \textbf{Hai-Bao Chen}$^{1,\ddag}$\\
  \\
  $^1$~SJTU\quad$^2$~StepFun
  \\
  $^{\dag}$~Project lead\quad$^{\ddag}$~Corresponding author\\
  \\
  \textcolor{cityblue}{
    \raisebox{-0.2\height}{\includegraphics[height=0.5cm]{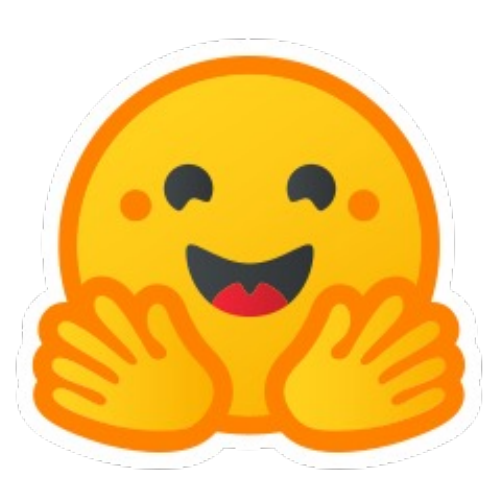}}~{\href{https://huggingface.co/datasets/OneIG-Bench/OneIG-Bench}{\textbf{Dataset}}}
    \quad
    \raisebox{-0.2\height}{\includegraphics[height=0.5cm]{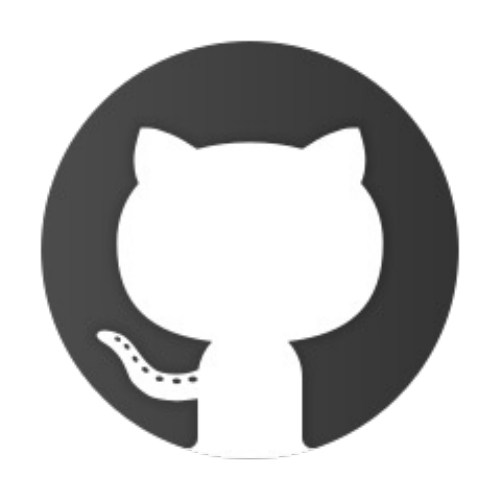}}~{\href{https://github.com/OneIG-Bench/OneIG-Benchmark}{\textbf{Code}}
    \quad
    \raisebox{-0.2\height}{\includegraphics[height=0.5cm]{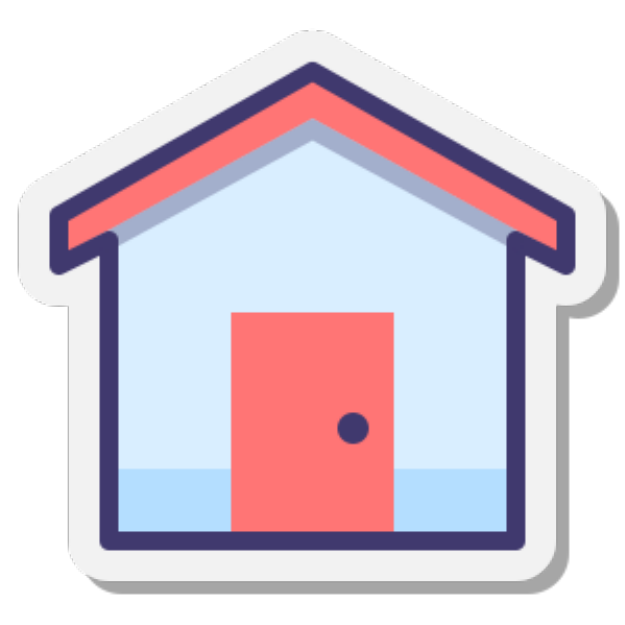}}}~{\href{https://oneig-bench.github.io}{\textbf{Project Page}}}}
}

\begin{document}


\definecolor{colorfirst}{RGB}{252,141,89}
\definecolor{colorsecond}{RGB}{253,187,132}
\definecolor{colorthird}{RGB}{253,212,158}
\definecolor{colorfourth}{RGB}{254,232,200}
\definecolor{colorfifth}{RGB}{255,247,236}
\definecolor{mygreen}{RGB}{112, 180, 143}
\definecolor{myred}{RGB}{242, 128, 128}
\definecolor{citypink}{RGB}{227, 108, 194}
\definecolor{cityblue}{RGB}{128, 159, 225}
\newcommand{\rankfirst}[0]{\cellcolor{colorfirst}}
\newcommand{\ranksecond}[0]{\cellcolor{colorsecond}}
\newcommand{\rankthird}[0]{\cellcolor{colorthird}}
\newcommand{\rankfourth}[0]{\cellcolor{colorfourth}}
\newcommand{\rankfifth}[0]{\cellcolor{colorfifth}}
\DeclareRobustCommand{\legendsquare}[1]{%
  \textcolor{#1}{\rule{2ex}{2ex}}%
}
\DeclareRobustCommand{\legendsquarebox}[1]{%
  \tikz[] \draw[black, fill=#1, line width=0.4pt] (0,0) rectangle (1.5ex,1.5ex);%
}
\newcommand{\cmark}{\textcolor{mygreen}{\ding{51}}}%
\newcommand{\xmark}{\textcolor{myred}{\ding{55}}}%


\maketitle

\begin{abstract}

Text-to-image (T2I) models have garnered significant attention for generating high-quality images aligned with text prompts. 
However, rapid T2I model advancements reveal limitations in early benchmarks, lacking comprehensive evaluations, for example, the evaluation on reasoning,  text rendering and style. 
Notably, recent state-of-the-art models, with their rich knowledge modeling capabilities, show promising results on the image generation problems requiring strong reasoning ability, yet existing evaluation systems have not adequately addressed this frontier.
To systematically address these gaps, we introduce \textbf{OneIG-Bench}, a meticulously designed comprehensive benchmark framework for fine-grained evaluation of T2I models across multiple dimensions, including prompt-image  alignment, text rendering precision, reasoning-generated content, stylization, and diversity. 
By structuring the evaluation, this benchmark enables in-depth analysis of model performance, helping researchers and practitioners pinpoint strengths and bottlenecks in the full pipeline of image generation. Specifically, \textbf{OneIG-Bench} enables flexible evaluation by allowing users to focus on a particular evaluation subset. Instead of generating images for the entire set of prompts, users can generate images only for the prompts associated with the selected dimension and complete the corresponding evaluation accordingly.
Our codebase and dataset are now publicly available to facilitate reproducible evaluation studies and cross-model comparisons within the T2I research community.

\end{abstract}

\section{Introduction}\label{sec:intro}
Recent years have witnessed remarkable advancements in text-to-image (T2I) models across image quality, semantic alignment, text rendering precision, and knowledge-driven reasoning in image generation~\cite{Rombach_2022_CVPR,podell2023sdxl,esser2024scaling,2024sd3.5,flux2024,ramesh2022hierarchical,openai2024gpt4o_image}. However, the development of evaluation systems has significantly lagged behind model progress: most existing benchmarks remain confined to single-dimensional assessments, lacking comprehensiveness. For instance, T2ICompBench~\cite{huang2023t2icompbench}, GenEval~\cite{ghosh2023geneval}, and DSG-1k~\cite{Cho2024DSG} focus on short-text semantic understanding, while DPG-Bench~\cite{hu2024ella} introduces dense prompt evaluation but lackly coverage of limited dimensions like style and text. Although WorldGenBench~\cite{zhang2025worldgenbench} addresses world knowledge and reasoning, a more comprehensive multi-dimensional evaluation framework is urgently needed to provide scientific model assessments and guide technological development.

\begin{figure}[ht]
    \centering
    \setlength{\abovecaptionskip}{1pt}
    \includegraphics[width=.92\linewidth]{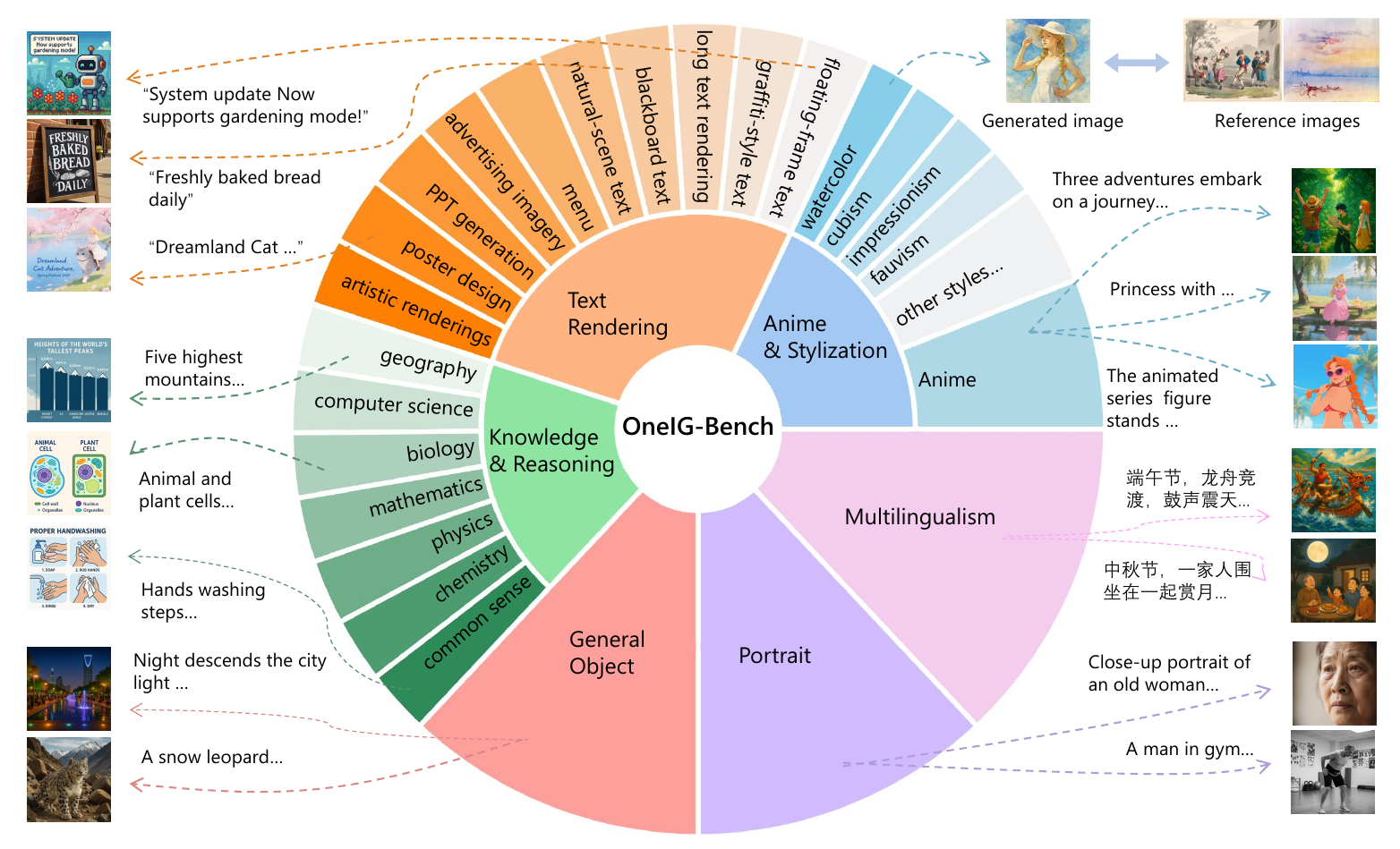}
    \caption{\small \textbf{Overview of OneIG-Bench}. OneIG-Bench comprises six core categories, each designed to evaluate targeted capabilities across distinct generative dimensions, with approximately 200 carefully curated prompts per category to ensure comprehensive coverage of diverse scenarios.}
    \label{fig:teaser}
    \vspace{-0.4cm}
\end{figure}

To drive the advancement of text-to-image models, we posit that the development of a holistic benchmark framework capable of evaluating models across multiple interconnected dimensions is imperative for fostering rigorous, comprehensive assessment. We introduce \textbf{OneIG-Bench} shown in Figure~\ref{fig:teaser}, comprising over 1000 omni-dimensional prompts, primarily sourced from real-world user inputs, systematically designed to comprehensively evaluate text-to-image models across diverse generative capabilities. Based on distinct generative themes, we classify the evaluation dataset into six core assessment categories: \textit{General Object}, \textit{Portrait}, \textit{Anime and Stylization}, \textit{Text Rendering}, \textit{Knowledge and Reasoning} and \textit{Multilingualism}. By leveraging our framework for the taxonomic differentiation of generative themes, we enable a more nuanced evaluation of models' multidimensional capabilities. Consequently, this systematic assessment allows users to identify models with tailored capabilities that align precisely with their specific application requirements.

For different evaluation dimensions, we have carefully devised quantitative indicators, taking into account various factors to ensure the comprehensiveness and objectivity of the evaluation. These indicators are designed to precisely measure the model's performance in different aspects related to specific subject domains, enabling a more accurate and in-depth assessment. Specifically for the evaluation of \textit{General object}, \textit{Portrait}, \textit{Anime and Stylization}, we have developed an evaluation system for the ability to comply with input prompts. Regarding \textit{Anime and Stylization}, we also incorporate the stylistic similarity into our evaluation system, specifically designed to assess models' capabilities in reproducing diverse artistic styles. In the \textit{Text Rendering} evaluation segment, we focus on three core metrics: the edit distance between generated text and ground truth, the text completion rate in one visual output, and the overall text generation accuracy. For the \textit{Knowledge and Reasoning} part, our assessment centers on whether the model possesses the required domain knowledge and can accurately interpret user intent, thereby enabling the generation of semantically coherent and logically consistent images. For the \textit{Multilingualism} part, we evaluate the alignment between the images generated from cultural element prompts and the corresponding cultural elements. Additionally, we also evaluate the diversity across all six dimensions. 

We summarize our key contributions in the following three points:
\label{check:contributions2}
\begin{itemize}
\vspace{-0.15cm}
 \item We present \textbf{OneIG-Bench}, which consists of six prompt sets, with the first five — 245 \textit{Anime and Stylization}, 244 \textit{Portrait}, 206 \textit{General Object}, 200 \textit{Text Rendering}, and 225 \textit{Knowledge and Reasoning} prompts — each provided in both English and Chinese, and 200 \textit{Multilingualism} prompts, designed for the comprehensive evaluation of current text-to-image models.
 
\vspace{-0.15cm}
\item A systematic quantitative evaluation is developed to facilitate objective capability ranking through standardized metrics, enabling direct comparability across models. Specifically, our evaluation framework allows T2I models to generate images only for prompts associated with a particular evaluation dimension, and to assess performance accordingly within that dimension.

\vspace{-0.15cm}
\item State-of-the-art open-sourced methods as well as the proprietary model are evaluated based on our proposed benchmark to facilitate the development of text-to-image research.

\end{itemize}

\section{Related Works}\label{sec:related}
\subsection{Text to Image Models}

Text-to-image (T2I) generation aims to develop models that produce images semantically consistent with given text descriptions. 
Early explorations during the generative adversarial network (GAN)~\cite{goodfellow2020generative,sauer2023stylegan} era laid foundational work, but these methods suffered significant limitations due to mode collapse, often failing to generate even simple subject matters accurately.
In recent years, generative approaches based on the diffusion model paradigm have emerged as a dominant trend~\cite{ho2020denoising,song2020denoising}. Notable advancements include Unet-based architectures like Stable Diffusion XL~\cite{podell2023sdxl}, Diffusion Image Transformer (DiT)-based models~\cite{chen2024pixart,chen2023pixart,ma2024sit}, double-stream MMDiT frameworks~\cite{2024sd3.5,esser2024scaling}, and hybrid designs combining double-stream and single-stream networks~\cite{2025cogview4, xie2025sana,flux2024}. These models have achieved remarkable progress in generating high-quality images that closely align with textual semantics.
With technological advancements, the evaluation framework for T2I models must evolve from single-dimensional attribute assessments (e.g., color, shape) to multi-layered evaluations, encompassing semantic alignment, stylistic consistency, and text rendering accuracy. 
Meanwhile, autoregressive-based models~\cite{chen2020generative,radford2019language,ramesh2021zero,sun2024autoregressive,yu2022scaling,team2024chameleon,wang2024emu3,xie2024show} have demonstrated unique strengths in knowledge modeling and complex text comprehension, necessitating the integration of systematic reasoning capability evaluations into the assessment framework. In summary, the rapid development of T2I generation underscores the urgent need for a comprehensive and rigorous evaluation system to accurately measure model performance, identify strengths and weaknesses, and foster sustainable progress in the field.

\subsection{Text to Image Evaluation}

In the early stages of text-to-image development, researchers typically employed some metrics to evaluate image generation quality, such as FID~\cite{heusel2017gans}(Fréchet Inception Distance), SSIM~\cite{wang2004image}(Structural Similarity Index Measure), PSNR(Peak Signal to Noise Ratio), etc. However, these methods do not provide a comprehensive understanding of the model's capabilities and fail to capture the model's ability to comprehend higher-order semantics. 

In recent years, the technology of text-to-image models has been evolving rapidly, and the corresponding evaluation system urgently needs to be innovated and upgraded. Taking the evaluation of Stable Diffusion 1.5 ~\cite{Rombach_2022_CVPR} and Stable Diffusion XL ~\cite{podell2023sdxl}  as examples, most of the existing benchmark evaluations  (Attend~\cite{chefer2023attend}, Hrs-bench~\cite{bakr2023hrs}, CC500~\cite{feng2022training}) focus on judging the degree of restoration of the core elements in the prompts. 
In the face of the rapid evolution of model technology, it has become difficult to comprehensively and accurately measure the actual performance and innovative potential of the models.
Subsequently, evaluation methods such as PartiPrompt~\cite{yu2022scaling}, DrawBench~\cite{saharia2022photorealistic}, TIFA~\cite{hu2023tifa}, Gecko~\cite{wiles2024revisiting}, EvalAlign~\cite{tan2024evalalign}, T2ICompBench~\cite{huang2023t2icompbench}, T2ICompBench++~\cite{huang2025t2i}, GenEval~\cite{ghosh2023geneval}, EvalMuse~\cite{han2024evalmuse40kreliablefinegrainedbenchmark}, DPG-Bench~\cite{hu2024ella}, and GenAI-Bench~\cite{li2024genai} introduced Visual Language Models (CLIP~\cite{radford2021learning},BLIP~\cite{li2022blip}, MLLMs~\cite{bai2025qwen2,wang2024qwen2,chen2024internvl}) as evaluators.
These approaches aim to maximize the utilization of model capabilities for jointly assessing prompts and images. However, these evaluation methods primarily focus on the prompt following ability of text-to-image models, often neglecting other critical aspects. Some alternative methods(MJHQ-30K~\cite{li2024playground}, HPSv2~\cite{wu2023human}, Pick-a-pic~\cite{kirstain2023pick}) have attempted to assess the aesthetic quality of images generated by these models. 
Recently, to keep up with the advances in text-to-image models, evaluation frameworks such as WISE~\cite{niu2025wise}, WorldGenBench~\cite{zhang2025worldgenbench}, Commonsense-T2I~\cite{fu2024commonsenseT2I}, and PhyBench~\cite{qiu2025phybench} have been developed to assess the models' knowledge and reasoning capabilities.

\section{Benchmark}\label{sec:benchmark}
\subsection{Benchmark Overview}

With the rapid advancement of text-to-image (T2I) models,
the existing evaluation frameworks for T2I models urgently require improvement to measure model strengths and weaknesses comprehensively.
Early studies have explored evaluation methods from diverse perspectives, such as compositional text-to-image generation tasks~\cite{chefer2023attend,huang2023t2icompbench,fu2024commonsenseT2I}, including concept correlation, attribute binding (focusing on color attributes), and spatial relationship modeling.
However, evaluations solely focusing on compositional content exhibit significant limitations~\cite{ghosh2023geneval,hu2024ella,han2024evalmuse40kreliablefinegrainedbenchmark}, failing to adequately address broader natural language understanding and other quantitative image assessment dimensions.

In response, recent evaluation frameworks like GenEval~\cite{ghosh2023geneval}, EvalMuse~\cite{han2024evalmuse40kreliablefinegrainedbenchmark}, and DPG-Bench~\cite{hu2024ella} adopt an object-centric structured paradigm to quantify T2I model performance on specific tasks. Nevertheless, these methods predominantly rely on vision-language models (VLMs)~\cite{Qwen,Qwen2.5}, object detection models~\cite{cheng2021mask2former,chen2019mmdetection} or visual question answering (VQA)~\cite{antol2015vqa} models for element-level alignment assessment, suffering from notable deficiencies in evaluating style consistency and text rendering accuracy, with a lack of high-precision metrics. 
Additionally, human-based evaluation is prohibitively costly in terms of time and resources, making automated evaluation with limited prompts increasingly critical.
Notably, with the rapid development of reasoning-oriented models~\cite{openai2024gpt4o_image}, this study proactively introduces reasoning task evaluation to accurately measure models' knowledge representation and reasoning-driven image generation capabilities. 

Table~\ref{tab:comparison} systematically reviews the advantages and disadvantages of recent evaluation methods, presenting a comprehensive framework named OneIG-Bench from six dimensions: scene coverage, prompt distribution diversity, evaluation content, multilingualism support, automation level and leaderboard availability. OneIG-Bench covers core T2I scenarios (e.g., style image generation, text rendering, reasoning-based drawing), supports multi-format prompt inputs (including long/short texts and phrase/tag-based prompts), and employs customized automated metrics for different scenarios. Experimental results demonstrate that this framework effectively identifies performance bottlenecks and strengths of current models, providing a scientific basis for T2I model optimization. \textbf{Hereafter, unless otherwise specified, OneIG-Bench refers to OneIG-Bench-EN.}

\begin{table}[h]
\centering
\scriptsize
\renewcommand{\arraystretch}{1.2} 
    \caption{\small \textbf{Comparison between OneIG-Bench and other previous benchmarks}. In the column of prompt diversity, \textbf{L} denotes long prompt, \textbf{S} denotes the short prompt, \textbf{NP} denotes the natural language prompt, \textbf{T} denotes the tag-based prompt, and \textbf{P} denotes the phrase-based prompt.}
    \vspace{0.2cm}
    \resizebox{\columnwidth}{!}{
    \begin{tabular}{l|cccc|ccc|cccc|c|c|c}
        \toprule
        \multirow{2}{*}{\raisebox{-4\height}{\textbf{Benchmark}}}
 & \multicolumn{4}{c|}{\textbf{Scenes}} & \multicolumn{3}{c|}{\textbf{Prompt Diversity}} & \multicolumn{4}{c|}{\textbf{Evalution}} 
&  \multirow{2}{*}{\rotatebox{90}{\textbf{Multilingualism}}} 
&  \multirow{2}{*}{\rotatebox{90}{\textbf{Auto Eval}}}
&  \multirow{2}{*}{\rotatebox{90}{\textbf{Leaderboard}}} \\
        \cline{2-12} 
& \rotatebox{90}{\textbf{General}} 
& \rotatebox{90}{\textbf{Style}} 
& \rotatebox{90}{\textbf{Text}} 
& \rotatebox{90}{\textbf{Reason}} 
& \rotatebox{90}{\textbf{Length}} 
& \rotatebox{90}{\textbf{Type}} 
& \rotatebox{90}{\textbf{Count}} 
& \rotatebox{90}{\textbf{Alignment~~~}} 
& \rotatebox{90}{\textbf{Text}} 
& \rotatebox{90}{\textbf{Style}} 
& \rotatebox{90}{\textbf{Diversity}} 
& & \\
        \midrule
        PartiPrompt~\cite{yu2022scaling}  & \cmark       & \xmark                        & \xmark                    & \xmark                                & L, S                         & NP&  1,600                   & \xmark                      & \xmark             & \xmark                & \xmark       & \xmark      & \xmark                 & \xmark \\
        DrawBench~\cite{saharia2022photorealistic}   & \cmark       & \xmark                        & \cmark                    & \xmark                                & S                               & NP&  200                        & \xmark                               & \xmark                      & \xmark                         & \xmark              & \xmark        & \xmark                 & \xmark \\
        TIFA~\cite{hu2023tifa}         & \cmark       & \xmark                        & \xmark                    & \xmark                                & S                               & NP & 4,000                   & \cmark                      & \xmark             & \xmark          & \xmark      & \xmark             & \cmark                 & \xmark \\
        T2ICompBench~\cite{huang2023t2icompbench} & \cmark       & \xmark                        & \xmark                    & \xmark                                & S                               & NP& 6,000                     & \cmark                      & \xmark             & \xmark                & \xmark     & \xmark        & \cmark                 & \xmark \\
        GenEval ~\cite{li2024genai}      & \cmark       & \xmark                        & \xmark                    & \xmark                                & S                               & NP&   553                        & \cmark                               & \xmark                      & \xmark                         & \xmark            & \xmark          & \cmark                 & \xmark \\
        EVALALIGN~\cite{tan2024evalalign}    & \cmark       & \cmark                        & \cmark                    & \cmark                                & S                               & NP& 100             & \cmark                               & \xmark                      & \xmark                         & \xmark            & \xmark          & \cmark                 & \xmark \\
        WISE~\cite{niu2025wise}         & \cmark       & \xmark                        & \xmark                    & \cmark                                & S                               &NP &  1,000               & \cmark                               & \xmark                      & \xmark                         & \xmark             & \xmark         & \cmark                 & \xmark \\
        EvalMuse~\cite{han2024evalmuse40kreliablefinegrainedbenchmark}    & \cmark       & \xmark                        & \xmark                    & \xmark                                & S                               & NP& 199                       & \cmark                               & \xmark                      & \xmark                         & \xmark            & \xmark          & \cmark                 & \cmark \\
        DPG-Bench ~\cite{hu2024ella}   & \cmark       & \xmark                        & \xmark                    & \xmark                                & L                                & NP &   1,065                 & \cmark                               & \xmark                      & \xmark                         & \xmark              & \xmark        & \cmark                 & \xmark \\
        \midrule
        \textbf{OneIG-Bench}  & \cmark       & \cmark                        & \cmark                    & \cmark                                & L, S          & NP,T,P & $\text{2,440}^*$                & \cmark                               & \cmark                      & \cmark                         & \cmark             & \cmark         & \cmark                 & \cmark \\ 
        \bottomrule
    \end{tabular}
    \label{tab:comparison}
    }
    \vspace{0.1cm}
    \begin{tablenotes}
        \small
        \item{
        $*$:\textbf{OneIG-Bench} consists of two subsets: \textbf{OneIG-Bench-EN} and \textbf{OneIG-Bench-ZH}. \textbf{OneIG-Bench-EN} includes $245$ \textit{Anime and Stylization} prompts, $244$ \textit{Portrait} prompts, $206$ \textit{General Objectz} prompts, $200$ \textit{Text Rendering} prompts, and 225 \textit{Knowledge and Reasoning} prompts. \textbf{OneIG-Bench-ZH} comprises manually translated and verified Chinese versions of the prompts in \textbf{OneIG-Bench-EN}, along with additional prompts from the Multilingualism category, totaling $1,320$ prompts. In total, \textbf{OneIG-Bench} contains $2,440$ prompts.}
    \end{tablenotes}
    \vspace{-0.3cm}
\end{table}

\subsection{Benchmark Construction}

\begin{figure}[t]
    \centering
    \setlength{\abovecaptionskip}{1pt}
    \includegraphics[width=.98\linewidth]{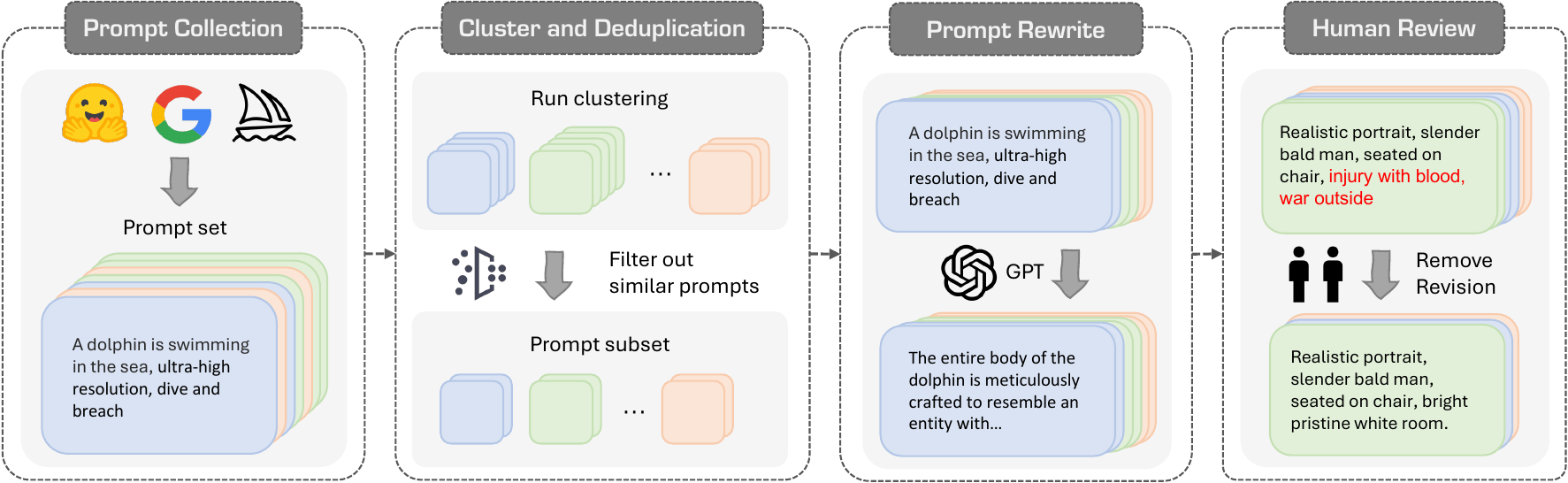}
    \vspace{0.2cm}
    \caption{\small \textbf{The construction pipeline of OneIG-Bench}. The construction pipeline comprises four methodical steps to derive our final assessment prompts, ensuring the diversity and comprehensiveness of the benchmark.}
    \label{fig:construction}
    \vspace{-0.35cm}
\end{figure}

In constructing the evaluation prompt set, as illustrated in Figure~\ref{fig:construction}, we have established five core steps to ensure the diversity and comprehensiveness of the prompt set, aligning with the design principles of our benchmark framework. 

In the first step, we curated prompts and generation scenes by filtering publicly accessible internet data, user inputs, and some established datasets, thereby ensuring that the benchmark focuses on content aligned with real-world user needs rather than rare or specialized contexts.

In the second step, we apply a clustering approach to balance the distribution of prompts across different scenes and semantic dimensions, ensuring that no single category dominates. Within each cluster, we identify some prompts with high semantic overlap, which can compromise evaluation diversity; thus, we implement a deduplication pipeline that filters out redundant prompts based on cosine similarity to the cluster center embeddings. Besides, we endeavor to maintain a relatively consistent proportion of prompts across the five defined dimensions during prompt selection.

Following deduplication and subset curation, we employ a large language model (LLM) to rewrite the original prompts. Concurrently, constraints are applied to the word-level length distribution of prompts, enabling a structured analysis of model performance across varying text complexities. The prompt corpus is intentionally structured into three length categories: concise texts (fewer than 30 words), mid-complexity scenarios (30–60 words), and elaborate texts (exceeding 60 words). The corresponding ratio of three categories is around 1:2:1.

Finally, we performed manual reviews to filter out prompts containing sensitive content or conflicting semantics, ensuring the rationality of all benchmark prompts. This critical step not only enhances the dataset’s quality and reliability but also guarantees its suitability for fair and unbiased model evaluation across diverse generative scenarios.

Through a rigorously designed construction pipeline, this structured evaluation framework facilitates a granular assessment of model performance across multiple dimensions.

\section{Evaluation}\label{sec:evaluation}
\subsection{Metrics}
\label{subsec:metrics}
To illustrate the metrics used in our benchmark, we provide the following general definitions. Given the text prompt set $\mathbb{T} = \{T_1,T_2,...,T_n\}$, for each $T_k \in \mathbb{T}$, the generated images are defined as $\mathbb{G}^k = \{G_1^k, G_2^k, ...,G_m^k\}$, where $m$ denotes the number of generated images, and a total of $m \times n$ images are generated as evaluation images.. To evaluate the style score, for each style-specific prompt $T_k \in \mathbb{T}$, we define $\mathbb{R} = \{R_1^k, R_2^k, R_3^k\}$ as the set of corresponding style reference images, where $l$ denotes the number of reference images. For text rendering, the original target text string in $T_k$ is defined as $s_k$, and the generated string in the corresponding image is defined as $\hat{s}_k$.

\textbf{Semantic alignment.} We follow the method introduced in DSG~\cite{Cho2024DSG} on \textit{General Object}, \textit{Portrait}, \textit{Anime and Stylization} to assess the semantic matching degree of each text-image sample. For each prompt, we initially leverage GPT-4o~\cite{openai2024gpt4o} to generate a question dependency graph. In the process of constructing the graph, our focus lies in formulating questions related to the overall information, spatial relationships, and the attributes of diverse objects. During the evaluation, Qwen2.5-VL-7B~\cite{Qwen2.5} is utilized to answer questions derived from the corresponding prompt and the generated image. A score of 1 is assigned for each correctly answered question. However, when calculating the aggregate score for a prompt, leaf node scores are conditionally validated: they contribute to the total score only if the root node question is answered correctly; otherwise, leaf node scores are reset to 0. The final score for each prompt is computed as the sum of validated scores divided by the total number of questions. 

\textbf{Text Rendering.} To accurately evaluate the text-generation capability of text-to-image models, we designed specialized text evaluation metrics. 
To extract the generated string $\hat{s}_k$, we first use a state-of-the-art Vision-Language Model (VLM, e.g., Qwen2.5-VL-7B~\cite{Qwen2.5}) to parse the text string and then clean it by removing symbols and consecutive spaces
The metrics are as follows:

(1) \textbf{Edit Distance (ED)}: 
    It is defined as the average edit distance between the generated text of evaluation images and the ground-truth text to be generated. We define the edit distance score of the \(i\)-th evaluation image as $\text{ED}_i=\mathcal{L}(\hat{s}_i,s_i)$, where $\mathcal{L}(\cdot) $ denotes the Levenshtein distance function. Therefore, the overall edit distance score of the model is: $\text{ED} = \frac{1}{n \times m} \sum_{i=1}^{n \times m} \text{ED}_i.$
    
(2) \textbf{Completion Rate (CR)}: 
  It is defined as the proportion of the number of evaluation images with completely correct generated text to the total number of evaluation images. We define that the score for the \(i\)-th generated image in this criterion is \(\text{CR}_i = 1 \) if and only if the edit distance score of the \(i\)-th image is 0,\textit{ i.e.}, \( \text{ED}_i = 0 \). Therefore, the overall CR score of the model is defined as $\text{CR} = \sum_{i=1}^{m\times n}\text{CR}_i / ( m\times n)$.

(3) \textbf{Word Accuracy (WAC)}: It is defined as a metric representing the ratio of all correctly generated words to the total number of words in the original target text strings among all prompts.

Based on our analysis on the evaluation results, we define the edit distance upper bound as $\phi$, and edit distance exceeding $\phi$ indicate deficiencies in the model's text-rendering capability. To facilitate metric ranking and readability, we integrated three metrics into a composite metric and defined the text score($\mathrm{S}_{\text{text}}$) as follows:

\begin{equation}
    \mathrm{S}_{\text{text}} = 1 - \min(\phi, \text{ED}) * (1 - \text{CR}) * (1 - \text{WAC}) / \phi
    \label{eq:s_text}
\end{equation}
where $\phi=100$ in OneIG-Bench. Considering that Chinese characters typically occupy twice as many bytes as English letters, we use $\phi=50$ in Equation~\ref{eq:s_text} when computing the text score for OneIG-Bench-ZH, in order to maintain a comparable normalization scale.

\textbf{Knowledge and Reasoning.} We perform the evaluations using GPT-4o~\cite{openai2024gpt4o} and LLM2CLIP~\cite{huang2024llm2clip}. Specifically, GPT-4o is responsible for generating the textual reasoning answers, which serve as the core reference for evaluation. LLM2CLIP then measures the alignment between text and image by calculating the cosine similarity between the GPT-4o-generated answer and the corresponding generated image.

\textbf{Style.} For stylization evaluation, we curated multiple reference images per style and employed a dual-style extraction framework to mitigate bias and enhance robustness. Specifically, the CSD~\cite{somepalli2024measuring} model and OneIG style image encoder(fine-tuned from CLIP~\cite{radford2021learning}, using images generated by CSGO ~\cite{xing2024csgo}) are leveraged to encode the images and generate the corresponding embeddings. Subsequently, for each encoder, we quantitatively assess the model’s style capacity by computing the cosine similarity between the style embeddings of generated images and those of reference images. For each generated image, we choose the maximum similarity as the score of the image. The style similarity($\mathrm{S}_{[\text{csd}, \text{oneig}]}$) of one style image encoder is defined as:
\begin{equation}
    \mathrm{S}_{[\text{csd}, \text{oneig}]} = \frac{1}{n} \sum_{k=1}^n \left[ \frac{1}{m} \sum_{i=1}^{m} \left( \max_{j} \cos(\mathcal{F}(G_i^k), \mathcal{F}(R_j^k)) \right) \right]
\end{equation}

where $\mathcal{F}(\cdot)$ denotes the corresponding style image encoder. The final style score $\mathrm{S}_{\text{style}}$ is defined as $\mathrm{S}_{\text{style}} = (\mathrm{S}_{\text{csd}} + \mathrm{S}_{\text{oneig}}) / 2$.

\textbf{Diversity.} In addition, we apply a form of similarity calculation to evaluate the diversity of the generation of the model introduced in ~\cite{gandikota2025distilling}. The calculation of diversity is defined as follows: for a given model, we first compute the pairwise cosine similarity between every pair of images generated from the same prompt within a set of multiple generated outputs. These similarities are averaged per prompt to yield an intra-prompt similarity score. We then aggregate these intra-prompt averages across all prompts in the evaluation dataset using a global mean, resulting in an overall diversity metric. Following ~\cite{gandikota2025distilling}, we also applied DreamSim~\cite{fu2023dreamsim} to compute the cosine similarity and the formula is as follows: 
\begin{equation}
    \text{SIM}_{ij}^k = \cos(\mathcal{F}(G_i^k), \mathcal{F}(G_j^k))
\end{equation}

where $\text{SIM}_{ij}^k$ denotes the cosine similarity between images generated by one text prompt, $\mathcal{F}(\cdot)$ represents DreamSim~\cite{fu2023dreamsim} model. Thus, the diversity score($\mathrm{S}_{\text{diversity}}$) can be defined:
\begin{equation}
    \mathrm{S}_{\text{diversity}} = \frac{1}{n} \sum_{k=1}^{n} \left[ \frac{1}{C_m^2} \sum_{i=1}^m \sum_{j=i+1}^m \left(1 - \text{SIM}_{ij}^k\right)\right].
\end{equation}

\subsection{Results and Analysis}
\label{subsec:results}

We evaluate a range of well-known image generation models on our benchmark, including unified multimodal models (Janus-Pro~\cite{chen2025janus}, BLIP3-o~\cite{chen2025blip3}, BAGEL~\cite{deng2025bagel}, Show-o2~\cite{xie2025showo2}, OmniGen2~\cite{wu2025omnigen2}), open-source models (Stable Diffusion 1.5~\cite{Rombach_2022_CVPR}, Stable Diffusion XL~\cite{podell2023sdxl}, Stable Diffusion 3.5~\cite{2024sd3.5}, Flux.1-dev~\cite{flux2024}, CogView4~\cite{2025cogview4}, SANA~\cite{xie2025sana}, Lumina-Image 2.0~\cite{lumina2}, and HiDream-I1-Full~\cite{2025hidreami1}) with A800 GPUs, as well as closed-source models (Imagen3~\cite{2024Imagen3}, Recraft V3~\cite{2024recraftv3}, Kolors 2.0~\cite{2025Kolors2}, Seedream 3.0~\cite{gao2025seedream}, Imagen4~\cite{2025Imagen4} and GPT-4o~\cite{openai2024gpt4o_image}). To present a comprehensive comparison, we aggregate evaluation metrics across multiple dimensions, as summarized in Table~\ref{tab:overall_quantitative}. \textbf{We define the sets of images generated based on the OneIG-Bench prompt categories \textit{General Object} $\mathcal{O}$, \textit{Portrait} $\mathcal{P}$, \textit{Anime and Stylization} $\mathcal{A}$ (prompts without stylization), $\mathcal{S}$ (prompts with stylization), \textit{Text Rendering} $\mathcal{T}$, \textit{Knowledge and Reasoning} $\mathcal{KR}$ and \textit{Multilingualism} $\mathcal{L}$}. A more detailed, fine-grained analysis of individual dimensions on OneIG-Bench follows in the subsequent sections and the appendix.

\begin{table}[h]
    \centering
    \scriptsize
    \caption{\small \textbf{Overall quantitative comparison of different methods on OneIG-Bench}. The table showcases the results of five core metrics for various methods. \legendsquare{colorfirst} \legendsquare{colorsecond} \legendsquare{colorthird} \legendsquare{colorfourth} \legendsquare{colorfifth} indicate the first, second, third, fourth, and fifth performance, respectively.}
    \vspace{0.2cm}
     \resizebox{\columnwidth}{!}{
    \begin{tabular}{lccccc}
        \toprule
        \textbf{Method} & \textbf{~~~Alignment~}$\uparrow$ \textbf{~~}& \textbf{~~~~~~~Text~}$\uparrow$ \textbf{~~~~~~}& \textbf{~~~~Reasoning~}$\uparrow$ \textbf{~~~}& \textbf{~~~~~~~Style~}$\uparrow$ \textbf{~~~~~~}& \textbf{~~~~Diversity~}$\uparrow$ \textbf{~~~}\\
        \midrule
        Assessment Sets & $\mathcal{O}, \mathcal{P}, \mathcal{A},  \mathcal{S}$ & $\mathcal{T}$ & $\mathcal{KR}$ & $\mathcal{S}$ & $\mathcal{O}, \mathcal{P}, \mathcal{A},  \mathcal{S}, \mathcal{T}, \mathcal{KR}$ \\
        \midrule
        Janus-Pro~\cite{chen2025janus} & 0.553  & 0.001  &   0.139     & 0.276 & \ranksecond{0.365} \\
        BLIP3-o~\cite{chen2025blip3} & 0.711  & 0.013  &   0.223      & 0.361 & 0.229 \\
        BAGEL~\cite{deng2025bagel} & 0.769  & 0.244  &   0.173    & 0.367 & 0.251
        \\
        BAGEL+CoT~\cite{deng2025bagel} & 0.793 & 0.020  &   0.206    & \rankfourth{0.390} & 0.209 \\
        Show-o2-1.5B~\cite{xie2025showo2} & 0.798 & 0.002 & 0.219& 0.317 & 0.186 \\
        Show-o2-7B~\cite{xie2025showo2} & 0.817 & 0.002 & 0.226 & 0.317 & 0.177\\
        OmniGen2~\cite{wu2025omnigen2} & 0.804 & 0.680 & 0.271 & 0.377 & 0.242 \\       
        \midrule
        Stable Diffusion 1.5~\cite{Rombach_2022_CVPR} & 0.565 & 0.010 & 0.207 & 0.383 & \rankfirst{0.429} \\
        Stable Diffusion XL~\cite{podell2023sdxl} & 0.688 & 0.029 & 0.237 & 0.332 & \rankfourth{0.296} \\
        Stable Diffusion 3.5 Large~\cite{2024sd3.5} & 0.809 & 0.629 & 0.294 & 0.353 & 0.225 \\
        Flux.1-dev~\cite{flux2024} & 0.786 & 0.523 & 0.253 & 0.368 & 0.238 \\
        CogView4~\cite{2025cogview4} & 0.786 & 0.641 & 0.246 & 0.353 & 0.205 \\
        SANA-1.5 1.6B (PAG)~\cite{xie2025sana} & 0.762 & 0.054 & 0.209 & \rankfifth{0.387} & 0.222 \\
        SANA-1.5 4.8B (PAG)~\cite{xie2025sana} & 0.765 & 0.069 & 0.217 & \rankthird{0.401} & 0.216 \\
        Lumina-Image 2.0~\cite{lumina2} & 0.819 & 0.106 & 0.270 & 0.354 & 0.216 \\
        HiDream-I1-Full~\cite{2025hidreami1} & \rankfourth{0.829} & \rankfifth{0.707} & \rankfourth{0.317} & 0.347 & 0.186 \\
        \midrule
        Imagen3~\cite{2024Imagen3} & \rankthird{0.843} & 0.343 & \rankfifth{0.313} & 0.359 & 0.188 \\
        Recraft V3~\cite{2024recraftv3} & 0.810 & \rankfourth{0.795} & \rankthird{0.323} & 0.378 & 0.205 \\
        Kolors 2.0~\cite{2025Kolors2} & \rankfifth{0.820} & 0.427 & 0.262 & 0.360 & \rankthird{0.300} \\
        Seedream 3.0~\cite{gao2025seedream} & 0.818 & \rankfirst{0.865} & 0.275 & \ranksecond{0.413} & \rankfifth{0.277} \\
Imagen4~\cite{2025Imagen4} & \rankfirst{0.857} & \rankthird{0.805} & \ranksecond{0.338} & 0.377 & 0.199\\
        GPT-4o~\cite{openai2024gpt4o_image} & \ranksecond{0.851} & \ranksecond{0.857} & \rankfirst{0.345} & \rankfirst{0.462} & 0.151 \\
        \bottomrule
    \end{tabular}
    }
    \label{tab:overall_quantitative}
\end{table}

\begin{table}[ht]
    \centering
    \scriptsize
    \caption{\small \textbf{Overall quantitative comparison of different methods on OneIG-Bench-ZH}. The table showcases the results of five core metrics for various methods. \legendsquare{colorfirst} \legendsquare{colorsecond} \legendsquare{colorthird} \legendsquare{colorfourth} \legendsquare{colorfifth} indicate the first, second, third, fourth, and fifth performance, respectively.}
    \vspace{0.2cm}
    \resizebox{0.95\columnwidth}{!}{
    \begin{tabular}{lccccc}
        \toprule
        \textbf{Method} &
        \textbf{Alignment~}$\uparrow$ & 
        \textbf{~~~~~~Text~}$\uparrow$ \textbf{~~~~} & 
        \textbf{Reasoning~}$\uparrow$ & 
        \textbf{~~~~~~Style~}$\uparrow$ \textbf{~~~~} & 
        \textbf{Diversity~}$\uparrow$ \\
        \midrule
        Assessment Sets & \makecell[l]{ $~~~~~\mathcal{O}_{zh}, \mathcal{P}_{zh},$ \\ $\mathcal{A}_{zh}, \mathcal{S}_{zh}, \mathcal{L}_{zh}$} & $\mathcal{T}_{zh}$ & $\mathcal{KR}_{zh}$ & $\mathcal{S}_{zh}$ & \makecell[l]{$\mathcal{O}_{zh}, \mathcal{P}_{zh}, \mathcal{A}_{zh}, \mathcal{S}_{zh}, $ \\$~~~~\mathcal{T}_{zh}, \mathcal{KR}_{zh}, \mathcal{L}_{zh}$} \\
        \midrule
        Janus-Pro~\cite{chen2025janus} & 0.324          & 0.148          & 0.104        & 0.264          & \rankfirst{0.358} \\
        BLIP3-o~\cite{chen2025blip3} & 0.608          & 0.092          & 0.213        & \rankfourth{0.369}          & 0.233          \\
        BAGEL~\cite{deng2025bagel} & 0.672          & \rankfourth{0.365}          & 0.186 & \rankfifth{0.357}          & \rankfourth{0.268}          \\
        BAGEL+CoT~\cite{deng2025bagel} & \rankfifth{0.719}   & 0.127  &   0.219    & \rankthird{0.385} & 0.197 \\
        \midrule
        Cogview4~\cite{2025cogview4} & 0.700          & 0.193          & \rankfourth{0.236}        & 0.348          & 0.214          \\
        Lumina-Image 2.0~\cite{lumina2} & \rankfourth{0.731}          & 0.136          & 0.221        & 0.343          & 0.240           \\
        HiDream-I1-Full~\cite{2025hidreami1} & 0.620          & \rankfifth{0.205}          & \rankthird{0.256}        & 0.304          & \rankthird{0.300}            \\
        \midrule
        Kolors 2.0~\cite{2025Kolors2} & \rankthird{0.738}          & \rankthird{0.502}          & \rankfifth{0.226}        & 0.331          & \ranksecond{0.333}          \\
        Seedream 3.0~\cite{gao2025seedream} & \ranksecond{0.793}          & \rankfirst{0.928} & \ranksecond{0.281}       & \ranksecond{0.397}          & \rankfifth{0.243}          \\
        GPT-4o~\cite{openai2024gpt4o_image} & \rankfirst{0.812} & \ranksecond{0.650}           & \rankfirst{0.300} & \rankfirst{0.449} & 0.159      \\
        \bottomrule
    \end{tabular}
    }
    \label{tab:overall_quantitative_zh}
\end{table}

\begin{figure}[ht]
    \centering
    \setlength{\abovecaptionskip}{1pt}
    \includegraphics[width=.98\linewidth]{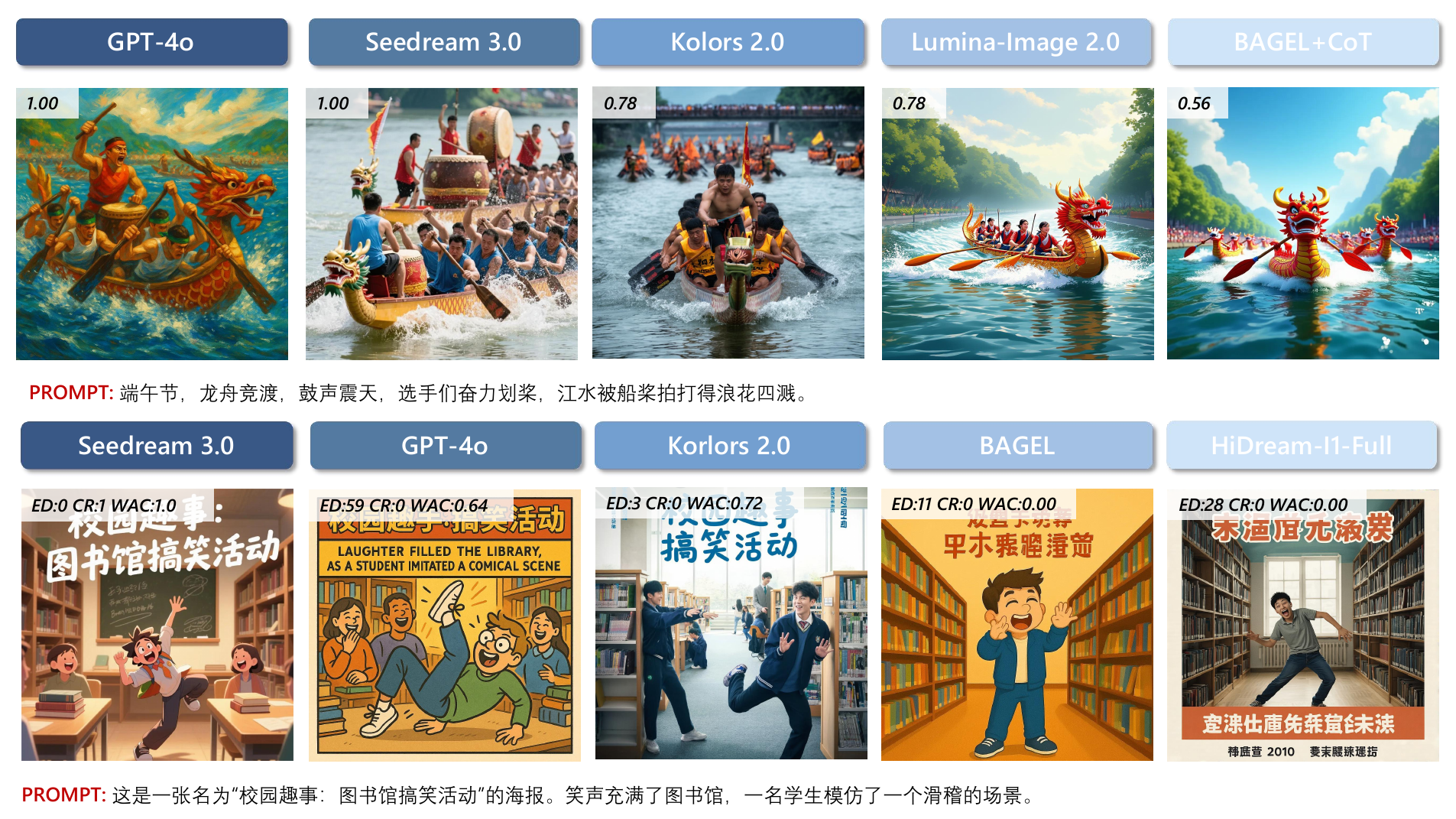}
    \caption{\small \textbf{An illustration of the generation results and the corresponding scores on OneIG-Bench-ZH.} The first row shows the alignment results and the second row shows the the text rendering results. And the evaluation scores are displayed in the upper left corner of the image.}
    \label{fig:ranking_zh}
\end{figure}

Meanwhile, we evaluate a relatively smaller set of well-known image generation models on OneIG-Bench-ZH, including unified multimodal models (Janus-Pro~\cite{chen2025janus}, BLIP3-o~\cite{chen2025blip3}, BAGEL~\cite{deng2025bagel}), open-source models (CogView4~\cite{2025cogview4}, Lumina-Image 2.0~\cite{lumina2}, and HiDream-I1-Full~\cite{2025hidreami1}) using A800 GPUs, as well as closed-source models (Kolors 2.0~\cite{2025Kolors2}, Seedream 3.0~\cite{gao2025seedream}, and GPT-4o~\cite{openai2024gpt4o_image}). In OneIG-Bench-ZH, \textit{Multilingualism} part consists of 100 culture-related prompts and 100 portrait-related prompts. To provide a clear overall comparison, the evaluation results across multiple dimensions are summarized in Table~\ref{tab:overall_quantitative_zh}. On OneIG-Bench-ZH, GPT-4o~\cite{openai2024gpt4o_image} demonstrates outstanding performance, ranking first across most evaluation dimensions in Table~\ref{tab:overall_quantitative_zh} and Figure~\ref{fig:ranking_zh}.
In contrast, Seedream 3.0~\cite{gao2025seedream} excels particularly in Chinese text rendering, significantly outperforming GPT-4o.
However, most models show limited capability in generating Chinese text, with many nearly incapable of producing legible Chinese characters.
It is also worth noting that for most models, performance on alignment, reasoning, and style dimensions is slightly weaker on OneIG-Bench-ZH than on OneIG-Bench, indirectly reflecting that their ability to understand and generate Chinese semantics still requires further improvement.

\subsubsection{Semantic Alignment and Diversity}

In the alignment dimension shown in Table~\ref{tab:overall_quantitative}, Imagen4~\cite{2025Imagen4}, GPT-4o~\cite{openai2024gpt4o_image} and Imagen3~\cite{2024Imagen3} consistently outperform other models, with Imagen4 and GPT-4o showing superior alignment performance.
Furthermore, as indicated in Table~\ref{tab:alignment}, most models achieve significantly higher alignment accuracy when responding to natural language prompts than to tag-based or phrase-based prompts. A possible explanation is that tag and phrase prompts may introduce ambiguity, such as attribute confusion between entities or logical inconsistencies, which complicates semantic alignment.
Longer prompts tend to involve greater semantic complexity and structural variation, often resulting in lower alignment scores. 
Notably, models incorporating T5~\cite{raffel2020exploring} or other large language models appear more robust in handling long prompts, exhibiting less semantic degradation.

\begin{table}[ht]
\renewcommand{\arraystretch}{1.2}
\centering
\scriptsize
\caption{\small \textbf{The alignment and diversity evaluation results}. \textbf{NP} denotes the natural
language prompt. \textbf{T\&P} denotes the tag-based and phrase-based prompt. \textbf{Short}, \textbf{Medium} and \textbf{Long} represent the length of the prompts, where \textbf{Short} denote the number of words is less than 30, \textbf{Medium} denotes the number between 30 and 60, and \textbf{Long} denotes the number exceeding 60. \legendsquare{colorfirst} \legendsquare{colorsecond} \legendsquare{colorthird} \legendsquare{colorfourth} \legendsquare{colorfifth} indicate the first, second, third, fourth, and fifth performance respectively.}
\vspace{0.2cm}
\scalebox{0.95}{
\begin{tabular}{l|cc|ccc|cc|ccc}
\toprule
\multirow{2}{*}{\textbf{Method}} & \multicolumn{5}{c|}{\textbf{Alignment}$\uparrow$} & \multicolumn{5}{c}{\textbf{Diversity}$\uparrow$}  \\
\cline{2-11}
& \textbf{NP} & \textbf{T\&P} & \textbf{Short} & \textbf{Medium} & \textbf{Long} & \textbf{NP} & \textbf{T\&P} & \textbf{Short} & \textbf{Medium} & \textbf{Long} \\ 
\midrule
Janus-Pro~\cite{chen2025janus} & 0.557  & 0.533  & 0.609   & 0.548 & 0.515 & \ranksecond{0.372} & \ranksecond{0.304} & \ranksecond{0.407} & \ranksecond{0.332} & \ranksecond{0.347} \\
BLIP3-o~\cite{chen2025blip3} & 0.719  & 0.671  & 0.754   & 0.712 & 0.674 &   0.237 & 0.161 & 0.283 & 0.192 & 0.198 \\
BAGEL~\cite{deng2025bagel} & 0.776  & 0.734  & 0.782   & 0.769 & 0.759 &    0.257 & 0.197 & 0.344 & 0.190 & 0.194 \\
BAGEL+CoT~\cite{deng2025bagel} & 0.798 & 0.767 & 0.824 & 0.793 & 0.767 &    0.214 & 0.164 & 0.244 & 0.184 & 0.189 \\
Show-o2-1.5B~\cite{xie2025showo2} & 0.805 & 0.760 & 0.800 & 0.799 & 0.793 & 0.191 & 0.142 & 0.226 & 0.162 & 0.157 \\
Show-o2-7B~\cite{xie2025showo2} & 0.825 & 0.778 & 0.825 & 0.819 & 0.807 & 0.182 & 0.130 & 0.216 & 0.152 & 0.151 \\
OmniGen2~\cite{wu2025omnigen2} & 0.812 & 0.768 & 0.809 & 0.805 & 0.799 & 0.250 & 0.174 & 0.329 & 0.185 & 0.190 \\
\midrule
Stable Diffusion 1.5~\cite{Rombach_2022_CVPR} & 0.570 & 0.541 & 0.616 & 0.558 & 0.537 & \rankfirst{0.434} & \rankfirst{0.381} & \rankfirst{0.481} & \rankfirst{0.389} & \rankfirst{0.403} \\
Stable Diffusion XL~\cite{podell2023sdxl} & 0.688 & 0.685 & 0.732 & 0.685 & 0.657 & \rankfourth{0.303} & \rankfourth{0.239} & \rankfifth{0.337} & \rankthird{0.263} & \rankthird{0.278} \\
Stable Diffusion 3.5 Large~\cite{2024sd3.5} & 0.818 & 0.762 & 0.826 & 0.808 & 0.795 & 0.229 & 0.194 & 0.267 & 0.194 & 0.206 \\
Flux.1-dev~\cite{flux2024} & 0.791 & 0.759 & 0.794 & 0.785 & 0.780 & 0.243 & 0.190 & 0.302 & 0.194 & 0.199 \\
CogView4~\cite{2025cogview4} & 0.796 & 0.737 & 0.792 & 0.788 & 0.777 & 0.211 & 0.153 & 0.277 & 0.158 & 0.159 \\
SANA-1.5 1.6B(PAG)~\cite{xie2025sana} & 0.769 & 0.726 & 0.770 & 0.764 & 0.752 & 0.231 & 0.153 & 0.284 & 0.177 & 0.192 \\
SANA-1.5 4.8B(PAG)~\cite{xie2025sana} & 0.773 & 0.729 & 0.782 & 0.767 & 0.749 & 0.223 & 0.154 & 0.264 & 0.181 & 0.191 \\
Lumina-Image 2.0~\cite{lumina2} & \rankfifth{0.825} & 0.788 & 0.829 & 0.819 & \rankfourth{0.812} & 0.224 & 0.149 & 0.282 & 0.171 & 0.180 \\
HiDream-I1-Full~\cite{2025hidreami1} & \rankfourth{0.834} & \rankfifth{0.806} & \rankfourth{0.849} & \rankfourth{0.834} & 0.806 & 0.192 & 0.142 & 0.260 & 0.139 & 0.140 \\
\midrule
Imagen3~\cite{2024Imagen3} & \rankthird{0.849} & \rankfourth{0.809} & \rankthird{0.859} & \rankthird{0.841} & \rankthird{0.832} & 0.189 & 0.173 & 0.246 & 0.146 & 0.153 \\
Recraft V3~\cite{2024recraftv3} & 0.816 & 0.781 & 0.838 & 0.809 & 0.788 & 0.209 & 0.178 & 0.246 & 0.178 & 0.180 \\
Kolors 2.0~\cite{2025Kolors2} & 0.824 & 0.798 & \rankfifth{0.847} & 0.814 & \rankfifth{0.807} & \rankthird{0.308} & \rankfifth{0.230} & \rankthird{0.359} & \rankfourth{0.261} & \rankfourth{0.259} \\
Seedream 3.0~\cite{gao2025seedream} & 0.818 & \rankthird{0.815} & 0.838 & \rankfifth{0.825} & 0.789 & \rankfifth{0.280} & \rankthird{0.246} & \rankfourth{0.342} & \rankfifth{0.235} & \rankfifth{0.227} \\
Imagen4~\cite{2025Imagen4} & \rankfirst{0.860} & \rankfirst{0.843} & \rankfirst{0.875} & \rankfirst{0.854} & \rankfirst{0.847} & 0.199 & 0.197 & 0.276 & 0.147 & 0.149 \\
GPT-4o~\cite{openai2024gpt4o_image} & \ranksecond{0.857} & \ranksecond{0.820} & \ranksecond{0.869} & \ranksecond{0.851} & \ranksecond{0.838} & 0.154 & 0.124 & 0.177 & 0.134 & 0.134 \\
\bottomrule
\end{tabular}
}
\label{tab:alignment}
\end{table}

Diversity is informative when assessed among models with comparable levels of alignment. In real-world applications, generative models are generally expected to produce varied outputs while maintaining close adherence to the input prompts. 
Although Stable Diffusion 1.5~\cite{Rombach_2022_CVPR} and Janus-Pro~\cite{chen2025janus} achieve notably high diversity scores, these results are less indicative of true generative quality, as they largely stem from the models’ inconsistent preservation of semantic alignment in the generated images. And Kolors 2.0~\cite{2025Kolors2} exhibits outstanding diversity without compromising its alignment performance, and is regarded as a model with excellent diversity performance.

While stylization can be viewed as a specific facet of semantic alignment, performance in this dimension does not entirely coincide with overall alignment outcomes. GPT-4o~\cite{openai2024gpt4o_image} retains a clear lead in stylization, while both Seedream 3.0~\cite{gao2025seedream} and the SANA series methods also exhibit strong performance
Notably, Stable Diffusion 1.5~\cite{Rombach_2022_CVPR} demonstrates impressive stylization capabilities despite its relatively poor performance in semantic alignment. This may be attributed to its data cleaning process, which likely preserved a broad range of stylistic patterns and enabled the model to generate images with distinct stylistic characteristics.

\subsubsection{Text Rendering}

\begin{table}[ht]
\renewcommand{\arraystretch}{1.2}
\centering
\scriptsize
\caption{\small \textbf{The text rendering evaluation results.} In the table, \textbf{Short}, \textbf{Medium} and \textbf{Long} represent the length of the prompts, where \textbf{Short} denote the number of words is less than 30, \textbf{Medium} denotes the length between 30 and 60, and \textbf{Long} denotes the length exceeding 60. \legendsquare{colorfirst} \legendsquare{colorsecond} \legendsquare{colorthird} \legendsquare{colorfourth} \legendsquare{colorfifth} indicate the first, second, third, fourth, and fifth performance respectively.}
\vspace{0.2cm}
\resizebox{\textwidth}{!}{
    \begin{tabular}{l|ccc|ccc|ccc}
    \toprule
    \multirow{2}{*}{\textbf{Method}} & \multicolumn{3}{c|}{\textbf{Edit Distance (ED)}$\downarrow$} & \multicolumn{3}{c|}{\textbf{Completion Rate (CR)}$\uparrow$} & \multicolumn{3}{c}{\textbf{Word Accuracy (WAC)}$\uparrow$} \\
    \cline{2-10} 
    & \textbf{Short} & \textbf{Medium} & \textbf{Long} & \textbf{Short} & \textbf{Medium} & \textbf{Long} & \textbf{Short} & \textbf{Medium} & \textbf{Long} \\ 
    \midrule
Janus-Pro\cite{chen2025janus} & 33.041 & 59.695  & 295.020 & 0.000 & 0.000 & 0.000 & 0.001 & 0.001 & 0.000 \\
BLIP3-o\cite{chen2025blip3} & 31.627 & 59.584  & 258.510 & 0.005 & 0.000 & 0.000 & 0.014 & 0.022 & 0.012 \\
BAGEL\cite{deng2025bagel} & 25.650 & 44.453  & 242.415 & 0.005 & 0.079 & 0.000 & 0.128 & 0.288 & 0.148 \\
BAGEL+CoT\cite{deng2025bagel} & 30.550         & 57.397         & 255.220      & 0.000          & 0.005          & 0.000 & 0.024 & 0.033 & 0.011 \\
Show-o2-1.5B~\cite{xie2025showo2} & 32.205 & 59.232 & 296.580 & 0.000 & 0.000 & 0.000 & 0.002 & 0.004 & 0.001 \\
Show-o2-7B~\cite{xie2025showo2} & 31.709 & 59.179 & 294.720 & 0.000 & 0.000 & 0.000 & 0.004 & 0.006 & 0.001 \\
OmniGen2~\cite{wu2025omnigen2} & \rankthird{17.000} & \rankfifth{36.145} & 181.525 & \rankthird{0.255} & 0.145 & 0.000 & \rankthird{0.543} & \rankthird{0.614} & \rankfifth{0.534} \\
\midrule
Stable Diffusion 1.5~\cite{Rombach_2022_CVPR}   & 36.227 & 60.480 & 290.245 & 0.000 & 0.011 & 0.000 & 0.004 & 0.020 & 0.004 \\
Stable Diffusion XL~\cite{podell2023sdxl}       & 35.045 & 61.824 & 290.470 & 0.000 & 0.008 & 0.000 & 0.020 & 0.056 & 0.029 \\
Stable Diffusion 3.5 Large~\cite{2024sd3.5}     & 19.459 & 38.711 & 248.280 & \ranksecond{0.432} & \ranksecond{0.255} & 0.005 & \ranksecond{0.749} & \ranksecond{0.740} & 0.512 \\
Flux.1-dev~\cite{flux2024}                      & 26.855 & 44.189 & 227.565 & \rankfourth{0.223} & 0.161 & 0.000 & 0.387 & 0.577 & 0.430 \\
CogView4~\cite{2025cogview4}                    & \rankfifth{17.173} & \rankthird{29.437} & 193.420 & \rankfifth{0.200} & 0.150 & \rankfourth{0.010} & \rankfourth{0.437} & \rankfifth{0.593} & 0.517 \\
SANA-1.5 1.6B (PAG)~\cite{xie2025sana}          & 31.573 & 61.566 & 288.225 & 0.059 & 0.003 & 0.000 & 0.143 & 0.090 & 0.031 \\
SANA-1.5 4.8B (PAG)~\cite{xie2025sana}          & 25.027 & 55.634 & 268.025 & 0.086 & 0.000 & 0.000 & 0.228 & 0.079 & 0.030 \\
Lumina-Image 2.0~\cite{lumina2}                 & 26.259 & 54.547 & 270.530 & 0.059 & 0.013 & 0.000 & 0.199 & 0.199 & 0.083 \\
HiDream-I1-Full~\cite{2025hidreami1}            & \ranksecond{14.464} & \rankfourth{31.026} & \rankfifth{177.765} & 0.195 & \rankfifth{0.166} & 0.005 & \rankfifth{0.435} & \rankfourth{0.602} & \rankfourth{0.576} \\
\midrule
Imagen3~\cite{2024Imagen3}                      & 34.005 & 83.171 & 239.565 & 0.068 & 0.071 & 0.000 & 0.279 & 0.447 & 0.371 \\
Recraft V3~\cite{2024recraftv3}                 & \rankfourth{17.050} & \ranksecond{26.945} & \rankfirst{74.181}  & 0.050 & 0.061 & \rankfifth{0.006} & 0.267 & 0.371 & 0.430 \\
Kolors 2.0~\cite{2025Kolors2}                   & 18.236 & 37.930 & 235.953 & 0.125 & 0.118 & 0.000 & 0.374 & 0.483 & 0.257 \\
Seedream 3.0~\cite{gao2025seedream}             & \rankfirst{7.204}  & \rankfirst{20.596} & \rankfourth{169.307} & \rankfirst{0.699} & \rankfirst{0.451} & \rankfirst{0.109} & \rankfirst{0.893} & \rankfirst{0.822} & \ranksecond{0.688} \\
Imagen4~\cite{2025Imagen4}                      & 18.273 & 42.918 & \rankthird{121.625} & 0.181 & \rankthird{0.184} & \rankthird{0.060} & 0.364 & 0.575 & \rankfirst{0.745} \\
GPT-4o~\cite{openai2024gpt4o_image}             & 18.255 & 37.850 & \ranksecond{85.430}  & 0.150 & \rankfourth{0.171} & \ranksecond{0.070} & 0.323 & 0.495 & \rankthird{0.673} \\
    \bottomrule
    \end{tabular}
    }
    \label{tab:text_eval}
\end{table}

We evaluate the performance of text-to-image models with a particular focus on text rendering, which assesses how accurately these models reproduce textual content within generated images. For clearer comparison across varying textual complexities, the images are categorized by prompt length.

As shown in Table~\ref{tab:text_eval}, Seedream 3.0~\cite{gao2025seedream} achieves the best performance across nearly all sub-dimensions of completion ratio and word accuracy count, as well as in edit distance for short and medium-length prompts. Closer inspection of the generated images reveals that the relatively higher edit distances observed for Seedream 3.0 are mainly associated with the difficulty of rendering long text inputs, particularly in structured formats such as articles and PowerPoint slides. 
Images generated with long text inputs may contain more frequent textual deviations. While Seedream 3.0 performs well on certain long prompts, the overall complexity of long-text rendering contributes to an increase in edit distance.

\begin{figure}[ht]
    \centering
    \setlength{\abovecaptionskip}{1pt}
    \includegraphics[width=.98\linewidth]{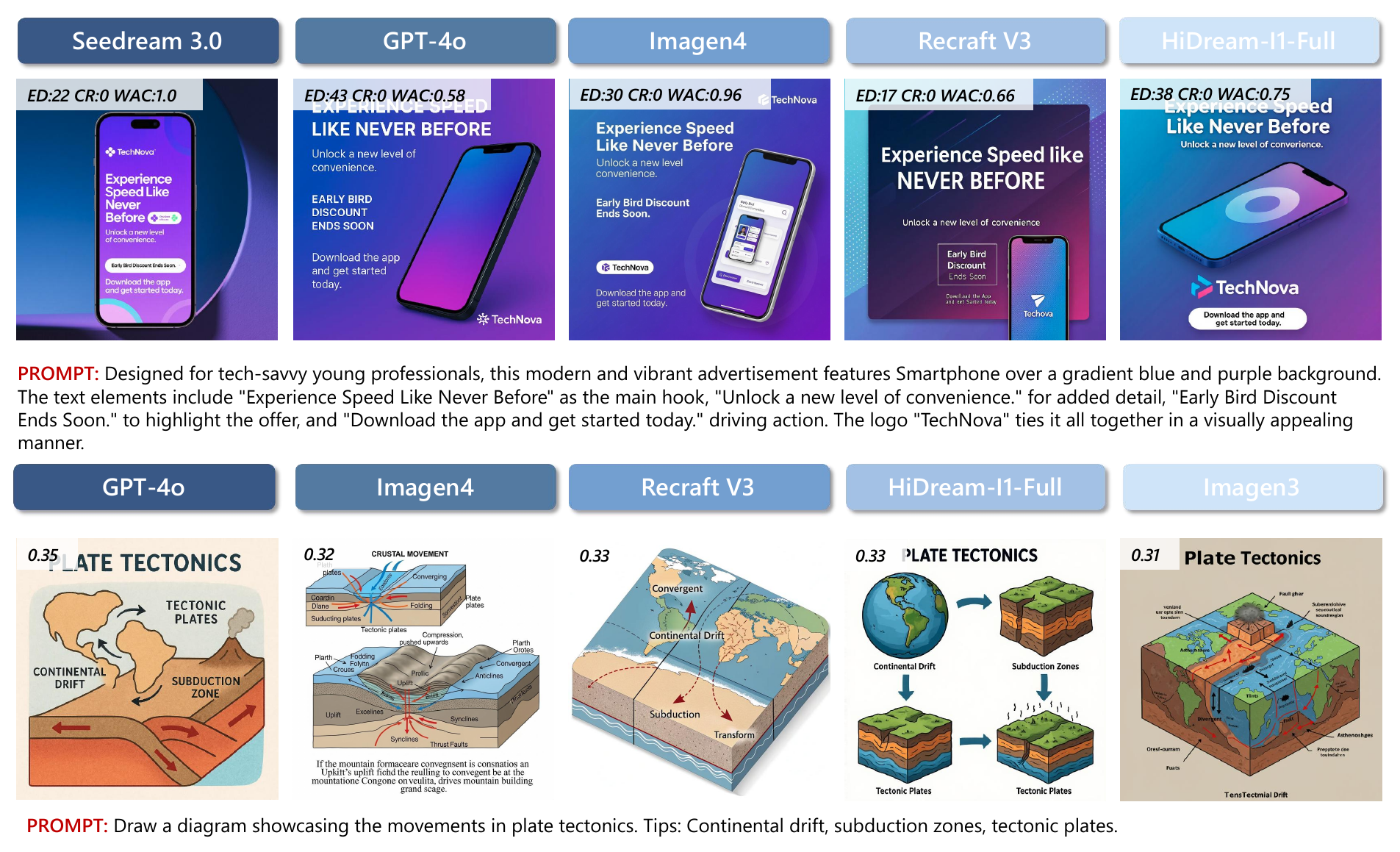}
    \caption{\small \textbf{An illustration of the generation results and the corresponding scores.} The first row shows the text rendering results and the second row shows the the knowledge and reasoning results. And the evaluation scores are displayed in the upper left corner of the image.}
    \label{fig:ranking}
    \vspace{-0.35cm}
\end{figure}

As shown in Figure~\ref{fig:ranking}, although GPT-4o~\cite{openai2024gpt4o_image} demonstrates strong visual accuracy, it shows no particular advantage in quantitative evaluation. This is mainly due to our strict evaluation criterion: any mismatch in capitalization (e.g., uppercase vs. lowercase) is counted as one unit of edit distance. This rule has a direct impact on GPT-4o’s overall rendering scores.

Compared to Imagen3~\cite{2024Imagen3}, Imagen4~\cite{2025Imagen4} shows a clear improvement in text rendering, with nearly double the performance on metrics such as ED and CR. Nonetheless, its overall visual appeal and the clarity of rendered text for long prompts still leave room for enhancement.

Notably, Recraft V3~\cite{2024recraftv3} consistently achieves excellent edit distance performance across all prompt lengths, with its values for long prompts showing a large gap from the second-best result. However, its performance in completion ratio and word accuracy count is relatively less impressive. This discrepancy may be attributed to its layout-first strategy~\cite{2024recraft_text}, in which a layout is generated prior to text insertion. This approach effectively reduces severe errors and prevents chaotic outputs by decomposing the original text rendering task into several relatively simpler subtasks, thereby significantly enhancing the reliability of text rendering in the final images. 

\subsubsection{Knowledge and Reasoning}

In the knowledge and reasoning dimension, shown in Table ~\ref{tab:reasoning} and in Figure~\ref{fig:ranking}, GPT-4o~\cite{openai2024gpt4o_image} demonstrates substantially stronger capabilities than other models. It consistently outperforms its counterparts in both knowledge retention and reasoning ability across nearly all subject categories evaluated. In general, closed-source models outperform open-source models in knowledge and reasoning capabilities. Notably, no single model has shown a particularly outstanding performance in specific subjects, indicating that the reasoning abilities of current models are largely derived from a balanced and conventional training dataset.

\begin{table}[ht]
\centering
\footnotesize
\caption{\small \textbf{The Knowledge and Reasoning evaluation results.} \legendsquare{colorfirst} \legendsquare{colorsecond} \legendsquare{colorthird} \legendsquare{colorfourth} \legendsquare{colorfifth} indicate the first, second, third, fourth, and fifth performance respectively.}
\vspace{0.2cm}
\scalebox{0.75}{
\begin{tabular}{l|ccccccc}
\toprule
\textbf{Method}   & \textbf{Geography}  & \textbf{Computer Science} & \textbf{Biology} & \textbf{Mathmatics} & \textbf{Physics} & \textbf{Chemistry} & \textbf{Common Sense}   \\
\midrule
Janus-Pro~\cite{chen2025janus} & 0.153  & 0.134   & 0.123 & 0.130  & 0.144 & 0.131 & 0.189 \\
BLIP3-o~\cite{chen2025blip3} & 0.210   & 0.225   & 0.219 & 0.223 & 0.219 & 0.232 & 0.275 \\
BAGEL~\cite{deng2025bagel} & 0.191  & 0.152   & 0.183 & 0.149 & 0.168 & 0.169 & 0.248 \\
BAGEL+CoT~\cite{deng2025bagel} & 0.206  & 0.184  & 0.206 & 0.200  & 0.212 & 0.208 & 0.273\\
Show-o2-1.5B~\cite{xie2025showo2} & 0.218 & 0.197 & 0.235 & 0.201 & 0.222 & 0.206 & 0.295 \\
Show-o2-7B~\cite{xie2025showo2} & 0.222 & 0.206 & 0.244 & 0.217 & 0.227 & 0.212 & 0.297 \\
OmniGen2~\cite{wu2025omnigen2} & 0.266 & 0.270 & 0.271 & 0.284 & 0.257 & 0.278 & 0.314 \\
\midrule
Stable Diffusion 1.5~\cite{Rombach_2022_CVPR} & 0.217 & 0.203 & 0.211 & 0.212 & 0.207 & 0.185 & 0.251 \\
Stable Diffusion XL~\cite{podell2023sdxl} & 0.246 & 0.216 & 0.244 & 0.235 & 0.234 & 0.239 & 0.289 \\
Stable Diffusion 3.5 Large~\cite{2024sd3.5} & 0.291 & 0.299 & 0.283 & \rankfifth{0.306} & 0.292 & 0.292 & 0.319 \\
Flux.1-dev~\cite{flux2024} & 0.239 & 0.257 & 0.247 & 0.265 & 0.247 & 0.253 & 0.298 \\
CogView4~\cite{2025cogview4} & 0.223 & 0.251 & 0.237 & 0.279 & 0.239 & 0.252 & 0.296 \\
SANA-1.5 1.6B(PAG)~\cite{xie2025sana} & 0.207 & 0.203 & 0.222 & 0.218 & 0.211 & 0.193 & 0.280 \\
SANA-1.5 4.8B(PAG)~\cite{xie2025sana} & 0.214 & 0.206 & 0.224 & 0.224 & 0.214 & 0.203 & 0.282 \\
Lumina-Image 2.0~\cite{lumina2} & 0.208 & 0.206 & 0.225 & 0.222 & 0.215 & 0.197 & 0.278 \\
HiDream-I1-Full~\cite{2025hidreami1} & \rankthird{0.324} & \rankfourth{0.324} & \rankfourth{0.305} & \rankfourth{0.312} & \rankfifth{0.318} & \rankfifth{0.305} & \rankthird{0.347} \\
\midrule
Imagen3~\cite{2024Imagen3} & \rankfifth{0.304} & \rankfifth{0.319} & 0.298 & 0.303 & \rankthird{0.320} & \rankfourth{0.315} & \rankfifth{0.338} \\
Recraft V3~\cite{2024recraftv3} & \rankfourth{0.323} & \rankthird{0.337} & \rankfifth{0.303} & \rankthird{0.320} & \rankfourth{0.319} & \rankthird{0.328} & \rankfourth{0.344} \\
Kolors 2.0~\cite{2025Kolors2} & 0.255 & 0.252 & 0.256 & 0.263 & 0.258 & 0.277 & 0.314 \\
Seedream 3.0~\cite{gao2025seedream} & 0.246 & 0.295 & \rankthird{0.313} & 0.253 & 0.297 & 0.270 & 0.277 \\
Imagen4~\cite{2025Imagen4} & \ranksecond{0.334}  & \ranksecond{0.346} & \ranksecond{0.314} & \rankfirst{0.343} & \ranksecond{0.342} & \ranksecond{0.346} & \ranksecond{0.351} \\
GPT-4o~\cite{openai2024gpt4o_image} & \rankfirst{0.351} & \rankfirst{0.348} & \rankfirst{0.323} & \ranksecond{0.334} & \rankfirst{0.350} & \rankfirst{0.355} & \rankfirst{0.364} \\

\bottomrule
\end{tabular}}
    \label{tab:reasoning}
    \vspace{-0.2cm}
\end{table}

The models' reasoning abilities can be categorized into five tiers as follows: the first tier includes GPT-4o~\cite{openai2024gpt4o_image}, Imagen4~\cite{2025Imagen4}, the second tier consists of Recraft V3~\cite{2024recraftv3}, HiDream-I1-Full~\cite{2025hidreami1}, and Imagen3~\cite{2024Imagen3}, the third tier includes only Stable Diffusion 3.5 Large~\cite{2024sd3.5}, and the fourth tier includes Seedream 3.0~\cite{gao2025seedream}, Lumina-Image 2.0~\cite{lumina2}, and Kolors 2~\cite{2025Kolors2}, other models form the last tier. The following figure visualizes the reasoning scores, which correspond closely with the aforementioned ranking.

\section{Conclusion}\label{sec:conclusion}

We introduce a comprehensive text-to-image benchmark, namely \textbf{OneIG-Bench}, which establishes a systematic framework for omni-dimensional nuanced evaluation through categorization of generation themes. 
Specifically, we have meticulously designed general scenarios including human figures and conventional objects, text rendering scenarios, and anime/style scenarios, and have crafted evaluation metrics for each scenario to comprehensively measure text-to-image performance. By decomposing evaluation into these discrete dimensions, the benchmark facilitates in-depth comparative analysis of models' strengths and limitations. This approach not only provides researchers with a rigorous evaluation framework but also serves as a guiding tool for identifying technical bottlenecks and prioritizing methodological innovations in the field.

\label{check:limitation}
\textbf{Limitation:} While this study presents a novel and systematic benchmark, several limitations should be acknowledged. (1) Knowledge and reasoning represents a relatively novel task in the image generation domain, and most existing models currently lack robust reasoning capabilities. While we have confirmed that our metric rankings align closely with human evaluations, there may exist more rational and effective evaluation approaches yet to be explored. (2) Moreover, aesthetic models tend to exhibit unexpected biases, while body quality assessment models often lack sufficient discriminative power and generalizability. We will further investigate both dimensions to develop a more robust and precise evaluation method.

\section*{Acknowledgements}\label{sec:acknowledgements}
\vspace{-0.1cm}
We are particularly grateful to Bizhu Huang, Shuli Gao, Kang An, and
Wen Sun for their invaluable support throughout the course of this research.

\newpage

{
    \clearpage
    \small
    \bibliographystyle{plain}
    \bibliography{main}

\begin{thebibliography}{10}

\bibitem{antol2015vqa}
Stanislaw Antol, Aishwarya Agrawal, Jiasen Lu, Margaret Mitchell, Dhruv Batra, C~Lawrence Zitnick, and Devi Parikh.
\newblock Vqa: Visual question answering.
\newblock In {\em Proceedings of the IEEE international conference on computer vision}, pages 2425--2433, 2015.

\bibitem{Qwen}
Jinze Bai, Shuai Bai, Shusheng Yang, Shijie Wang, Sinan Tan, Peng Wang, Junyang Lin, Chang Zhou, and Jingren Zhou.
\newblock Qwen-vl: A versatile vision-language model for understanding, localization, text reading, and beyond.
\newblock {\em arXiv preprint arXiv:2308.12966}, 2023.

\bibitem{bai2025qwen2}
Shuai Bai, Keqin Chen, Xuejing Liu, Jialin Wang, Wenbin Ge, Sibo Song, Kai Dang, Peng Wang, Shijie Wang, Jun Tang, et~al.
\newblock Qwen2. 5-vl technical report.
\newblock {\em arXiv preprint arXiv:2502.13923}, 2025.

\bibitem{Qwen2.5}
Shuai Bai, Keqin Chen, Xuejing Liu, Jialin Wang, Wenbin Ge, Sibo Song, Kai Dang, Peng Wang, Shijie Wang, Jun Tang, Humen Zhong, Yuanzhi Zhu, Mingkun Yang, Zhaohai Li, Jianqiang Wan, Pengfei Wang, Wei Ding, Zheren Fu, Yiheng Xu, Jiabo Ye, Xi~Zhang, Tianbao Xie, Zesen Cheng, Hang Zhang, Zhibo Yang, Haiyang Xu, and Junyang Lin.
\newblock Qwen2.5-vl technical report.
\newblock {\em arXiv preprint arXiv:2502.13923}, 2025.

\bibitem{bakr2023hrs}
Eslam~Mohamed Bakr, Pengzhan Sun, Xiaoqian Shen, Faizan~Farooq Khan, Li~Erran Li, and Mohamed Elhoseiny.
\newblock Hrs-bench: Holistic, reliable and scalable benchmark for text-to-image models.
\newblock In {\em Proceedings of the IEEE/CVF International Conference on Computer Vision}, pages 20041--20053, 2023.

\bibitem{2024flux_api}
black-forest labs.
\newblock The official api of flux-1.dev.
\newblock \url{https://api.us1.bfl.ai/scalar#tag/tasks/POST/v1/flux-dev}, 2024.

\bibitem{chefer2023attend}
Hila Chefer, Yuval Alaluf, Yael Vinker, Lior Wolf, and Daniel Cohen-Or.
\newblock Attend-and-excite: Attention-based semantic guidance for text-to-image diffusion models.
\newblock {\em ACM transactions on Graphics (TOG)}, 42(4):1--10, 2023.

\bibitem{chen2025blip3}
Jiuhai Chen, Zhiyang Xu, Xichen Pan, Yushi Hu, Can Qin, Tom Goldstein, Lifu Huang, Tianyi Zhou, Saining Xie, Silvio Savarese, et~al.
\newblock Blip3-o: A family of fully open unified multimodal models-architecture, training and dataset.
\newblock {\em arXiv preprint arXiv:2505.09568}, 2025.

\bibitem{chen2024pixart}
Junsong Chen, Chongjian Ge, Enze Xie, Yue Wu, Lewei Yao, Xiaozhe Ren, Zhongdao Wang, Ping Luo, Huchuan Lu, and Zhenguo Li.
\newblock Pixart-sigma: Weak-to-strong training of diffusion transformer for 4k text-to-image generation.
\newblock In {\em European Conference on Computer Vision}, pages 74--91. Springer, 2024.

\bibitem{chen2023pixart}
Junsong Chen, Jincheng Yu, Chongjian Ge, Lewei Yao, Enze Xie, Yue Wu, Zhongdao Wang, James Kwok, Ping Luo, Huchuan Lu, et~al.
\newblock Pixart-alpha: Fast training of diffusion transformer for photorealistic text-to-image synthesis.
\newblock {\em arXiv preprint arXiv:2310.00426}, 2023.

\bibitem{chen2019mmdetection}
Kai Chen, Jiaqi Wang, Jiangmiao Pang, Yuhang Cao, Yu~Xiong, Xiaoxiao Li, Shuyang Sun, Wansen Feng, Ziwei Liu, Jiarui Xu, et~al.
\newblock Mmdetection: Open mmlab detection toolbox and benchmark.
\newblock {\em arXiv preprint arXiv:1906.07155}, 2019.

\bibitem{chen2020generative}
Mark Chen, Alec Radford, Rewon Child, Jeffrey Wu, Heewoo Jun, David Luan, and Ilya Sutskever.
\newblock Generative pretraining from pixels.
\newblock In {\em International conference on machine learning}, pages 1691--1703. PMLR, 2020.

\bibitem{chen2025janus}
Xiaokang Chen, Zhiyu Wu, Xingchao Liu, Zizheng Pan, Wen Liu, Zhenda Xie, Xingkai Yu, and Chong Ruan.
\newblock Janus-pro: Unified multimodal understanding and generation with data and model scaling.
\newblock {\em arXiv preprint arXiv:2501.17811}, 2025.

\bibitem{chen2024internvl}
Zhe Chen, Jiannan Wu, Wenhai Wang, Weijie Su, Guo Chen, Sen Xing, Muyan Zhong, Qinglong Zhang, Xizhou Zhu, Lewei Lu, et~al.
\newblock Internvl: Scaling up vision foundation models and aligning for generic visual-linguistic tasks.
\newblock In {\em Proceedings of the IEEE/CVF conference on computer vision and pattern recognition}, pages 24185--24198, 2024.

\bibitem{cheng2021mask2former}
Bowen Cheng, Anwesa Choudhuri, Ishan Misra, Alexander Kirillov, Rohit Girdhar, and Alexander~G Schwing.
\newblock Mask2former for video instance segmentation.
\newblock {\em arXiv preprint arXiv:2112.10764}, 2021.

\bibitem{Cho2024DSG}
Jaemin Cho, Yushi Hu, Jason Baldridge, Roopal Garg, Peter Anderson, Ranjay Krishna, Mohit Bansal, Jordi Pont-Tuset, and Su~Wang.
\newblock Davidsonian scene graph: Improving reliability in fine-grained evaluation for text-to-image generation.
\newblock In {\em ICLR}, 2024.

\bibitem{2025Imagen4}
Google deepmind Imagen4~team.
\newblock Imagen4.
\newblock \url{https://storage.googleapis.com/deepmind-media/Model-Cards/Imagen-4-Model-Card.pdf}, 2025.

\bibitem{deng2025bagel}
Chaorui Deng, Deyao Zhu, Kunchang Li, Chenhui Gou, Feng Li, Zeyu Wang, Shu Zhong, Weihao Yu, Xiaonan Nie, Ziang Song, Guang Shi, and Haoqi Fan.
\newblock Emerging properties in unified multimodal pretraining.
\newblock {\em arXiv preprint arXiv:2505.14683}, 2025.

\bibitem{esser2024scaling}
Patrick Esser, Sumith Kulal, Andreas Blattmann, Rahim Entezari, Jonas M{\"u}ller, Harry Saini, Yam Levi, Dominik Lorenz, Axel Sauer, Frederic Boesel, et~al.
\newblock Scaling rectified flow transformers for high-resolution image synthesis.
\newblock In {\em Forty-first international conference on machine learning}, 2024.

\bibitem{feng2022training}
Weixi Feng, Xuehai He, Tsu-Jui Fu, Varun Jampani, Arjun Akula, Pradyumna Narayana, Sugato Basu, Xin~Eric Wang, and William~Yang Wang.
\newblock Training-free structured diffusion guidance for compositional text-to-image synthesis.
\newblock {\em arXiv preprint arXiv:2212.05032}, 2022.

\bibitem{fu2023dreamsim}
Stephanie Fu, Netanel Tamir, Shobhita Sundaram, Lucy Chai, Richard Zhang, Tali Dekel, and Phillip Isola.
\newblock Dreamsim: Learning new dimensions of human visual similarity using synthetic data, 2023.

\bibitem{fu2024commonsenseT2I}
Xingyu Fu, Muyu He, Yujie Lu, William~Yang Wang, and Dan Roth.
\newblock Commonsense-t2i challenge: Can text-to-image generation models understand commonsense?
\newblock {\em arXiv preprint arXiv:2406.07546}, 2024.

\bibitem{gandikota2025distilling}
Rohit Gandikota and David Bau.
\newblock Distilling diversity and control in diffusion models.
\newblock {\em arXiv preprint arXiv:2503.10637}, 2025.

\bibitem{gao2025seedream}
Yu~Gao, Lixue Gong, Qiushan Guo, Xiaoxia Hou, Zhichao Lai, Fanshi Li, Liang Li, Xiaochen Lian, Chao Liao, Liyang Liu, et~al.
\newblock Seedream 3.0 technical report.
\newblock {\em arXiv preprint arXiv:2504.11346}, 2025.

\bibitem{ghosh2023geneval}
Dhruba Ghosh, Hannaneh Hajishirzi, and Ludwig Schmidt.
\newblock Geneval: An object-focused framework for evaluating text-to-image alignment.
\newblock {\em Advances in Neural Information Processing Systems}, 36:52132--52152, 2023.

\bibitem{goodfellow2020generative}
Ian Goodfellow, Jean Pouget-Abadie, Mehdi Mirza, Bing Xu, David Warde-Farley, Sherjil Ozair, Aaron Courville, and Yoshua Bengio.
\newblock Generative adversarial networks.
\newblock {\em Communications of the ACM}, 63(11):139--144, 2020.

\bibitem{han2024evalmuse40kreliablefinegrainedbenchmark}
Shuhao Han, Haotian Fan, Jiachen Fu, Liang Li, Tao Li, Junhui Cui, Yunqiu Wang, Yang Tai, Jingwei Sun, Chunle Guo, and Chongyi Li.
\newblock Evalmuse-40k: A reliable and fine-grained benchmark with comprehensive human annotations for text-to-image generation model evaluation, 2024.

\bibitem{heusel2017gans}
Martin Heusel, Hubert Ramsauer, Thomas Unterthiner, Bernhard Nessler, and Sepp Hochreiter.
\newblock Gans trained by a two time-scale update rule converge to a local nash equilibrium.
\newblock {\em Advances in neural information processing systems}, 30, 2017.

\bibitem{2025hidreami1}
HiDream-ai.
\newblock Hidream-i1.
\newblock \url{https://github.com/HiDream-ai/HiDream-I1}, 2025.

\bibitem{ho2020denoising}
Jonathan Ho, Ajay Jain, and Pieter Abbeel.
\newblock Denoising diffusion probabilistic models.
\newblock {\em Advances in neural information processing systems}, 33:6840--6851, 2020.

\bibitem{hu2024ella}
Xiwei Hu, Rui Wang, Yixiao Fang, Bin Fu, Pei Cheng, and Gang Yu.
\newblock Ella: Equip diffusion models with llm for enhanced semantic alignment.
\newblock {\em arXiv preprint arXiv:2403.05135}, 2024.

\bibitem{hu2023tifa}
Yushi Hu, Benlin Liu, Jungo Kasai, Yizhong Wang, Mari Ostendorf, Ranjay Krishna, and Noah~A Smith.
\newblock Tifa: Accurate and interpretable text-to-image faithfulness evaluation with question answering.
\newblock In {\em Proceedings of the IEEE/CVF International Conference on Computer Vision}, pages 20406--20417, 2023.

\bibitem{huang2025t2i}
Kaiyi Huang, Chengqi Duan, Kaiyue Sun, Enze Xie, Zhenguo Li, and Xihui Liu.
\newblock T2i-compbench++: An enhanced and comprehensive benchmark for compositional text-to-image generation.
\newblock {\em IEEE Transactions on Pattern Analysis and Machine Intelligence}, 2025.

\bibitem{huang2023t2icompbench}
Kaiyi Huang, Kaiyue Sun, Enze Xie, Zhenguo Li, and Xihui Liu.
\newblock T2i-compbench: A comprehensive benchmark for open-world compositional text-to-image generation.
\newblock {\em Advances in Neural Information Processing Systems}, 36:78723--78747, 2023.

\bibitem{huang2024llm2clip}
Weiquan Huang, Aoqi Wu, Yifan Yang, Xufang Luo, Yuqing Yang, Liang Hu, Qi~Dai, Xiyang Dai, Dongdong Chen, Chong Luo, et~al.
\newblock Llm2clip: Powerful language model unlock richer visual representation.
\newblock {\em arXiv preprint arXiv:2411.04997}, 2024.

\bibitem{2024Imagen3}
Imagen-Team-Google.
\newblock Imagen 3, 2024.

\bibitem{kirstain2023pick}
Yuval Kirstain, Adam Polyak, Uriel Singer, Shahbuland Matiana, Joe Penna, and Omer Levy.
\newblock Pick-a-pic: An open dataset of user preferences for text-to-image generation.
\newblock {\em Advances in Neural Information Processing Systems}, 36:36652--36663, 2023.

\bibitem{flux2024}
Black~Forest Labs.
\newblock Flux.
\newblock \url{https://github.com/black-forest-labs/flux}, 2024.

\bibitem{li2024genai}
Baiqi Li, Zhiqiu Lin, Deepak Pathak, Jiayao Li, Yixin Fei, Kewen Wu, Tiffany Ling, Xide Xia, Pengchuan Zhang, Graham Neubig, et~al.
\newblock Genai-bench: Evaluating and improving compositional text-to-visual generation.
\newblock {\em arXiv preprint arXiv:2406.13743}, 2024.

\bibitem{li2024playground}
Daiqing Li, Aleks Kamko, Ehsan Akhgari, Ali Sabet, Linmiao Xu, and Suhail Doshi.
\newblock Playground v2.5: Three insights towards enhancing aesthetic quality in text-to-image generation, 2024.

\bibitem{li2022blip}
Junnan Li, Dongxu Li, Caiming Xiong, and Steven Hoi.
\newblock Blip: Bootstrapping language-image pre-training for unified vision-language understanding and generation.
\newblock In {\em International conference on machine learning}, pages 12888--12900. PMLR, 2022.

\bibitem{ma2024sit}
Nanye Ma, Mark Goldstein, Michael~S Albergo, Nicholas~M Boffi, Eric Vanden-Eijnden, and Saining Xie.
\newblock Sit: Exploring flow and diffusion-based generative models with scalable interpolant transformers.
\newblock In {\em European Conference on Computer Vision}, pages 23--40. Springer, 2024.

\bibitem{niu2025wise}
Yuwei Niu, Munan Ning, Mengren Zheng, Bin Lin, Peng Jin, Jiaqi Liao, Kunpeng Ning, Bin Zhu, and Li~Yuan.
\newblock Wise: A world knowledge-informed semantic evaluation for text-to-image generation.
\newblock {\em arXiv preprint arXiv:2503.07265}, 2025.

\bibitem{openai2024gpt4o}
OpenAI.
\newblock Gpt-4o system card.
\newblock \url{https://openai.com/index/gpt-4o-system-card/}, 2024.

\bibitem{openai2024gpt4o_image}
OpenAI.
\newblock Introducing 4o image generation.
\newblock \url{https://openai.com/index/introducing-4o-image-generation/}, 2025.

\bibitem{podell2023sdxl}
Dustin Podell, Zion English, Kyle Lacey, Andreas Blattmann, Tim Dockhorn, Jonas M{\"u}ller, Joe Penna, and Robin Rombach.
\newblock Sdxl: Improving latent diffusion models for high-resolution image synthesis.
\newblock {\em arXiv preprint arXiv:2307.01952}, 2023.

\bibitem{lumina2}
Qi~Qin, Le~Zhuo, Yi~Xin, Ruoyi Du, Zhen Li, Bin Fu, Yiting Lu, Xinyue Li, Dongyang Liu, Xiangyang Zhu, Will Beddow, Erwann Millon, Wenhai~Wang Victor~Perez, Yu~Qiao, Bo~Zhang, Xiaohong Liu, Hongsheng Li, Chang Xu, and Peng Gao.
\newblock Lumina-image 2.0: A unified and efficient image generative framework, 2025.

\bibitem{qiu2025phybench}
Shi Qiu, Shaoyang Guo, Zhuo-Yang Song, Yunbo Sun, Zeyu Cai, Jiashen Wei, Tianyu Luo, Yixuan Yin, Haoxu Zhang, Yi~Hu, et~al.
\newblock Phybench: Holistic evaluation of physical perception and reasoning in large language models.
\newblock {\em arXiv preprint arXiv:2504.16074}, 2025.

\bibitem{radford2021learning}
Alec Radford, Jong~Wook Kim, Chris Hallacy, Aditya Ramesh, Gabriel Goh, Sandhini Agarwal, Girish Sastry, Amanda Askell, Pamela Mishkin, Jack Clark, et~al.
\newblock Learning transferable visual models from natural language supervision.
\newblock In {\em International conference on machine learning}, pages 8748--8763. PmLR, 2021.

\bibitem{radford2021clip}
Alec Radford, Jong~Wook Kim, Chris Hallacy, Aditya Ramesh, Gabriel Goh, Sandhini Agarwal, Girish Sastry, Amanda Askell, Pamela Mishkin, Jack Clark, et~al.
\newblock Learning transferable visual models from natural language supervision.
\newblock In {\em International conference on machine learning}, pages 8748--8763. PmLR, 2021.

\bibitem{radford2019language}
Alec Radford, Jeffrey Wu, Rewon Child, David Luan, Dario Amodei, Ilya Sutskever, et~al.
\newblock Language models are unsupervised multitask learners.
\newblock {\em OpenAI blog}, 1(8):9, 2019.

\bibitem{raffel2020exploring}
Colin Raffel, Noam Shazeer, Adam Roberts, Katherine Lee, Sharan Narang, Michael Matena, Yanqi Zhou, Wei Li, and Peter~J Liu.
\newblock Exploring the limits of transfer learning with a unified text-to-text transformer.
\newblock {\em Journal of machine learning research}, 21(140):1--67, 2020.

\bibitem{ramesh2022hierarchical}
Aditya Ramesh, Prafulla Dhariwal, Alex Nichol, Casey Chu, and Mark Chen.
\newblock Hierarchical text-conditional image generation with clip latents.
\newblock {\em arXiv preprint arXiv:2204.06125}, 1(2):3, 2022.

\bibitem{ramesh2021zero}
Aditya Ramesh, Mikhail Pavlov, Gabriel Goh, Scott Gray, Chelsea Voss, Alec Radford, Mark Chen, and Ilya Sutskever.
\newblock Zero-shot text-to-image generation.
\newblock In {\em International conference on machine learning}, pages 8821--8831. Pmlr, 2021.

\bibitem{Rombach_2022_CVPR}
Robin Rombach, Andreas Blattmann, Dominik Lorenz, Patrick Esser, and Bj\"orn Ommer.
\newblock High-resolution image synthesis with latent diffusion models.
\newblock In {\em Proceedings of the IEEE/CVF Conference on Computer Vision and Pattern Recognition (CVPR)}, pages 10684--10695, June 2022.

\bibitem{saharia2022photorealistic}
Chitwan Saharia, William Chan, Saurabh Saxena, Lala Li, Jay Whang, Emily~L Denton, Kamyar Ghasemipour, Raphael Gontijo~Lopes, Burcu Karagol~Ayan, Tim Salimans, et~al.
\newblock Photorealistic text-to-image diffusion models with deep language understanding.
\newblock {\em Advances in neural information processing systems}, 35:36479--36494, 2022.

\bibitem{sauer2023stylegan}
Axel Sauer, Tero Karras, Samuli Laine, Andreas Geiger, and Timo Aila.
\newblock Stylegan-t: Unlocking the power of gans for fast large-scale text-to-image synthesis.
\newblock In {\em International conference on machine learning}, pages 30105--30118. PMLR, 2023.

\bibitem{somepalli2024measuring}
Gowthami Somepalli, Anubhav Gupta, Kamal Gupta, Shramay Palta, Micah Goldblum, Jonas Geiping, Abhinav Shrivastava, and Tom Goldstein.
\newblock Measuring style similarity in diffusion models.
\newblock {\em arXiv preprint arXiv:2404.01292}, 2024.

\bibitem{song2020denoising}
Jiaming Song, Chenlin Meng, and Stefano Ermon.
\newblock Denoising diffusion implicit models.
\newblock {\em arXiv preprint arXiv:2010.02502}, 2020.

\bibitem{2024sd3.5}
Stability-AI.
\newblock stable-diffusion-3.5-large.
\newblock \url{https://github.com/Stability-AI/sd3.5}, 2024.

\bibitem{sun2024autoregressive}
Peize Sun, Yi~Jiang, Shoufa Chen, Shilong Zhang, Bingyue Peng, Ping Luo, and Zehuan Yuan.
\newblock Autoregressive model beats diffusion: Llama for scalable image generation.
\newblock {\em arXiv preprint arXiv:2406.06525}, 2024.

\bibitem{tan2024evalalign}
Zhiyu Tan, Xiaomeng Yang, Luozheng Qin, Mengping Yang, Cheng Zhang, and Hao Li.
\newblock Evalalign: Evaluating text-to-image models through precision alignment of multimodal large models with supervised fine-tuning to human annotations.
\newblock {\em arXiv e-prints}, pages arXiv--2406, 2024.

\bibitem{team2024chameleon}
Chameleon Team.
\newblock Chameleon: Mixed-modal early-fusion foundation models.
\newblock {\em arXiv preprint arXiv:2405.09818}, 2024.

\bibitem{2025Kolors2}
Kuaishou~Kolors team.
\newblock Kolors2.0.
\newblock \url{https://app.klingai.com/cn/}, 2025.

\bibitem{2024recraft_text}
Recraft team.
\newblock How to create sota image generation with text recrafts ml team insights.
\newblock \url{https://www.recraft.ai/blog/how-to-create-sota-image-generation-with-text-recrafts-ml-team-insights}, 2024.

\bibitem{2024recraftv3}
Recraft team.
\newblock Recraft v3.
\newblock \url{https://www.recraft.ai/blog/recraft-introduces-a-revolutionary-ai-model-that-thinks-in-design-language?utm_source=ai-bot.cn}, 2024.

\bibitem{wang2024qwen2}
Peng Wang, Shuai Bai, Sinan Tan, Shijie Wang, Zhihao Fan, Jinze Bai, Keqin Chen, Xuejing Liu, Jialin Wang, Wenbin Ge, et~al.
\newblock Qwen2-vl: Enhancing vision-language model's perception of the world at any resolution.
\newblock {\em arXiv preprint arXiv:2409.12191}, 2024.

\bibitem{wang2024emu3}
Xinlong Wang, Xiaosong Zhang, Zhengxiong Luo, Quan Sun, Yufeng Cui, Jinsheng Wang, Fan Zhang, Yueze Wang, Zhen Li, Qiying Yu, et~al.
\newblock Emu3: Next-token prediction is all you need.
\newblock {\em arXiv preprint arXiv:2409.18869}, 2024.

\bibitem{wang2004image}
Zhou Wang, Alan~C Bovik, Hamid~R Sheikh, and Eero~P Simoncelli.
\newblock Image quality assessment: from error visibility to structural similarity.
\newblock {\em IEEE transactions on image processing}, 13(4):600--612, 2004.

\bibitem{wiles2024revisiting}
Olivia Wiles, Chuhan Zhang, Isabela Albuquerque, Ivana Kaji{\'c}, Su~Wang, Emanuele Bugliarello, Yasumasa Onoe, Pinelopi Papalampidi, Ira Ktena, Chris Knutsen, et~al.
\newblock Revisiting text-to-image evaluation with gecko: On metrics, prompts, and human ratings.
\newblock {\em arXiv preprint arXiv:2404.16820}, 2024.

\bibitem{wu2025omnigen2}
Chenyuan Wu, Pengfei Zheng, Ruiran Yan, Shitao Xiao, Xin Luo, Yueze Wang, Wanli Li, Xiyan Jiang, Yexin Liu, Junjie Zhou, Ze~Liu, Ziyi Xia, Chaofan Li, Haoge Deng, Jiahao Wang, Kun Luo, Bo~Zhang, Defu Lian, Xinlong Wang, Zhongyuan Wang, Tiejun Huang, and Zheng Liu.
\newblock Omnigen2: Exploration to advanced multimodal generation.
\newblock {\em arXiv preprint arXiv:2506.18871}, 2025.

\bibitem{wu2023human}
Xiaoshi Wu, Yiming Hao, Keqiang Sun, Yixiong Chen, Feng Zhu, Rui Zhao, and Hongsheng Li.
\newblock Human preference score v2: A solid benchmark for evaluating human preferences of text-to-image synthesis.
\newblock {\em arXiv preprint arXiv:2306.09341}, 2023.

\bibitem{xie2025sana}
Enze Xie, Junsong Chen, Yuyang Zhao, Jincheng Yu, Ligeng Zhu, Yujun Lin, Zhekai Zhang, Muyang Li, Junyu Chen, Han Cai, et~al.
\newblock Sana 1.5: Efficient scaling of training-time and inference-time compute in linear diffusion transformer, 2025.

\bibitem{xie2024show}
Jinheng Xie, Weijia Mao, Zechen Bai, David~Junhao Zhang, Weihao Wang, Kevin~Qinghong Lin, Yuchao Gu, Zhijie Chen, Zhenheng Yang, and Mike~Zheng Shou.
\newblock Show-o: One single transformer to unify multimodal understanding and generation.
\newblock {\em arXiv preprint arXiv:2408.12528}, 2024.

\bibitem{xie2025showo2}
Jinheng Xie, Zhenheng Yang, and Mike~Zheng Shou.
\newblock Show-o2: Improved native unified multimodal models.
\newblock {\em arXiv preprint arXiv:2506.15564}, 2025.

\bibitem{xing2024csgo}
Peng Xing, Haofan Wang, Yanpeng Sun, Qixun Wang, Xu~Bai, Hao Ai, Renyuan Huang, and Zechao Li.
\newblock Csgo: Content-style composition in text-to-image generation.
\newblock {\em arXiv preprint arXiv:2408.16766}, 2024.

\bibitem{yu2022scaling}
Jiahui Yu, Yuanzhong Xu, Jing~Yu Koh, Thang Luong, Gunjan Baid, Zirui Wang, Vijay Vasudevan, Alexander Ku, Yinfei Yang, Burcu~Karagol Ayan, et~al.
\newblock Scaling autoregressive models for content-rich text-to-image generation.
\newblock {\em arXiv preprint arXiv:2206.10789}, 2(3):5, 2022.

\bibitem{2025cogview4}
THUKEG Z.ai.
\newblock Cogview4.
\newblock \url{https://github.com/THUDM/CogView4}, 2025.

\bibitem{zhai2023siglip}
Xiaohua Zhai, Basil Mustafa, Alexander Kolesnikov, and Lucas Beyer.
\newblock Sigmoid loss for language image pre-training.
\newblock In {\em Proceedings of the IEEE/CVF international conference on computer vision}, pages 11975--11986, 2023.

\bibitem{zhang2025worldgenbench}
Daoan Zhang, Che Jiang, Ruoshi Xu, Biaoxiang Chen, Zijian Jin, Yutian Lu, Jianguo Zhang, Liang Yong, Jiebo Luo, and Shengda Luo.
\newblock Worldgenbench: A world-knowledge-integrated benchmark for reasoning-driven text-to-image generation, 2025.

\end{thebibliography}
}

\newpage

\appendix
\section{Appendix}
\subsection{Word Count Distribution Statistics}
\begin{figure}[h]
    \centering
    \begin{minipage}{0.5\linewidth}
        \centering
        \includegraphics[width=\linewidth]{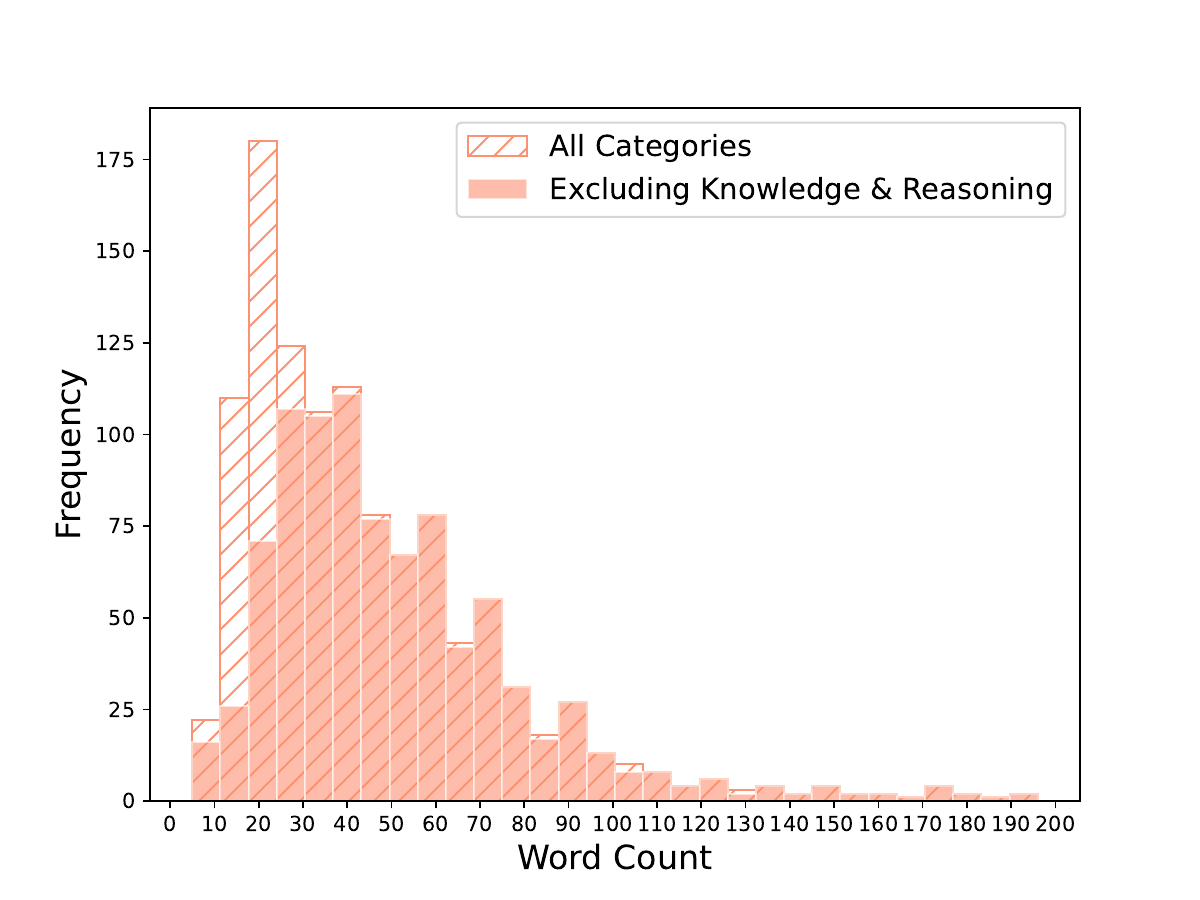}
        \caption{\small \textbf{Word Count of the Overall Prompts of OneIG-Bench}. The word count distribution of OneIG-Bench's prompts ranges from 0 to 200.}
        \label{fig:word_cnt_distribute}
    \end{minipage}
    \hfill
    \begin{minipage}{0.47\linewidth}
        \centering
        \includegraphics[width=\linewidth]{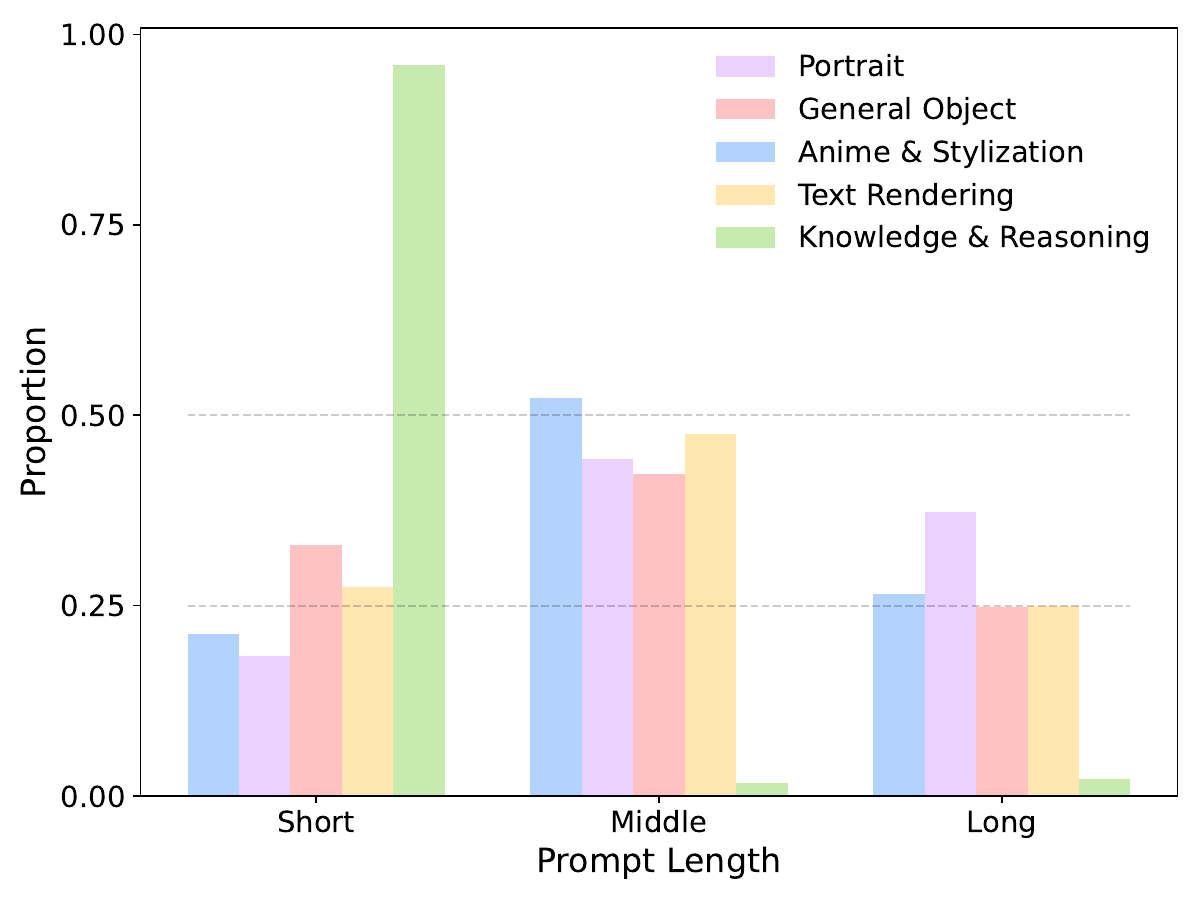}
        \caption{\small \textbf{The distribution of prompt word counts across \textbf{Short}, \text{Middle}, \text{Long} categories}.}
        \label{fig:categories}
    \end{minipage}
\end{figure}

The word count distribution of our OneIG-Bench prompts, as shown in Table~\ref{tab:distribution} and Figure~\ref{fig:word_cnt_distribute}, follows a \textbf{Short} : \textbf{Middle} : \textbf{Long} ratio of approximately 1:2:1. The choice of 30 and 60 words as the thresholds for distinguishing short, middle, and long prompts is based on the following reasoning: According to common rules for understanding token lengths, 1 word is approximately equal to 4/3 tokens, and 1-2 sentences are roughly equivalent to 30 tokens. This means that a simple 1-2 sentence prompt has a length of about 20-25 words. To ensure diversity in sentence structure and accuracy in stylization or portraiture in the image, we set the boundaries for short and middle prompts at 30 words. Furthermore, since some text encoders, such as CLIP~\cite{radford2021clip}, SigLIP~\cite{zhai2023siglip}, support a maximum of 77 tokens, a prompt of up to 60 words can generally be processed directly by these encoders.

\begin{table}[ht]
\centering
\caption{\small \textbf{Word count distribution of OneIG-Bench prompts in different categories}. In the table, \textbf{Avg} represents the average word count of prompts in different categories (including total and total w/o Knowledge \& Reasoning). \textbf{Short}, \textbf{Medium} and \textbf{Long} represent the length of the prompts, where \textbf{Short} denote the number of words is less than 30, \textbf{Medium} denotes the number between 30 and 60, and \textbf{Long} denotes the number exceeding 60. "K \& R" is the abbreviation for "Knowledge \& Reasoning".}
\setlength{\tabcolsep}{10pt} 
\renewcommand{\arraystretch}{1} 
\begin{tabular}{lcccc}
\toprule
\textbf{Category} & \textbf{Avg} & \textbf{Short} & \textbf{Middle} & \textbf{Long} \\
\midrule
Portrait & 56.4 & 0.184 & 0.443 & 0.373 \\
General Object & 46.5 & 0.330 & 0.422 & 0.248 \\
Anime \& Stylization & 50.6 & 0.212 & 0.522 & 0.265 \\
Text Rendering & 51.2 & 0.275 & 0.475 & 0.250 \\
Knowledge \& Reasoning & 20.5 &  0.960 & 0.018 & 0.022 \\
Total Distribution & 45.2 & 0.389 & 0.377 & 0.234 \\
Total Distribution w/o K \& R & 51.3 & 0.246 & 0.467 & 0.287 \\
\bottomrule
\end{tabular}
\label{tab:distribution}
\end{table}

The Portrait category, however, shows a slight deviation from this 1:2:1 distribution in Figure~\ref{fig:categories} due to the explicit requirement for portraits in the prompts, ensuring that the generated characters do not include stylized figures like those found in anime. As a result, the average word count for prompts in this category is higher than in other categories. On the other hand, the Knowledge \& Reasoning category, which focuses on reasoning tasks, does not revise the prompts to conform to the word count ratio, leading to a noticeably lower average word count compared to other categories. In general, excluding the Knowledge \& Reasoning category, OneIG-Bench prompts' word count results align closely with the 1:2:1 ratio.

\subsection{Implementation}
Our experiments on image generation with unified multimodal methods and open-source methods are configured according to Table~\ref{tab:config1}. For all methods, the CFG and step parameters follow the methods' default settings. To ensure consistency and ensure the quality of image generation, we increased the default steps for Stable Diffusion 3.5 Large~\cite{2024sd3.5} from 40 to 50. The number of inference steps for Flux.1-dev~\cite{flux2024} is set to be consistent with that used in the official API~\cite{2024flux_api}. With the exception of Stable Diffusion 1.5~\cite{Rombach_2022_CVPR}, which cannot generate images with a resolution of 1024 $\times$ 1024, all other methods are configured to generate images at a resolution of 1024 $\times$ 1024.

\begin{table}[h]
\vspace{-0.2cm}
    \centering
    \caption{\small \textbf{Configurations for unified multimodal and open-source methods}. \textbf{Size} represents the parameter size of the corresponding method. \textbf{CFG} represents the \textbf{guidance scale} of the corresponding method. \textbf{Resolution} represents the resolution of the image generated by the corresponding method. \textbf{Step} represents the number of inference steps during the image generation process.}
    \vspace{0.2cm}
     \resizebox{0.7\columnwidth}{!}{
    \begin{tabular}{lcccc}
        \toprule
        \textbf{Method} & \textbf{Size} & \textbf{CFG} & \textbf{Resolution} & \textbf{Step} \\
        \midrule
Janus-Pro~\cite{chen2025janus} & 7B & 5.0 & 384 $\times$ 384 & - \\
BLIP3-o~\cite{chen2025blip3} & 8B & 3.0 & 1024 $\times$ 1024 & - \\
BAGEL~\cite{deng2025bagel} & 7B & 4.0 & 1024 $\times$ 1024 & 50\\
BAGEL+CoT~\cite{deng2025bagel} & 7B & 4.0 & 1024 $\times$ 1024 & 50\\
Show-o2-1.5B~\cite{xie2025showo2} & 1.5B & 5.0 & 432 $\times$ 432 & 50 \\
Show-o2-7B~\cite{xie2025showo2} & 7B & 5.0 & 432 $\times$ 432 & 50 \\
OmniGen2~\cite{wu2025omnigen2} & 3B+4B & 4.0 & 1024 $\times$ 1024 & 50 \\
        \midrule
Stable Diffusion 1.5~\cite{Rombach_2022_CVPR} & 0.9B                              & 7.5                              & 512 $\times$ 512                              & 50                                \\
Stable Diffusion XL~\cite{podell2023sdxl}       & 2.6B                              & 5.0                              & 1024 $\times$ 1024                             & 50                                \\
Stable Diffusion 3.5 Large~\cite{2024sd3.5}     & 8.1B                              & 4.5                               & 1024 $\times$ 1024                             & 50                                \\
Flux.1-dev~\cite{flux2024}                      & 12B                               & 3.5                               & 1024 $\times$ 1024                             & 28                                \\
CogView4~\cite{2025cogview4}                    & 6B                                & 3.5                               & 1024 $\times$ 1024                             & 50                                \\
SANA-1.5 1.6B (PAG)~\cite{xie2025sana}          & 1.6B                              & 5.0                               & 1024 $\times$ 1024                             & 20                                \\
SANA-1.5 4.8B (PAG)~\cite{xie2025sana}          & 4.8B                              & 5.0                               & 1024 $\times$ 1024                             & 20                                \\
Lumina-Image 2.0~\cite{lumina2}                 & 2.6B                              & 4.0                              & 1024 $\times$ 1024                             & 50                                \\
HiDream-I1-Full~\cite{2025hidreami1}            & 17B                               & 5.0                               & 1024 $\times$ 1024                             & 50         \\
        \bottomrule
    \end{tabular}
    }
    \label{tab:config1}
\end{table}

For closed-source methods, we present the corresponding release or update dates of the methods in Table~\ref{tab:config2} to facilitate alignment with subsequent experimental results.
\begin{table}[ht]
\vspace{-0.12cm}
    \centering
    \caption{\small \textbf{Release/Update date of closed-source methods}. Release/Update date represents the version of the corresponding method when generating images.}
    \vspace{0.2cm}
     \resizebox{0.98\columnwidth}{!}{
    \begin{tabular}{ccccccc}
        \toprule
\textbf{Method}  & Imagen3~\cite{2024Imagen3} & Recraft V3~\cite{2024recraftv3} & Kolors 2.0~\cite{2025Kolors2} & Seedream 3.0~\cite{gao2025seedream} & Imagen4~\cite{2025Imagen4} & GPT-4o~\cite{openai2024gpt4o_image} \\
\midrule
\textbf{Release/Update Date} & 2025-01-23                                         & 2024-10-30                                              & 2025-04-15                                            & 2025-04-15                                                  & 2025-05-20 & 2025-04-29          \\
        \bottomrule
    \end{tabular}
    }
    \label{tab:config2}
\vspace{-0.2cm}
\end{table}

\subsection{The Details on Prompts Rewriting}

\begin{algorithm}
\caption{\textbf{Initial Prompts Rewritten by GPT-4o}}
\KwIn{$n$: the length of the initial prompts list, 

~~~~~~~~~~~~$\mathbb{P}_{\text{init}}$: the initial prompts $[p_1, p_2, \dots, p_n]$.}

\KwOut{$\mathbb{P}_{\text{rewritten}}$ : the rewritten prompts $[p'_1, p'_2, \dots, p'_n]$}


$\mathbb{P}_{\text{sorted}} \gets \text{sorted\_by\_word\_count}\left(\mathbb{P}_{\text{init}}\right)$

$R \gets 100 *\text{sorted} (\text{beta.rvs}(2.37, 2.86, n))$

\For{$i \gets 1$ \textbf{to} $n$}{

    $\text{initial\_prompt} \gets \mathbb{P}_{\text{sorted}}[i]$
    
    $\text{target\_word\_count} \gets R[i]$
    
    $\text{rewritten\_prompt} \gets \text{GPT-4o\_API}(\text{Prompt Template},\ \text{initial\_prompt},\ \text{target\_word\_count})$

    $\mathbb{P}_{\text{rewritten}}[i] \gets \text{rewritten\_prompt}$
}
$\textbf{return}~~\mathbb{P}_{\text{rewritten}}$
\label{alg:rewrite}
\end{algorithm}

As shown in Algorithm~\ref{alg:rewrite}, the initial prompts are first sorted based on their word count. Then, using a Beta distribution with parameters (2.37, 2.86), which roughly follows the ratio 0-0.3:0.3-0.6:0.6-1 $\approx$ 1:2:1, a list of desired prompt lengths is generated and subsequently sorted. The sorted prompts are then matched with the corresponding desired lengths, and the GPT-4o API is called to rewrite each prompt according to its specified length, resulting in the rewritten prompts. Without loss of generality, lengths in the range of 60-100 can be mapped to a range of greater, thereby generating longer prompts. In this process, the corresponding prompt for rewriting is as follows:

\begin{tcolorbox}[title=Prompt of Prompts Rewriting,colback=yellow!10!white,colframe=orange!80!black,fonttitle=\bfseries]
You are a precise rewriting assistant. \\
\textbf{Task Description:}
\begin{itemize}[itemsep=-1.4pt,leftmargin=2em]
\item Rewrite the \textcolor{blue}{\textless initial\_prompt \textgreater} according to the \textcolor{blue}{\textless target\_word\_count\textgreater} of the prompt.
\item For \textcolor{blue}{\textless initial\_prompt \textgreater} longer than the \textcolor{blue}{\textless target\_word\_count\textgreater} 
\begin{itemize}[itemsep=-1.4pt,leftmargin=2em]
\item Shorten the prompt by carefully removing specific but non-essential details.
\item Do not simply delete words or generalize the description.
  \end{itemize}
\item For \textcolor{blue}{\textless initial\_prompt \textgreater} shorter than the \textcolor{blue}{\textless target\_word\_count\textgreater} 
\begin{itemize}[itemsep=-1.4pt,leftmargin=2em]
\item Expand the prompt by adding specific, meaningful, and vivid details that enhance the scene. 
\item Do not introduce abstract or generalized commentary. 
  \end{itemize}
\item Ensure the rewritten prompt 
\begin{itemize}[itemsep=-1.4pt,leftmargin=2em]
\item The prompt should be coherent, natural, fluent, logically structured. 
\item Please maintain the initial tone and intent as much as possible. 
\end{itemize}
\end{itemize}
\textbf{Important:}
\begin{itemize}[itemsep=-1.4pt,leftmargin=2em]
\item Only output the final rewritten prompt without any additional words. 
\end{itemize}
\end{tcolorbox}

\subsection{The Details and Analysis on Stylization}
\begin{table}[ht]
\vspace{-0.3cm}
\centering
\footnotesize
\caption{\small \textbf{The styles in OneIG-Bench corresponding to specific categories}.}
\vspace{0.2cm}
\scalebox{1}{
\begin{tabular}{>{\centering\arraybackslash}m{1.6cm}|p{7cm}}
\toprule
\textbf{Category}   &  \textbf{Style}    \\
\midrule
Traditional  & abstract expressionism, art nouveau, Baroque, Chinese ink painting, cubism, fauvism, impressionism, line art, minimalism, pointillism, pop art, Rococo, Ukiyo-e \\
\midrule
Media     &  clay, crayon, graffiti, LEGO, pencil sketch, stone sculpture, watercolor\\
\midrule
Anime    &  Celluloid, Chibi,  comic, Cyberpunk, Ghibli, Impasto, Pixar, pixel art, 3d rendering,  \\ 
\bottomrule
\end{tabular}}
    \label{tab:style_name}
    \vspace{-0.2cm}
\end{table}
  
The Anime \& Stylization category encompasses a variety of styles, which are systematically grouped into three subcategories in Table~\ref{tab:style_name}: \textbf{Traditional}, \textbf{Media}, and \textbf{Anime}. The \textbf{Traditional} category primarily includes styles rooted in classical and historical art movements from around the world. The \textbf{Media} category includes styles defined by specific artistic media and material-based techniques. The \textbf{Anime} category represents a collection of stylized and detailed visual aesthetics commonly associated with animation and pop culture. And Table~\ref{tab:style} presents the scores of different methods across various style categories. The calculation process is as follows: for each method, the average style score is first calculated based on the images generated for each prompt within each style. Then, the score for each style category is obtained by averaging the scores of the individual styles within that category. It is clear that GPT-4o~\cite{openai2024gpt4o_image} demonstrates exceptional style-following ability across most categories, significantly outperforming other methods.

\begin{table}[ht]
\centering
\footnotesize
\caption{\small \textbf{The style scores on different categories of styles}. \legendsquare{colorfirst} \legendsquare{colorsecond} \legendsquare{colorthird} \legendsquare{colorfourth} \legendsquare{colorfifth} indicate the first, second, third, fourth, and fifth performance, respectively.}
\vspace{0.2cm}
\scalebox{1}{
\begin{tabular}{l|ccccccc}
\toprule
\textbf{Method}   & ~~~~\textbf{Traditional}~~~~  & ~~~~\textbf{Media}~~~~ & ~~~~\textbf{Anime}~~~~ \\
\midrule
Janus-Pro~\cite{chen2025janus} & 0.224 & 0.212 & 0.412 \\
BLIP3-o~\cite{chen2025blip3} & 0.381 & 0.287 & 0.390 \\
BAGEL~\cite{deng2025bagel} & 0.363 & 0.297 & \rankfifth{0.440} \\
BAGEL+CoT~\cite{deng2025bagel} & 0.372 & 0.329 & \ranksecond{0.499} \\
Show-o2-1.5B~\cite{xie2025showo2} & 0.269 & 0.288 & 0.410 \\
Show-o2-7B~\cite{xie2025showo2} & 0.283 & 0.295 & 0.380 \\
OmniGen2~\cite{wu2025omnigen2} & 0.360 & 0.314 & \rankthird{0.458} \\
\midrule
Stable Diffusion 1.5~\cite{Rombach_2022_CVPR}     & \ranksecond{0.483} & 0.298 & 0.349 \\
Stable Diffusion XL~\cite{podell2023sdxl}     & 0.316 & 0.307 & 0.339 \\
Stable Diffusion 3.5 Large~\cite{2024sd3.5} & 0.356 & 0.315 & 0.335 \\
Flux.1-dev~\cite{flux2024} & 0.367 & 0.298 & 0.391 \\
CogView4~\cite{2025cogview4} & 0.376 & 0.294 & 0.369 \\
SANA-1.5 1.6B(PAG)~\cite{xie2025sana}     & \rankfourth{0.438} & 0.331 & 0.370 \\
SANA-1.5 4.8B(PAG)~\cite{xie2025sana} & \rankthird{0.443} & \rankfifth{0.340} & 0.379 \\
Lumina-Image 2.0~\cite{lumina2}   & 0.351 & 0.325 & 0.360 \\
HiDream-I1-Full~\cite{2025hidreami1}  & 0.331 & 0.295 & 0.368 \\
\midrule
Imagen3~\cite{2024Imagen3}  & 0.378 & 0.309 & 0.371 \\
Recraft V3~\cite{2024recraftv3}   & \rankfifth{0.418} & \rankfourth{0.347} & 0.332 \\
Kolors 2.0~\cite{2025Kolors2}    & 0.370 & 0.336 & 0.360 \\
Seedream 3.0~\cite{gao2025seedream}   & 0.383 & \ranksecond{0.365} & \rankfirst{0.524} \\
Imagen4~\cite{2025Imagen4}   & 0.336 & \ranksecond{0.365} & \rankfourth{0.452} \\
GPT-4o~\cite{openai2024gpt4o_image}    & \rankfirst{0.532} & \rankfirst{0.404} & 0.411 \\
\bottomrule
\end{tabular}}
    \label{tab:style}
    \vspace{-0.2cm}
\end{table}

\subsection{Visualization Results}

We first give the best normalized polar visualization of SOTA methods on OneIG-Bench in Table~\ref{tab:polar_methods}. It can be observed that the closed-source methods outperform the other two categories of methods overall. And we selected some methods based on their overall performance across non-style metrics and showcased representative examples for each. For the style dimension, we further selected images from three finer-grained subcategories to illustrate the results. \textbf{In all visualizations, the methods are broadly selected based on overall performance. In the following figures, each image tile is labeled in the top-left corner with the score achieved by the corresponding method under the current evaluation metric.}

\begin{figure}[ht]
    \centering
    \setlength{\abovecaptionskip}{1pt}
    \includegraphics[width=.98\linewidth]{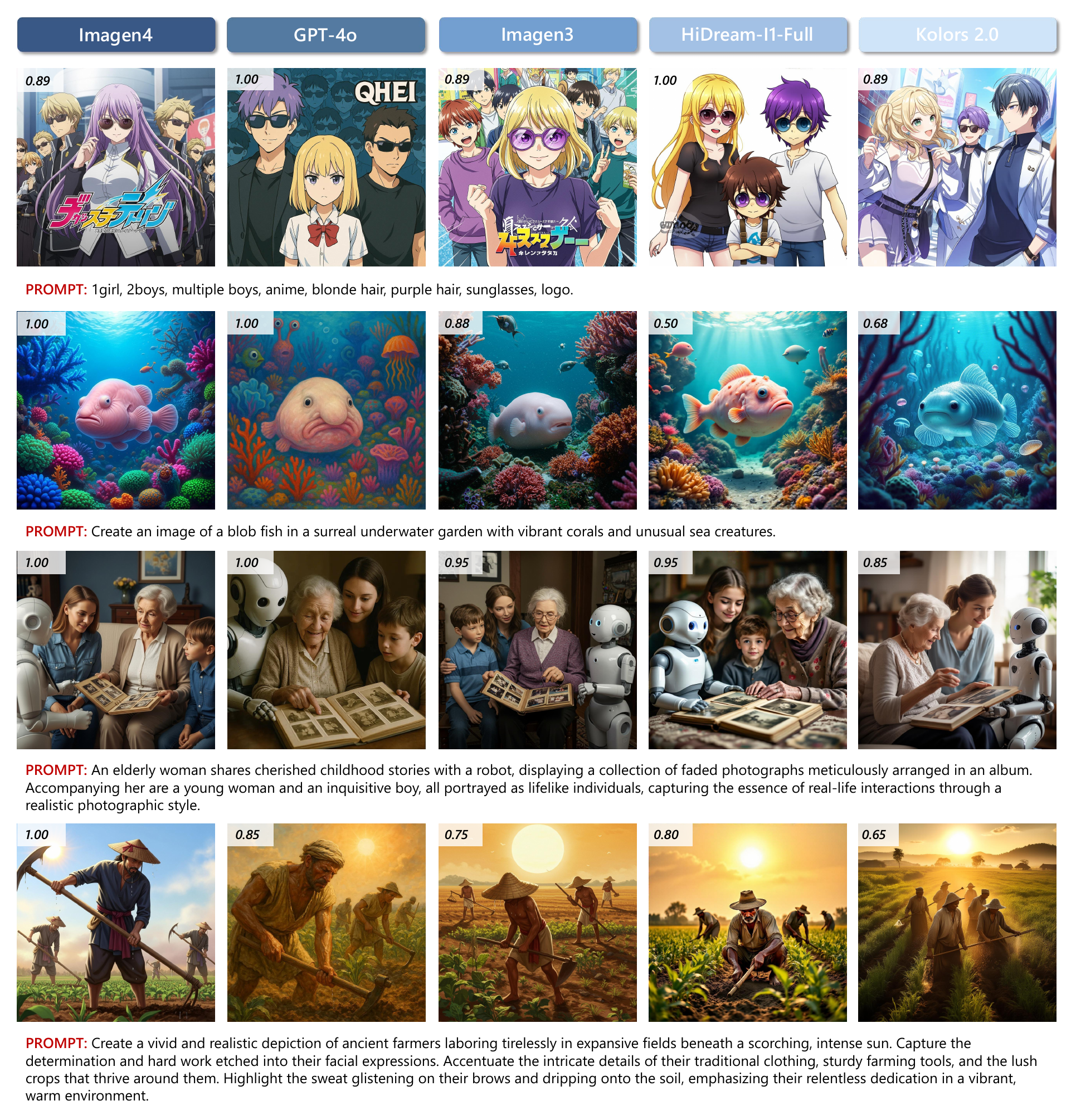}
    \caption{\small \textbf{Visualization results of SOTA methods.} \textbf{Alignment} of Imagen4~\cite{2025Imagen4}, GPT-4o~\cite{openai2024gpt4o_image}, Imagen3~\cite{2024Imagen3}, HiDream-I1-Full~\cite{2025hidreami1} and Kolors 2.0~\cite{2025Kolors2} are evaluated respectively. \textbf{Row 1} corresponds to \textbf{tag/phrase prompt}: The variation in the scores of the visual samples are mainly influenced by: "2 boys" and "logo". \textbf{Row 2} corresponds to \textbf{short prompt}: The variation in the scores of the visual samples are mainly influenced by: "blob fish", "surreal underwater" and "unusual sea creatures". \textbf{Row 3} corresponds to \textbf{middle prompt}: The variation in the scores of the visual samples are influenced by: "inquisitive boy", "realistic photography style" and whether each individual mentioned in the prompt is accurately and uniquely generated in the image. \textbf{Row 4} corresponds to \textbf{long prompt}: The variation in the scores of the visual samples are influenced by: "ancient farmers", "sweat", "facial expressions", "relentless dedication". The mentioned keywords may correspond to more than one question–answer pair.}
    \label{fig:vis_alignment}
    \vspace{-0.35cm}
\end{figure}

Figure~\ref{fig:vis_alignment} shows that Imagen4~\cite{2025Imagen4} and GPT-4o~\cite{openai2024gpt4o_image} demonstrate strong capabilities in semantic alignment. Notably, in multi-person generation tasks, many methods struggle to accurately fulfill requirements at the individual level and often exhibit confusion in assigning attributes to the correct subjects. In addition, some methods tend to overlook fine-grained details while focusing on the primary generation task, which significantly hinders their ability to achieve high alignment scores.

\begin{figure}[ht]
    \centering
    \setlength{\abovecaptionskip}{1pt}
    \includegraphics[width=.98\linewidth]{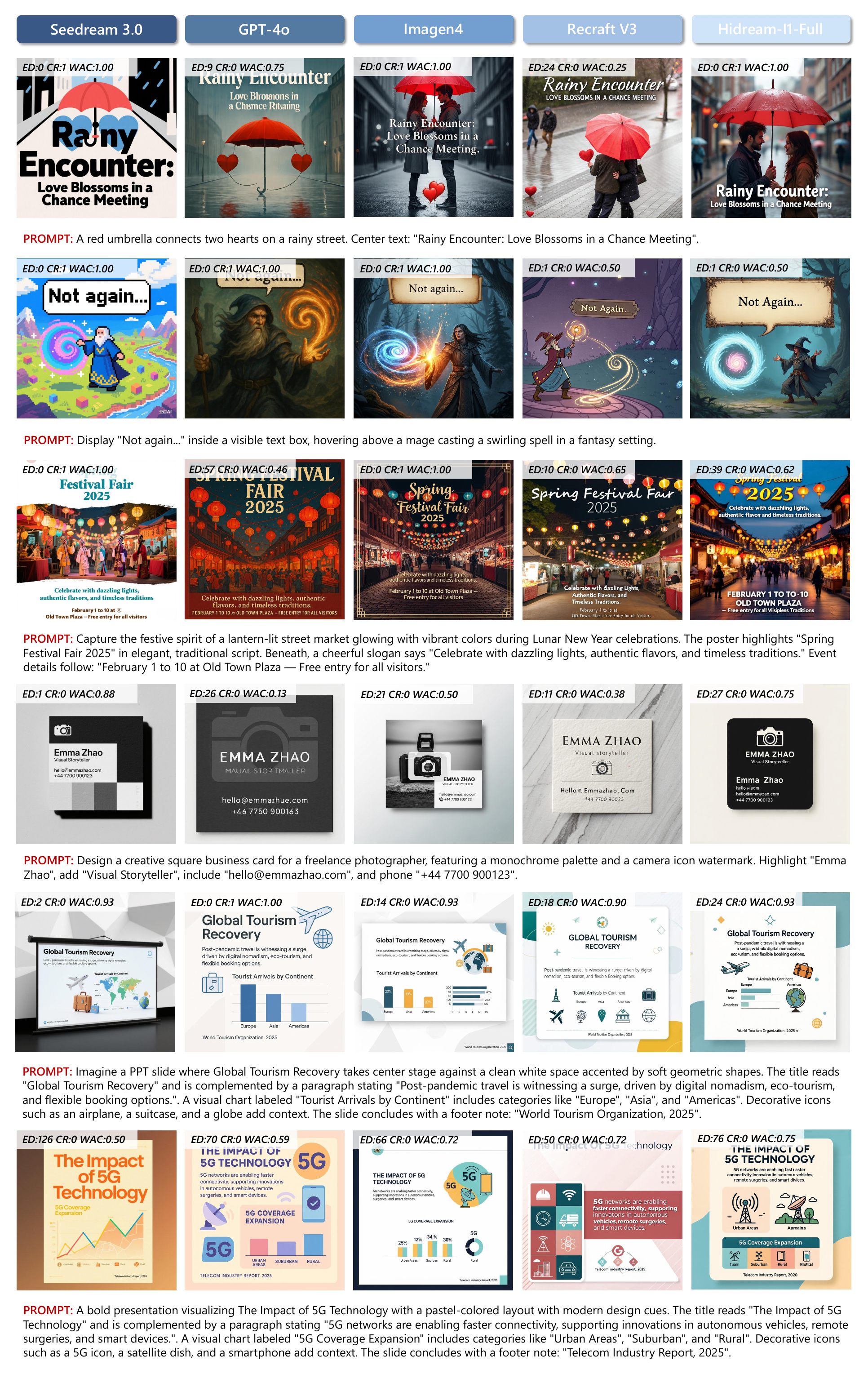}
    \caption{\small \textbf{Visualization results of SOTA methods.} \textbf{Text} of Seedream 3.0~\cite{gao2025seedream}, GPT-4o~\cite{openai2024gpt4o_image}, Imagen4~\cite{2025Imagen4}, Recraft V3~\cite{2024recraftv3}, HiDream-I1-Full~\cite{2025hidreami1} are evaluated respectively. \textbf{Row 1, 2} correspond to \textbf{short prompts}. \textbf{Row 3, 4} correspond to \textbf{middle prompts}. \textbf{Row 5, 6} correspond to \textbf{long prompts}.}
    \label{fig:vis_text}
\end{figure}

Text rendering is an important task for current generative methods. As shown in Figure~\ref{fig:vis_text}, Seedream 3.0~\cite{gao2025seedream} demonstrates high accuracy and aesthetic quality. Some methods, such as GPT-4o~\cite{openai2024gpt4o_image}, demonstrate limitations in adhering to case sensitivity (distinguishing between uppercase and lowercase letters), which compromises the textual accuracy of the rendered content. While the generated images may appear visually impressive, these subtle errors can lead to a noticeable divergence between subjective visual quality and objective evaluations based on metrics such as ED and WAC. Imagen4~\cite{2025Imagen4} demonstrates good accuracy in text generation, but the overall visual quality of the images is relatively poor. Recraft V3~\cite{2024recraftv3}, although rarely producing major errors in text generation, tends to make mistakes at the word level and suffers from inconsistencies in typography and layout coherence. Overall, HiDream-I1-Full~\cite{2025hidreami1} performs reasonably well in text rendering—it may not outperform Seedream 3.0, Recraft V3, or GPT-4o, but it is able to fulfill the basic prompt requirements.

\begin{figure}[ht]
    \centering
    \setlength{\abovecaptionskip}{1pt}
    \includegraphics[width=.98\linewidth]{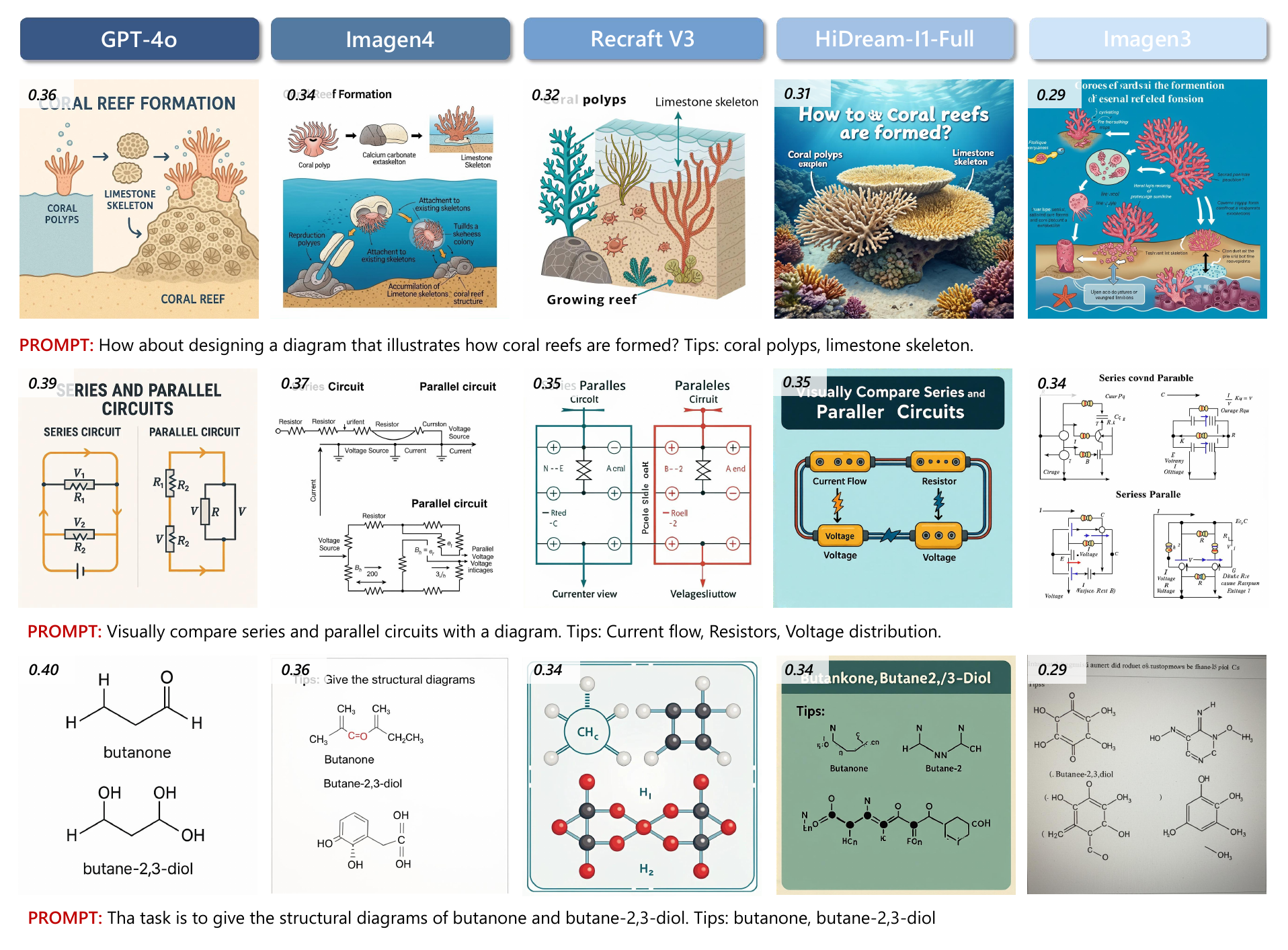}
    \caption{\small \textbf{Visualization results of SOTA methods.} \textbf{Reasoning} of GPT-4o~\cite{openai2024gpt4o_image}, Imagen4~\cite{2025Imagen4}, Recraft V3~\cite{2024recraftv3},  HiDream-I1-Full~\cite{2025hidreami1} and Imagen3~\cite{2024Imagen3} are evaluated respectively. \textbf{Row 1} aims to illustrate how coral reefs are formed, highlighting key steps such as the growth of coral polyps and the gradual accumulation of calcium carbonate structures. \textbf{Row 2} demonstrates the circuit diagrams of series and parallel circuits. \textbf{Row 3} focuses on the structural diagrams of butanone and butane-2,3-diol.}
    \label{fig:vis_reasoning}
    \vspace{-0.35cm}
\end{figure}

From a reasoning perspective, only GPT-4o~\cite{openai2024gpt4o_image} demonstrates both logical coherence and textual accuracy in Figure~\ref{fig:vis_reasoning}. Although Imagen4~\cite{2025Imagen4} and Recraft V3~\cite{2024recraftv3} fall short of GPT-4o in terms of clarity and correctness, they produce text that is generally readable. HiDream-I1-Full~\cite{2025hidreami1} provides limited textual and visual content but manages to convey a certain degree of knowledge and reasoning, albeit with insufficient accuracy. In contrast, Imagen3~\cite{2024Imagen3} tends to generate overly redundant outputs, with excessive textual and graphical elements that obscure the intended message, and often includes incorrect information. Therefore, Knowledge and Reasoning remains a critical area that warrants further investigation and refinement for generative methods.

\definecolor{Alignment1}{HTML}{dbe4ec}
\definecolor{Alignment2}{HTML}{b7c9d9}
\definecolor{Alignment3}{HTML}{93afc7}
\definecolor{Alignment4}{HTML}{6f94b4}
\definecolor{Alignment5}{HTML}{4b79a1}
\definecolor{Alignment6}{HTML}{4b79a1} 

\definecolor{Text1}{HTML}{d8ebe3}
\definecolor{Text2}{HTML}{c4e2d5}
\definecolor{Text3}{HTML}{b0d8c7}
\definecolor{Text4}{HTML}{9cceba}
\definecolor{Text5}{HTML}{89c4ac}
\definecolor{Text6}{HTML}{75ba9e}
\definecolor{Text7}{HTML}{61b190}
\definecolor{Text8}{HTML}{4ea782}
\definecolor{Text9}{HTML}{3a9d74}
\definecolor{Text10}{HTML}{3a9d74} 

\definecolor{Reasoning1}{HTML}{f4dcdc}
\definecolor{Reasoning2}{HTML}{ecc0c0}
\definecolor{Reasoning3}{HTML}{e3a3a3}
\definecolor{Reasoning4}{HTML}{da8787}
\definecolor{Reasoning5}{HTML}{d26b6b}
\definecolor{Reasoning6}{HTML}{c94f4f}
\definecolor{Reasoning7}{HTML}{c94f4f}

\definecolor{Style1}{HTML}{f9e8d8}
\definecolor{Style2}{HTML}{ecbb8b}
\definecolor{Style3}{HTML}{e08e3e}
\definecolor{Style4}{HTML}{e08e3e}

\definecolor{Diversity1}{HTML}{e5e5e7}
\definecolor{Diversity2}{HTML}{cbcbce}
\definecolor{Diversity3}{HTML}{b2b1b6}
\definecolor{Diversity4}{HTML}{98979d}
\definecolor{Diversity5}{HTML}{7e7d85}
\definecolor{Diversity6}{HTML}{7e7d85} 

\begin{table*}[t]
    \captionsetup{type=table}
    \caption{\setstretch{1.2}\selectfont \small \textbf{Best normalized polar visualization of SOTA methods on OneIG-Bench.} Anti-Clockwisely, \textbf{\underline{Alignment}}: \legendsquarebox{Alignment1} NP, \legendsquarebox{Alignment2}  T\&P, \legendsquarebox{Alignment3} Short, \legendsquarebox{Alignment4} Middle, \legendsquarebox{Alignment5} Long; \textbf{\underline{Text}}:  \legendsquarebox{Text1} ED-Short, \legendsquarebox{Text2} ED-Middle, \legendsquarebox{Text3} ED-Long, \legendsquarebox{Text4} CR-Short, \legendsquarebox{Text5} CR-Middle, \legendsquarebox{Text6} CR-Long, \legendsquarebox{Text7} WAC-Short, \legendsquarebox{Text8} WAC-Middle, \legendsquarebox{Text9} WAC-Long; 
    \textbf{\underline{Reasoning}}: \legendsquarebox{Reasoning1} Geography, \
\legendsquarebox{Reasoning2} Computer Science, \
\legendsquarebox{Reasoning3} Biology, \
\legendsquarebox{Reasoning4} Mathematics, \
\legendsquarebox{Reasoning5} Physics, \
\legendsquarebox{Reasoning6} Chemistry, \
\legendsquarebox{Reasoning7} Common Sense;
\textbf{\underline{Style}}: \legendsquarebox{Style1} Traditional, \legendsquarebox{Style2} Media, \legendsquarebox{Style3} Anime; 
\textbf{\underline{Diversity}}: \legendsquarebox{Diversity1} NP, \legendsquarebox{Diversity2}  T\&P, \legendsquarebox{Diversity3} Short, \legendsquarebox{Diversity4} Middle, \legendsquarebox{Diversity5} Long. The absolute average values of each evaluation metric are list on legends.
    }
    \label{tab:polar_methods}
    \centering
    \vspace{0.2cm}
    \resizebox{\linewidth}{!}{
    \begin{tabular}{ccc}
    \toprule
    \texttt{Janus-Pro} & \texttt{BLIP3-o} & \texttt{BAGEL} \\
    \includegraphics[width=0.32\linewidth]{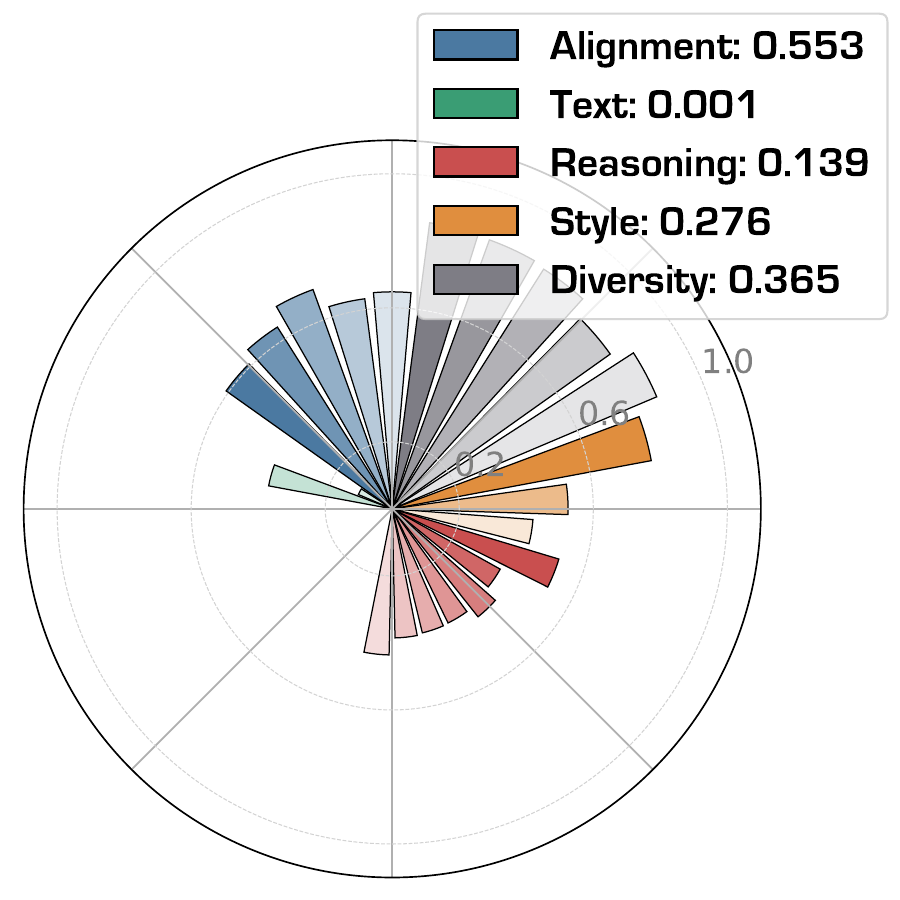} &
    \includegraphics[width=0.32\linewidth]{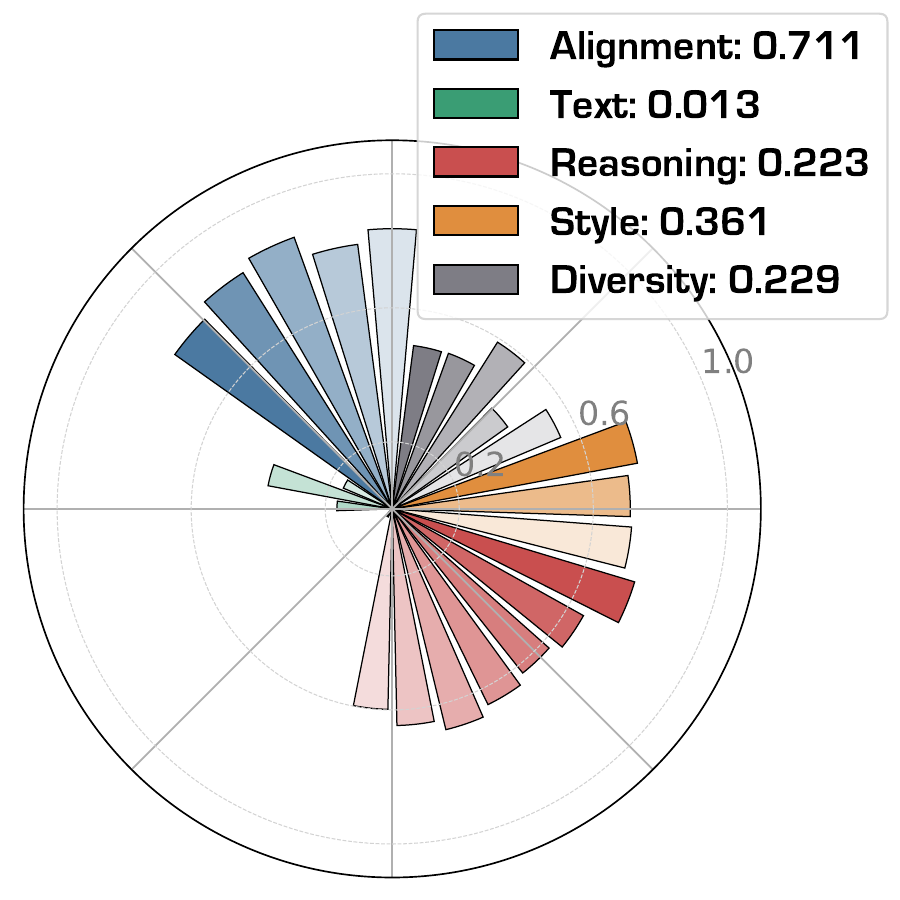} &
    \includegraphics[width=0.32\linewidth]{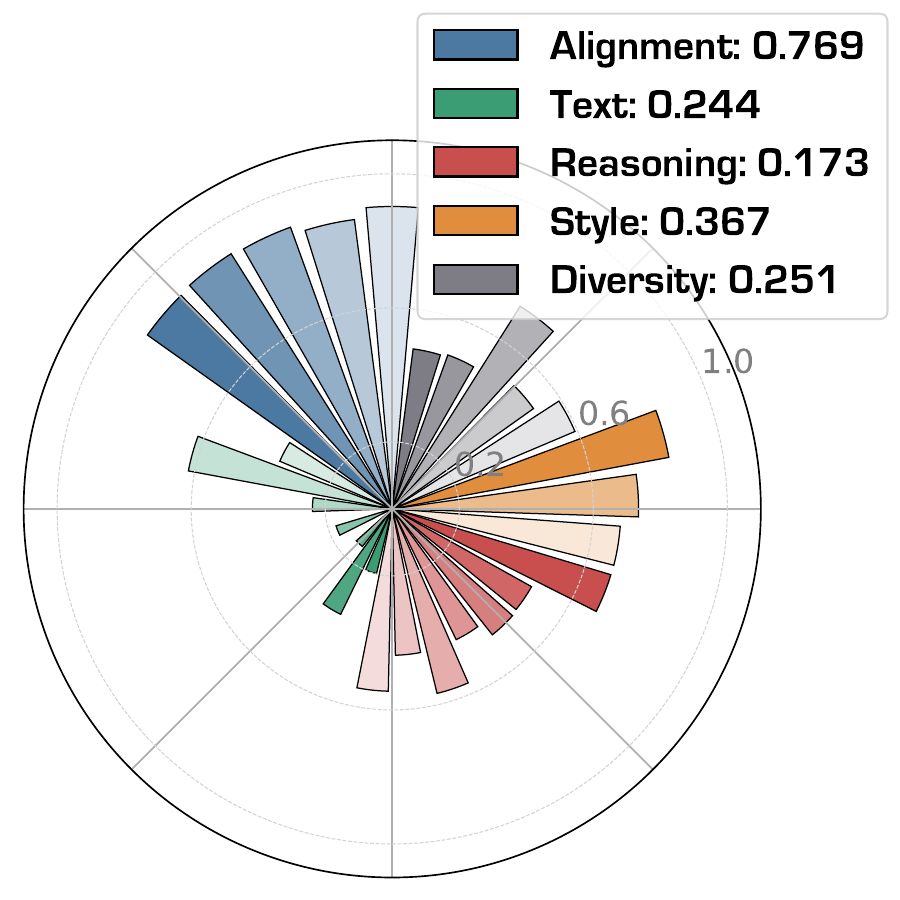} \\
    \midrule
    \texttt{BAGEL+CoT} & \texttt{Show-o2-1.5B(432)} & \texttt{Show-o2-7B(432)} \\
    \includegraphics[width=0.32\linewidth]{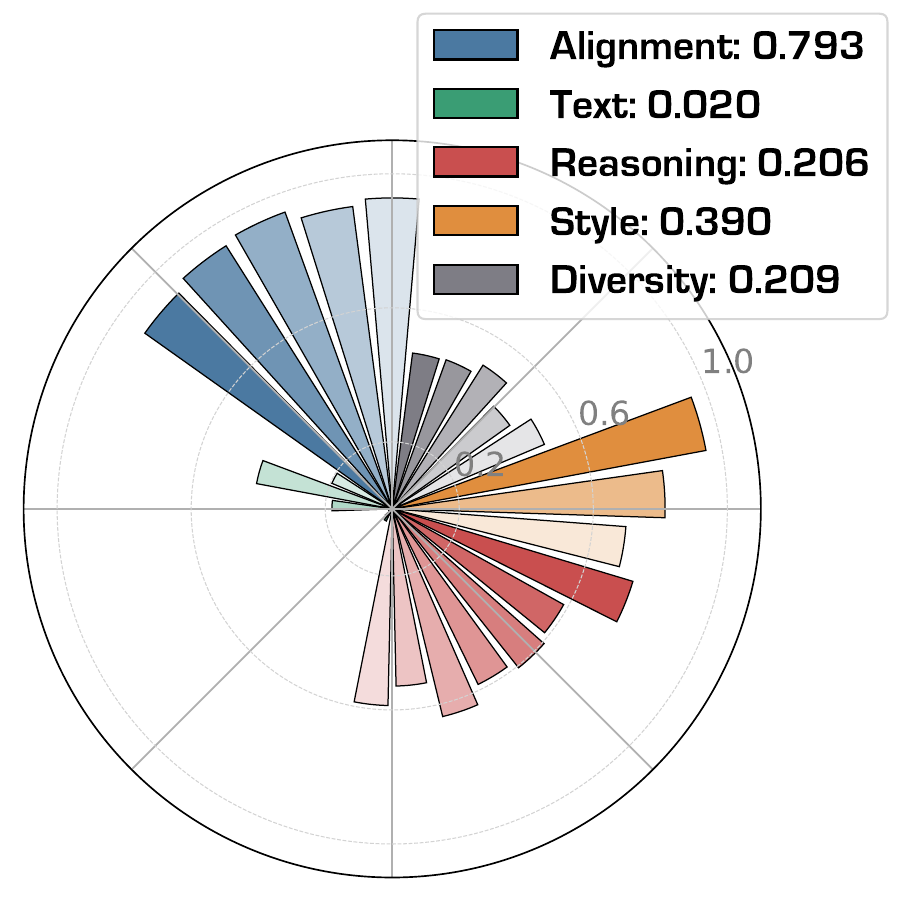} &
    \includegraphics[width=0.32\linewidth]{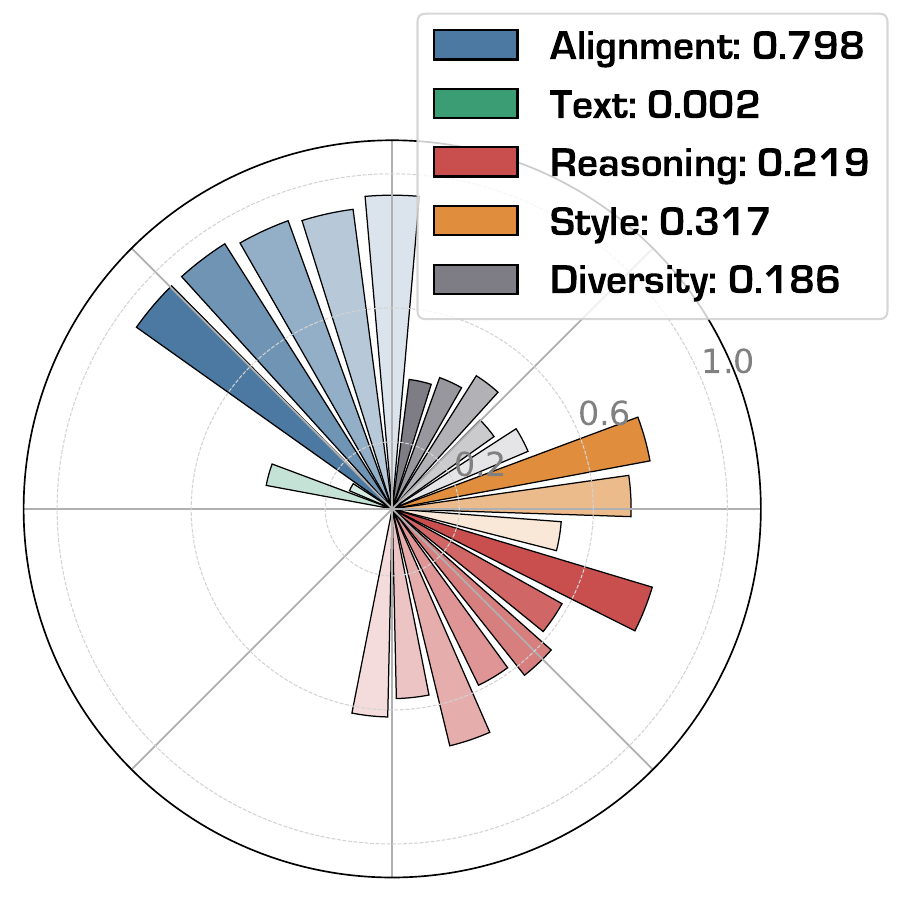} & 
    \includegraphics[width=0.32\linewidth]{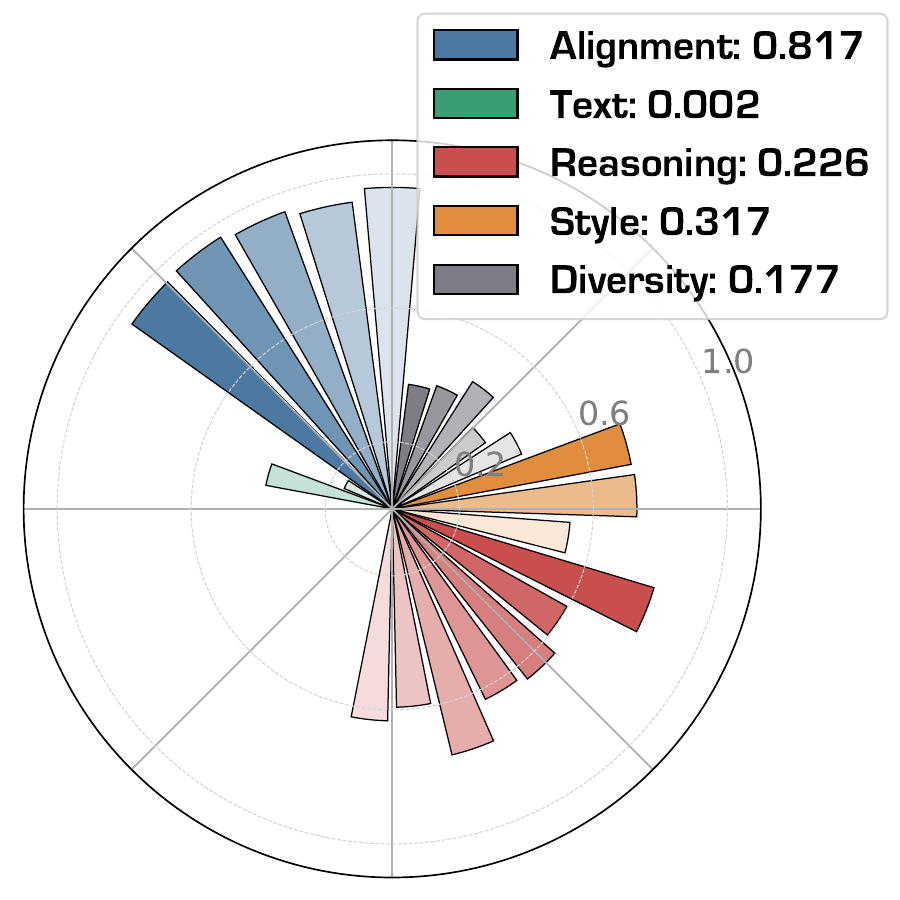} \\
    \midrule
    \texttt{OmniGen2} & \texttt{Stable Diffusion 1.5} & \texttt{Stable Diffusion XL} \\
    \includegraphics[width=0.32\linewidth]{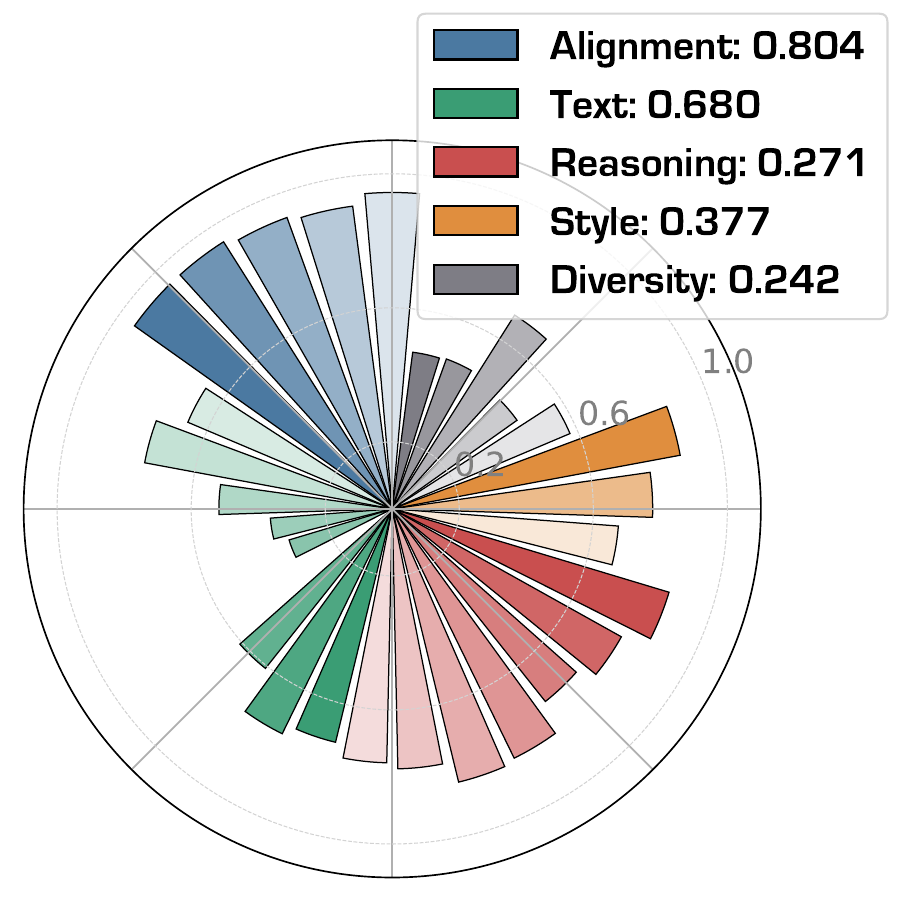} &
    \includegraphics[width=0.32\linewidth]{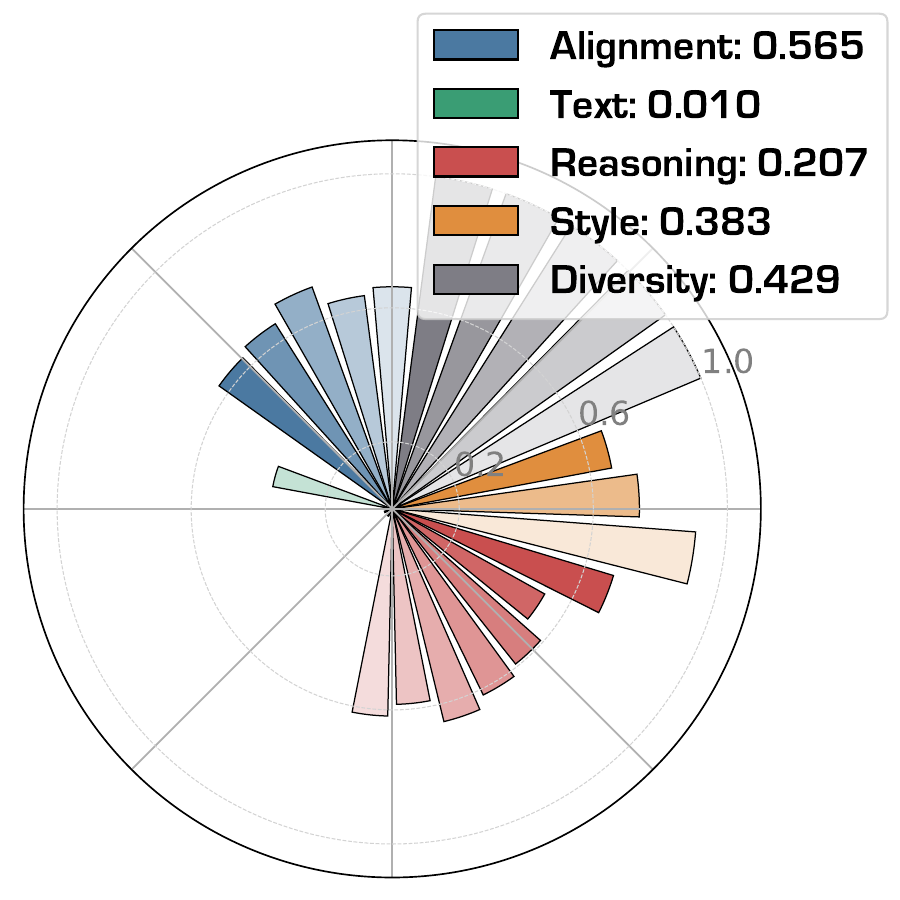} &
    \includegraphics[width=0.32\linewidth]{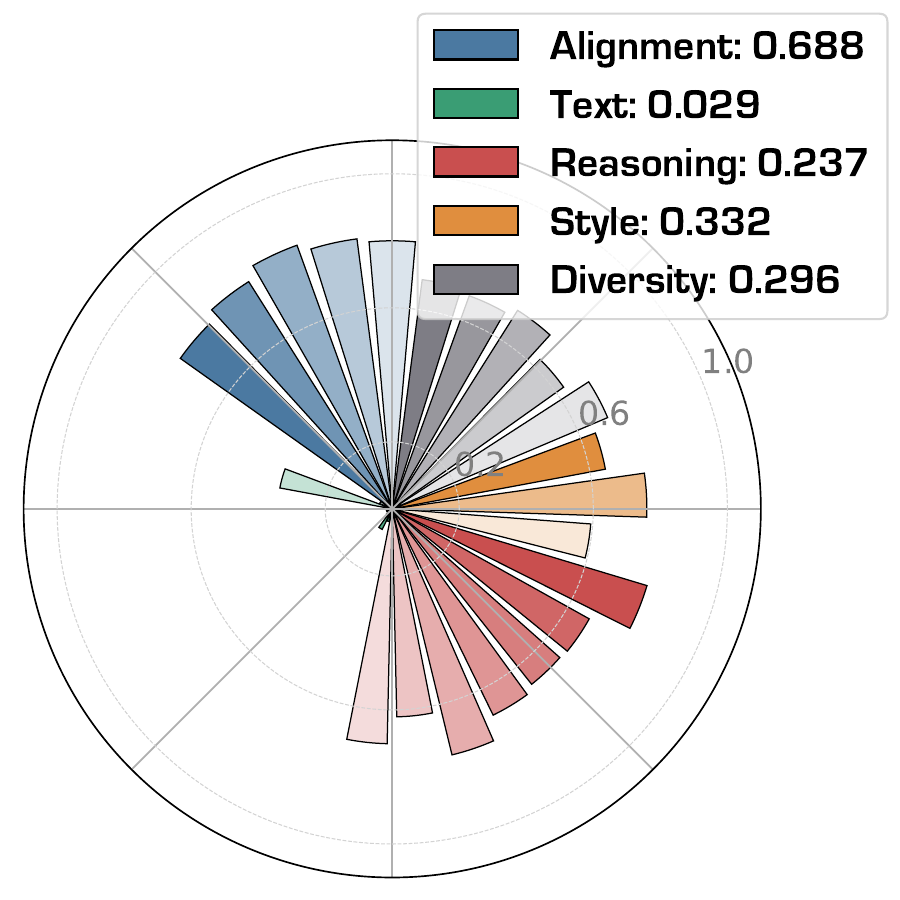} \\
    \midrule
    \texttt{Stable Diffusion 3.5 Large} & \texttt{Flux.1-dev} & \texttt{CogView4} \\
    \includegraphics[width=0.32\linewidth]{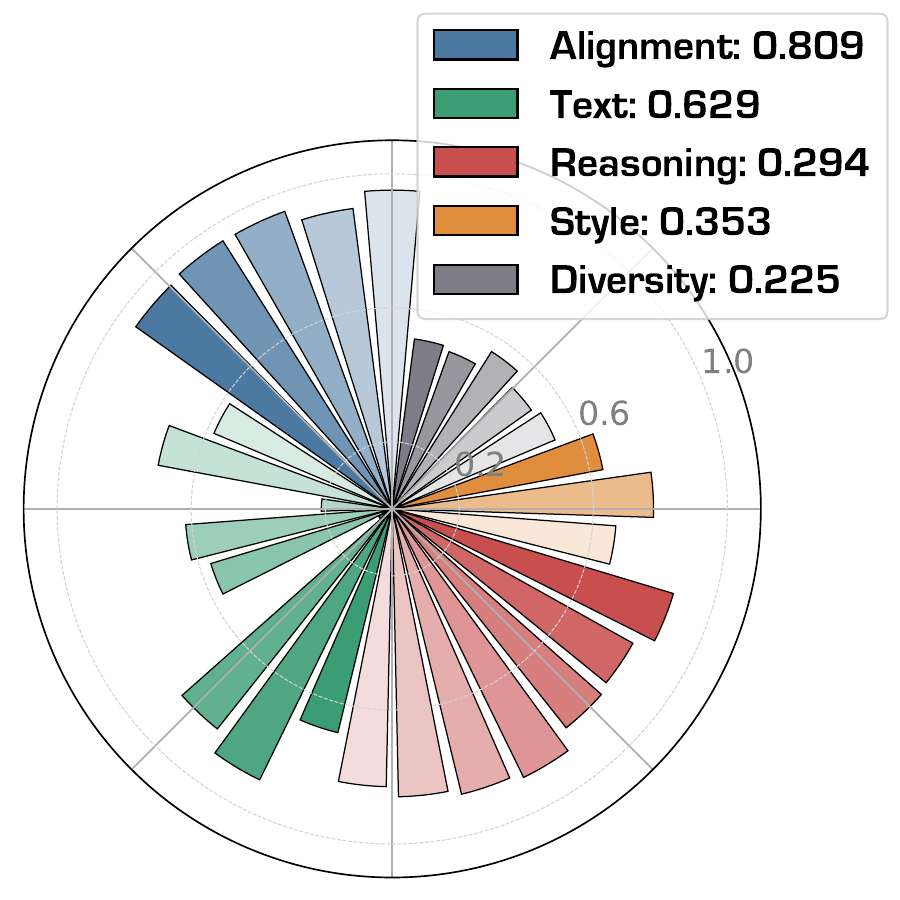} &
    \includegraphics[width=0.32\linewidth]{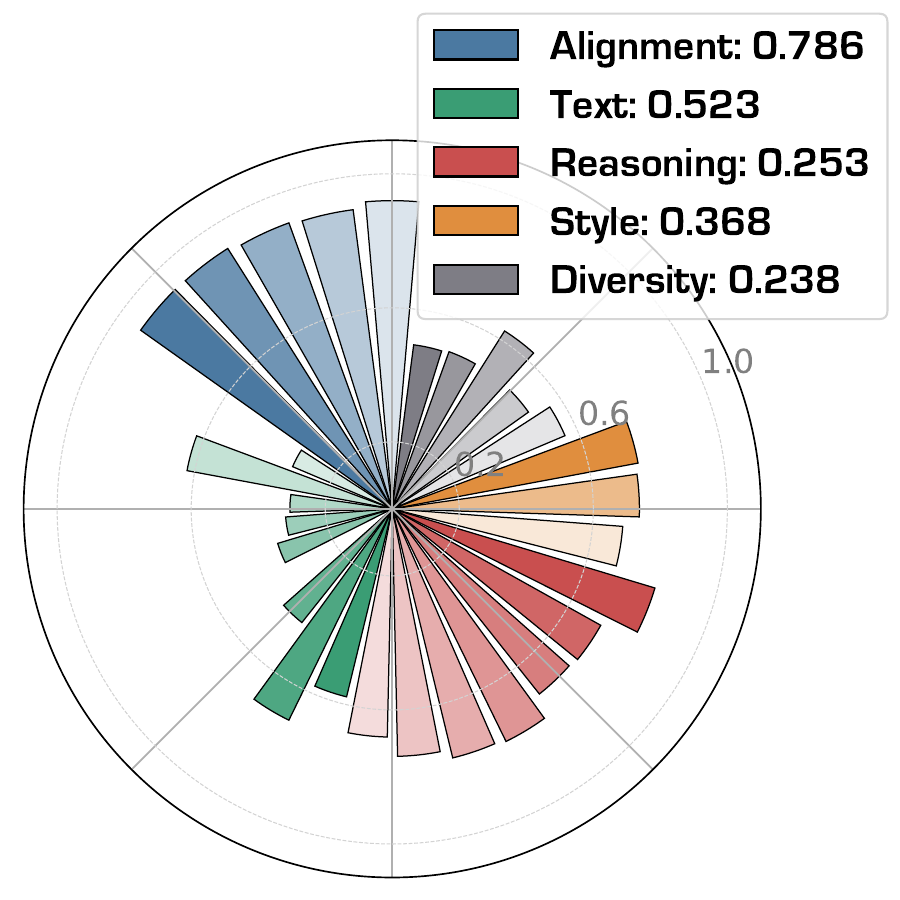} &
    \includegraphics[width=0.32\linewidth]{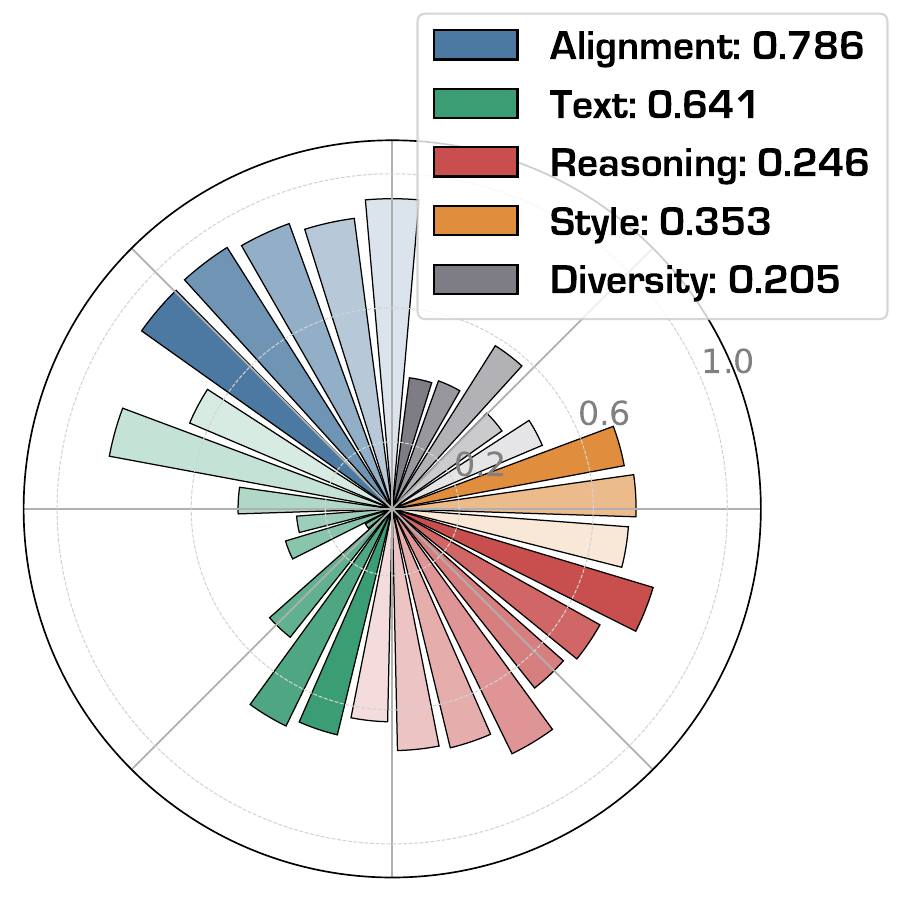} \\
    \bottomrule
    \end{tabular}}
    \vspace{-0.4cm}
\end{table*}


\begin{table*}[t]
    \captionsetup{type=table}
    \ContinuedFloat
    \caption{\setstretch{1.2}\selectfont \small \textbf{[Continued] Best normalized polar visualization of SOTA methods on OneIG-Bench.} Anti-Clockwisely, \textbf{\underline{Alignment}}: \legendsquarebox{Alignment1} NP, \legendsquarebox{Alignment2}  T\&P, \legendsquarebox{Alignment3} Short, \legendsquarebox{Alignment4} Middle, \legendsquarebox{Alignment5} Long; \textbf{\underline{Text}}:  \legendsquarebox{Text1} ED-Short, \legendsquarebox{Text2} ED-Middle, \legendsquarebox{Text3} ED-Long, \legendsquarebox{Text4} CR-Short, \legendsquarebox{Text5} CR-Middle, \legendsquarebox{Text6} CR-Long, \legendsquarebox{Text7} WAC-Short, \legendsquarebox{Text8} WAC-Middle, \legendsquarebox{Text9} WAC-Long; 
    \textbf{\underline{Reasoning}}: \legendsquarebox{Reasoning1} Geography, \
\legendsquarebox{Reasoning2} Computer Science, \
\legendsquarebox{Reasoning3} Biology, \
\legendsquarebox{Reasoning4} Mathematics, \
\legendsquarebox{Reasoning5} Physics, \
\legendsquarebox{Reasoning6} Chemistry, \
\legendsquarebox{Reasoning7} Common Sense;
\textbf{\underline{Style}}: \legendsquarebox{Style1} Traditional, \legendsquarebox{Style2} Media, \legendsquarebox{Style3} Anime; 
\textbf{\underline{Diversity}}: \legendsquarebox{Diversity1} NP, \legendsquarebox{Diversity2}  T\&P, \legendsquarebox{Diversity3} Short, \legendsquarebox{Diversity4} Middle, \legendsquarebox{Diversity5} Long. The absolute average values of each evaluation metric are list on legends.
    }
    \centering
    \vspace{0.2cm}
    \resizebox{\linewidth}{!}{
    \begin{tabular}{ccc}
    \toprule
    \texttt{SANA-1.5 1.6B (PAG)} & \texttt{SANA-1.5 4.8B (PAG)} & \texttt{Lumina-Image 2.0} \\
    \includegraphics[width=0.32\linewidth]{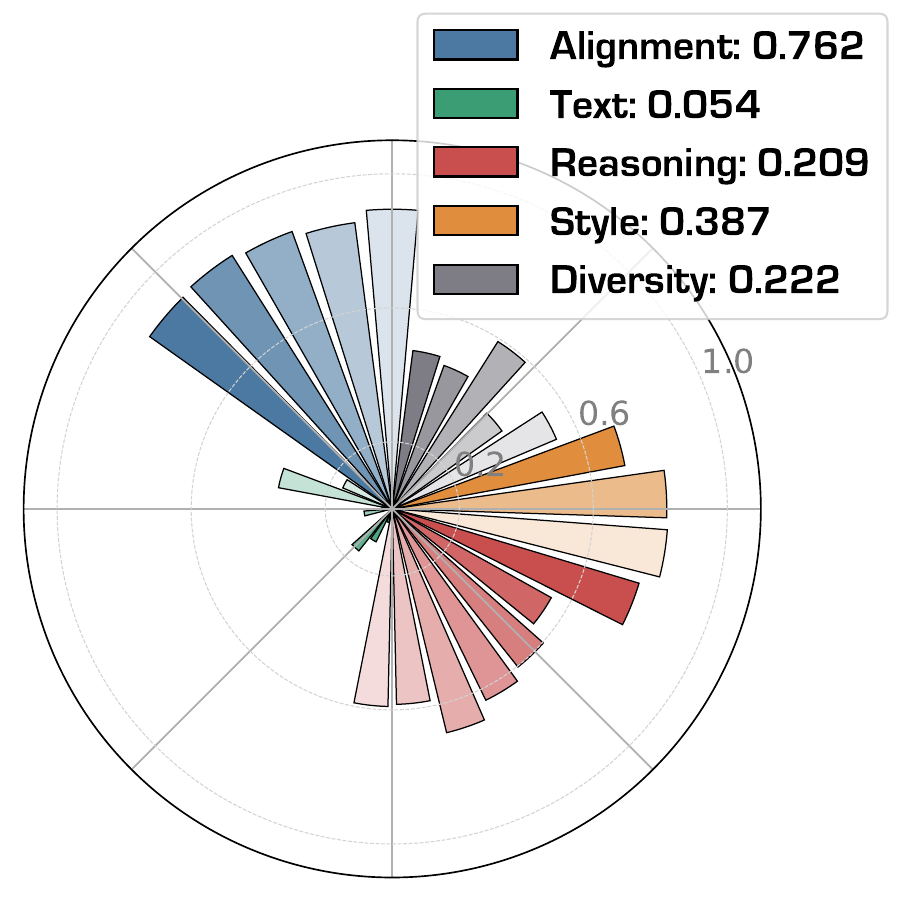} &
    \includegraphics[width=0.32\linewidth]{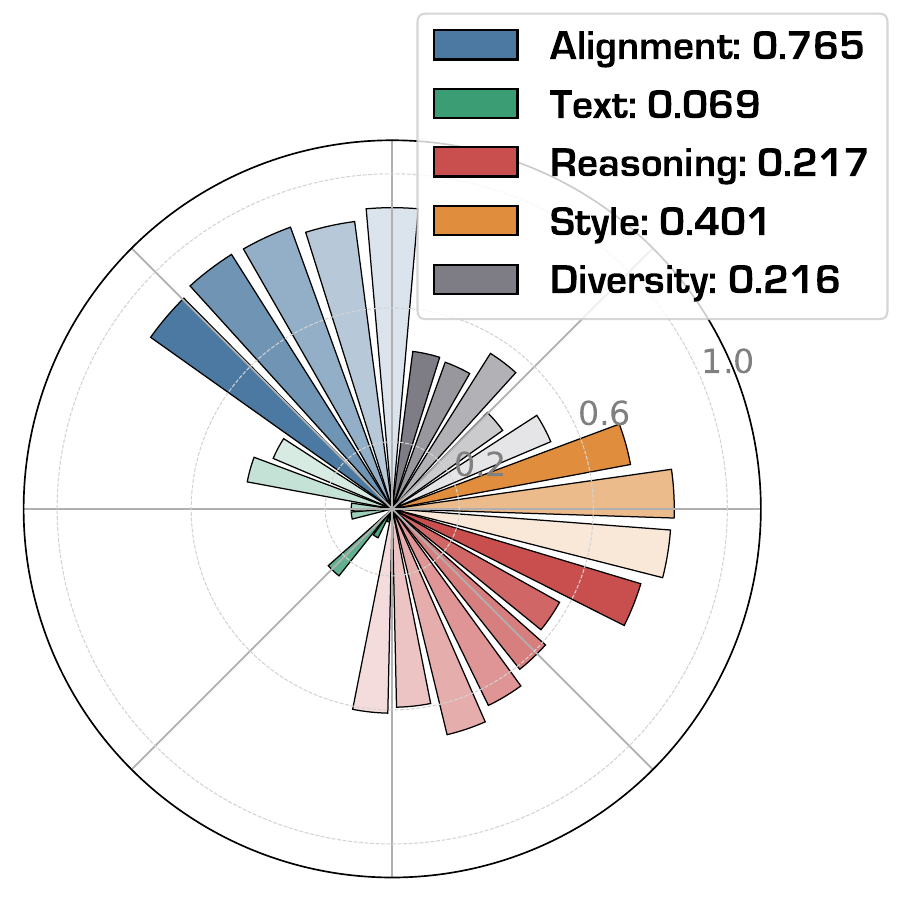} &
    \includegraphics[width=0.32\linewidth]{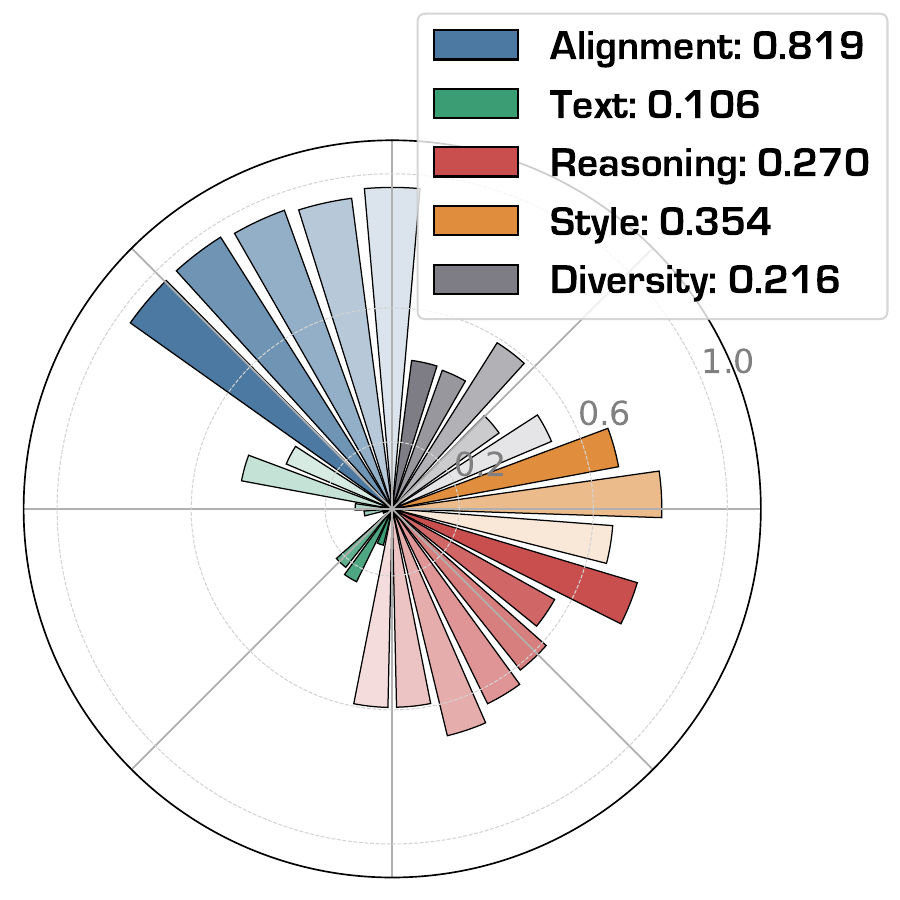} \\
    \midrule
    \texttt{HiDream-I1-Full} & \texttt{Imagen3} & \texttt{Recraft V3} \\
    \includegraphics[width=0.32\linewidth]{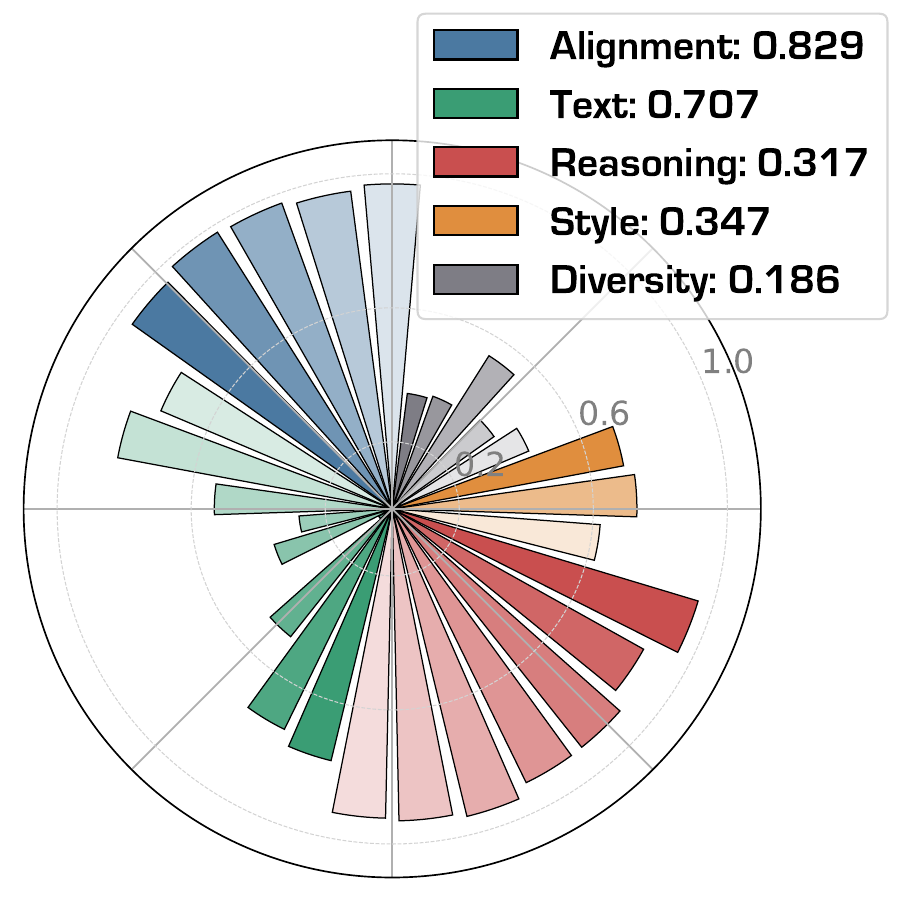} &
    \includegraphics[width=0.32\linewidth]{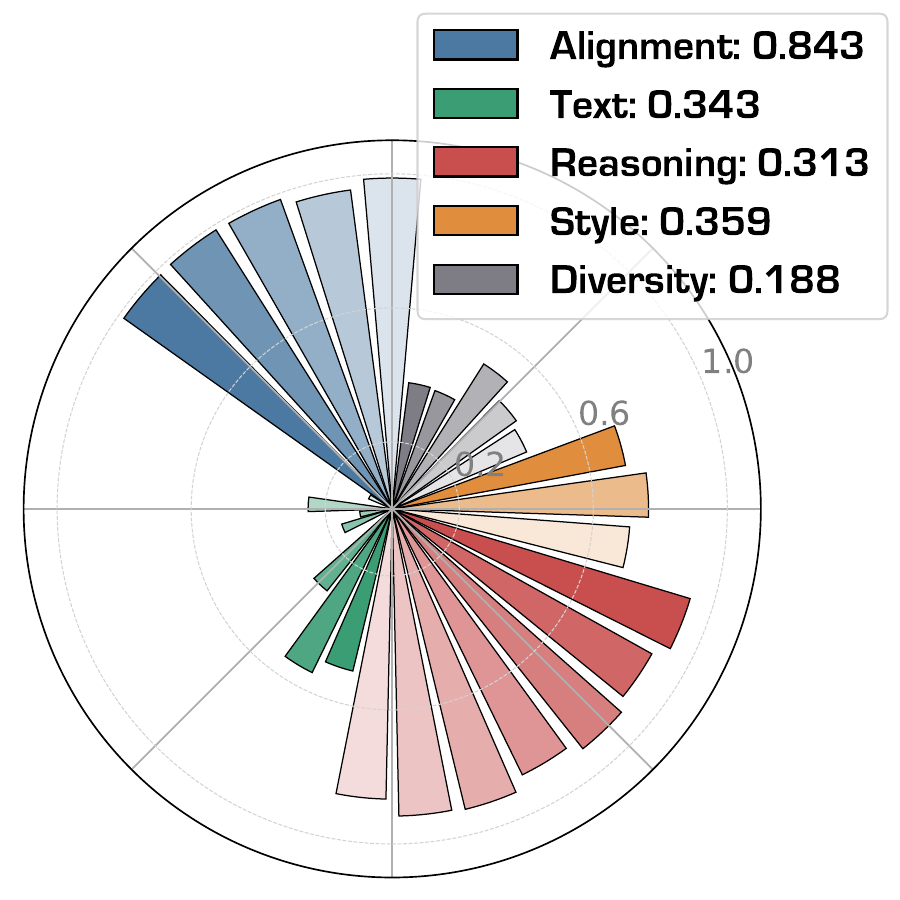} &
    \includegraphics[width=0.32\linewidth]{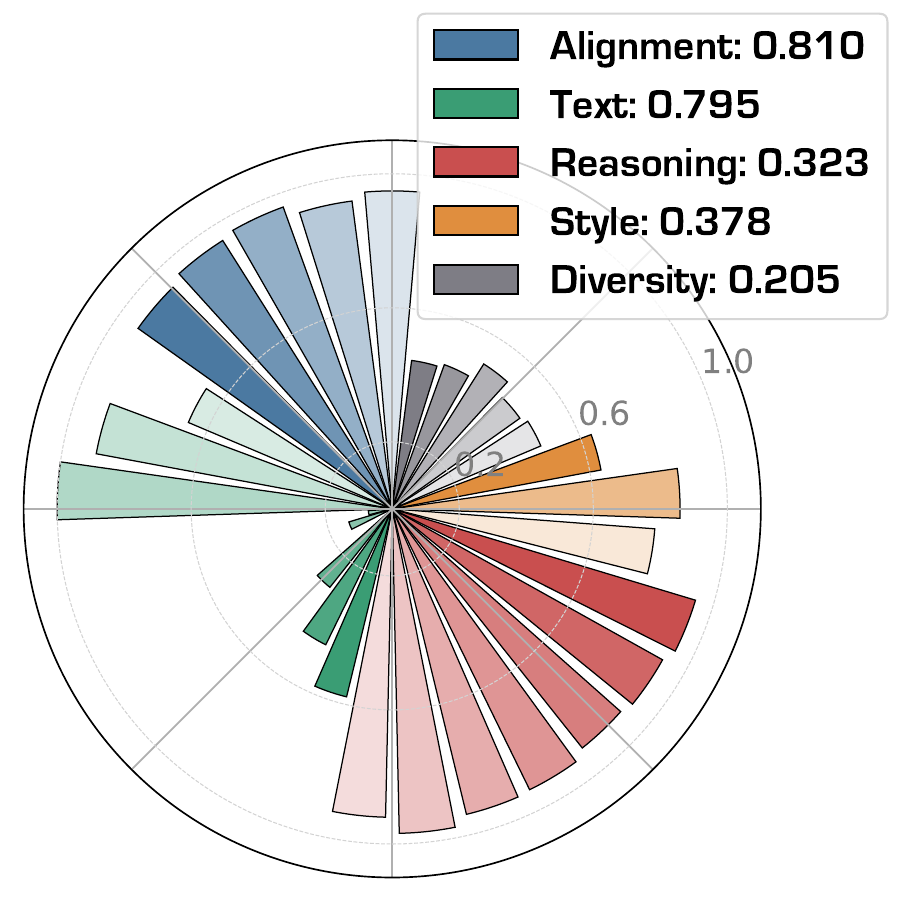} \\
    \midrule
    \texttt{Kolors 2.0} & \texttt{Seedream 3.0} & \texttt{Imagen4} \\
    \includegraphics[width=0.32\linewidth]{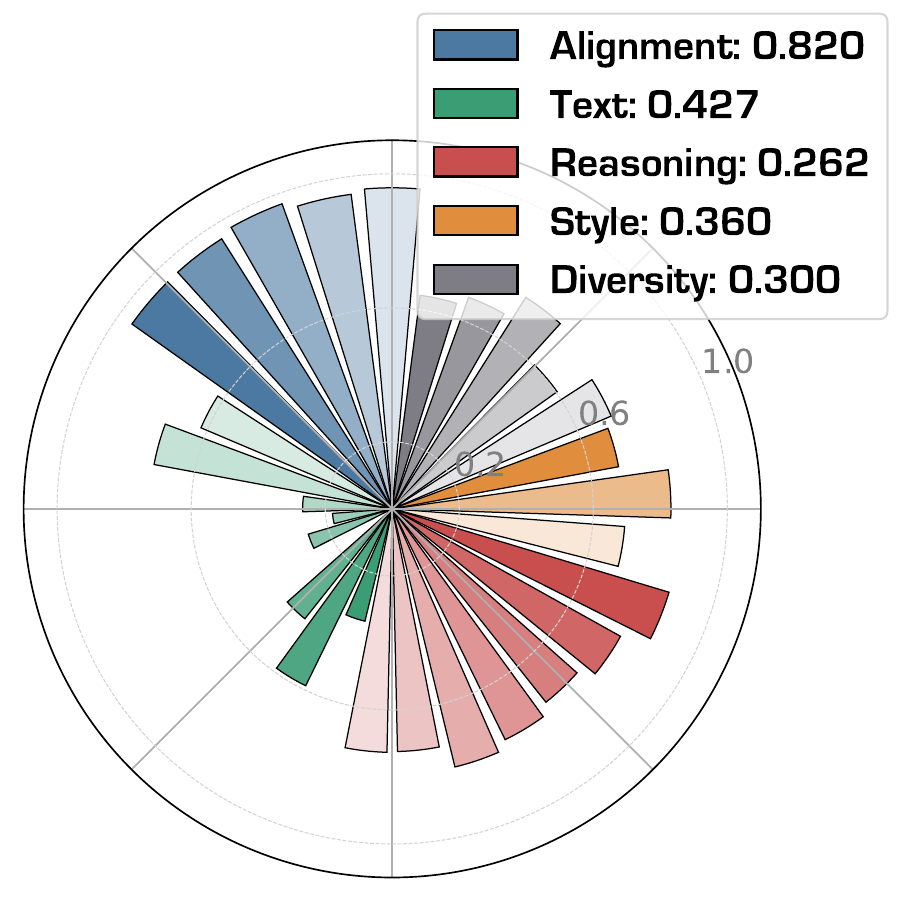} &
    \includegraphics[width=0.32\linewidth]{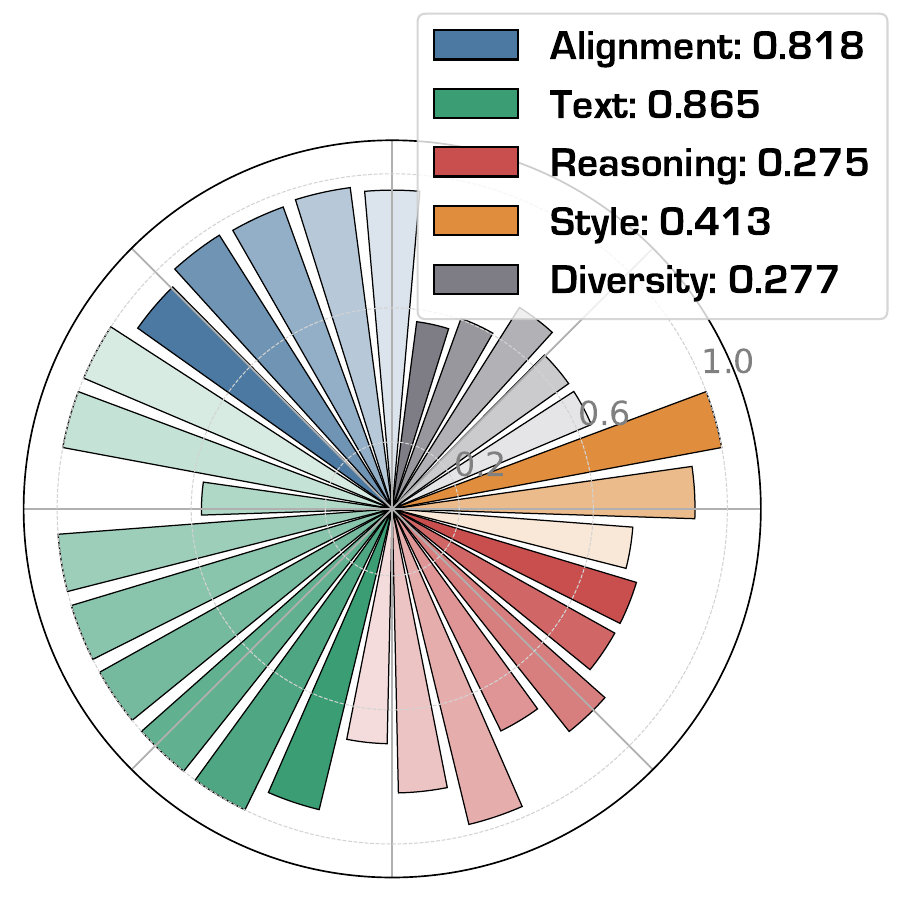} &
    \includegraphics[width=0.32\linewidth]{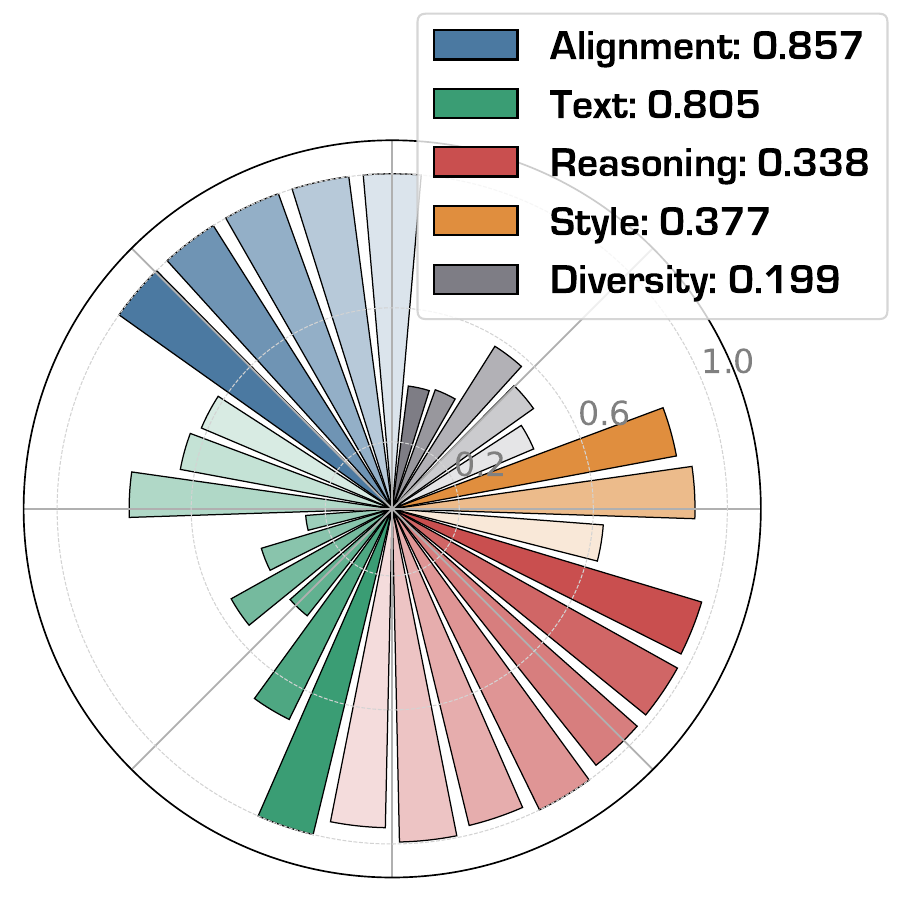} \\
    \midrule
    \texttt{GPT-4o} & & \\
    \includegraphics[width=0.32\linewidth]{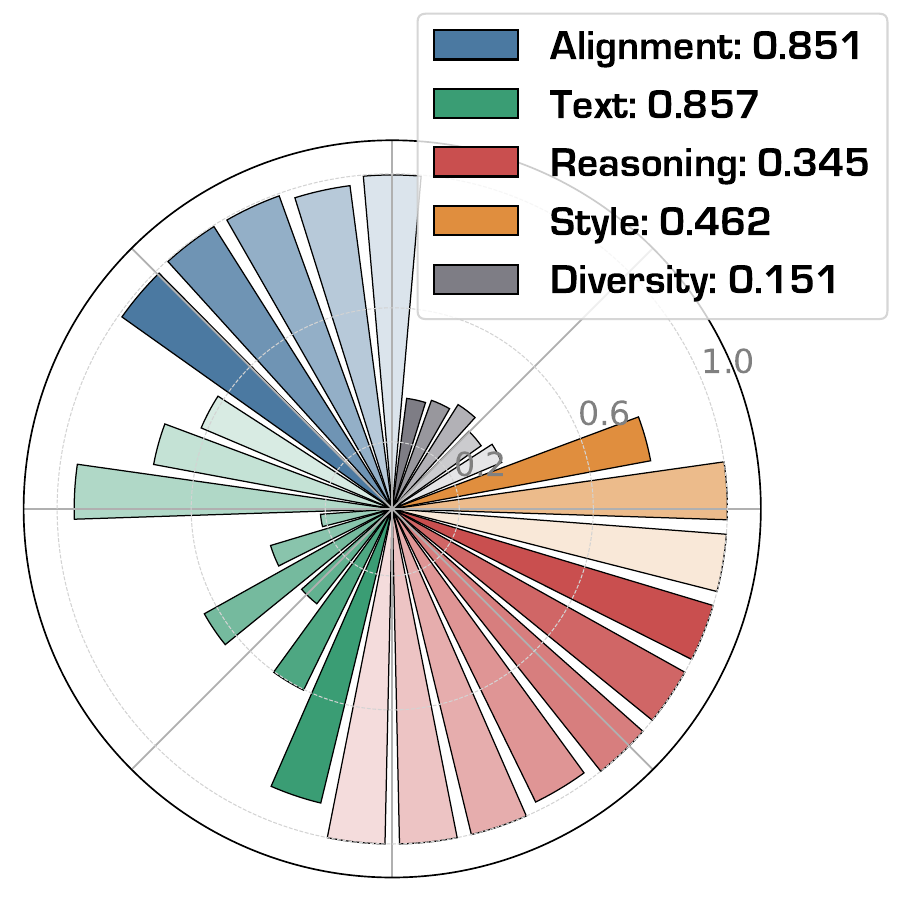} & & \\
    \bottomrule
    \end{tabular}}
    \vspace{-0.4cm}
\end{table*}

\begin{figure}[ht]
    \centering
    \setlength{\abovecaptionskip}{1pt}
    \includegraphics[width=.98\linewidth]{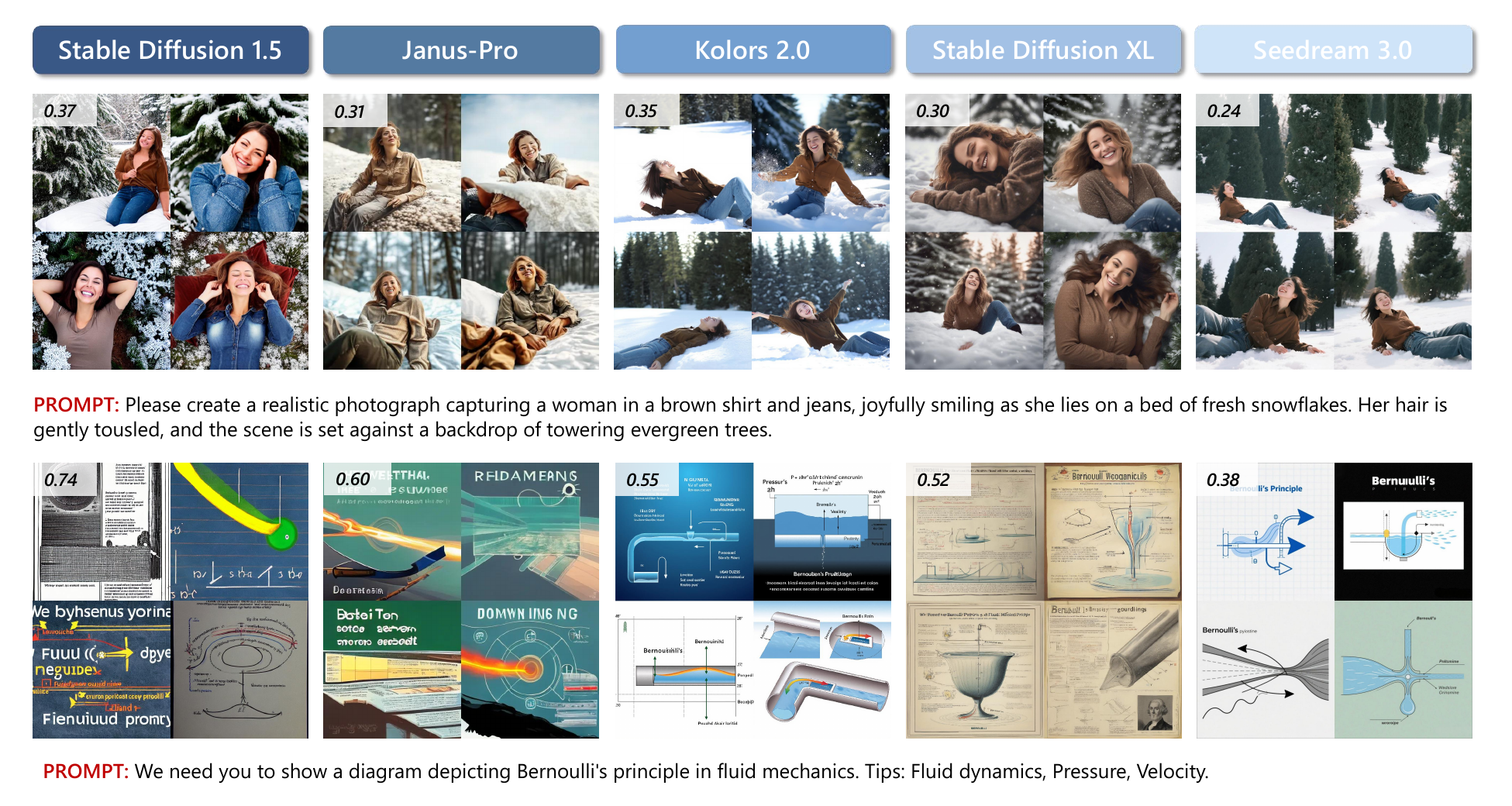}
    \caption{\small \textbf{Visualization results of SOTA methods.} \textbf{Diversity} of Stable Diffusion 1.5~\cite{Rombach_2022_CVPR}, Janus-Pro~\cite{chen2025janus}, Kolors 2.0~\cite{2025Kolors2}, Stable Diffusion XL~\cite{podell2023sdxl} and Seedream 3.0~\cite{gao2025seedream} are evaluated respectively.}
    \label{fig:vis_diversity}
    \vspace{-0.35cm}
\end{figure}

The visualization of diversity results is presented in Figure~\ref{fig:vis_diversity}, where the ranking of diversity scores aligns well with visual inspection, supporting the validity of our proposed diversity metrics, although the diversity observed in some methods may partly stem from insufficient alignment capabilities.

\begin{figure}[ht]
    \centering
    \setlength{\abovecaptionskip}{1pt}
    \includegraphics[width=.86\linewidth]{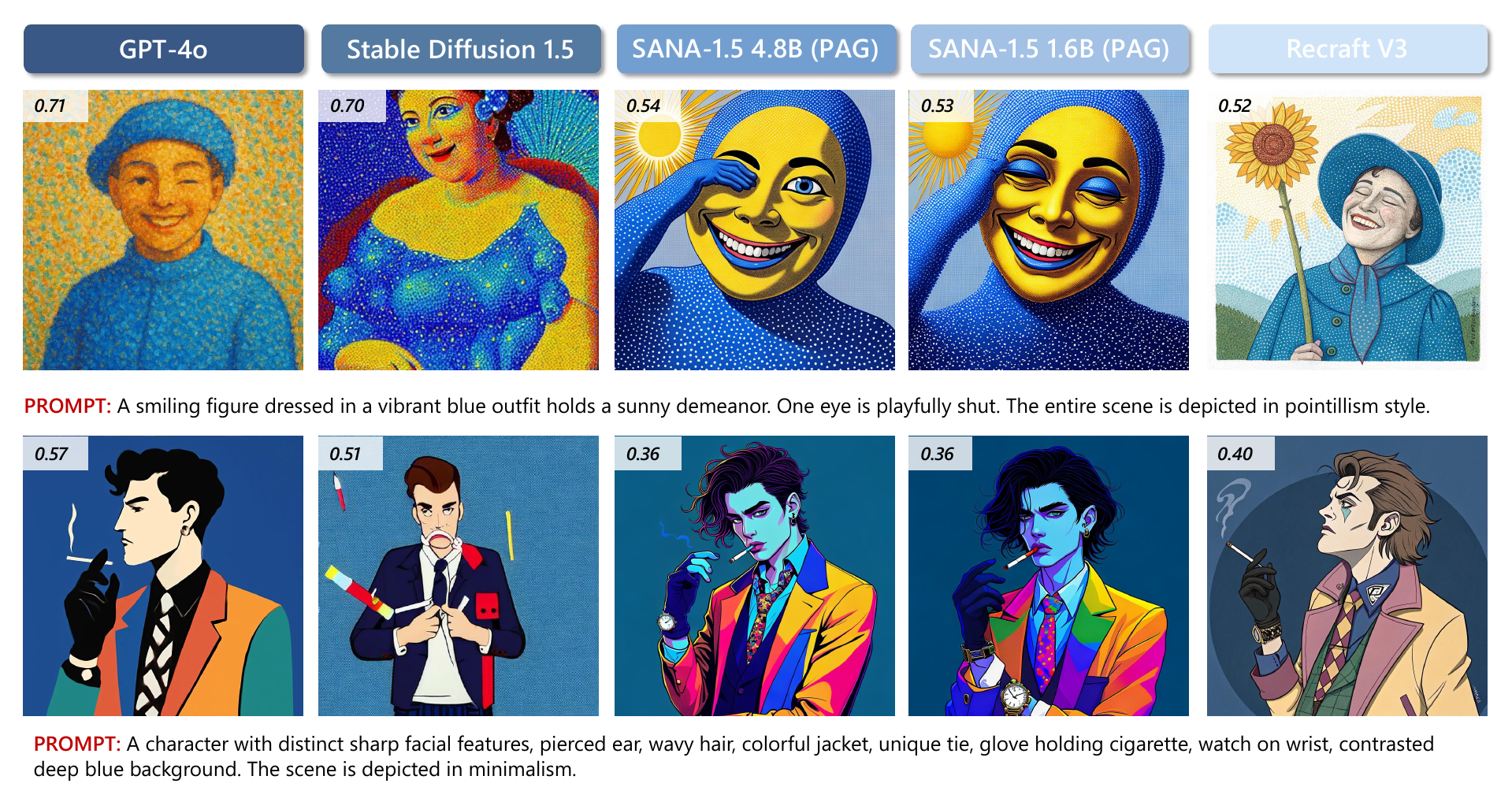}
    \caption{\small \textbf{Visualization results of SOTA methods.} \textbf{Traditional} styles of GPT-4o~\cite{openai2024gpt4o_image}, Stable Diffusion 1.5~\cite{Rombach_2022_CVPR}, SANA-1.5 4.8B (PAG) and 1.6B (PAG)~\cite{xie2025sana}, and Recraft V3~\cite{2024recraftv3} are evaluated respectively. The styles are  \textbf{pointillism} and  \textbf{minimalism}.}
    \label{fig:traditional_style}
    \vspace{-0.35cm}
\end{figure}

\begin{figure}[ht]
    \centering
    \setlength{\abovecaptionskip}{1pt}
    \includegraphics[width=.86\linewidth]{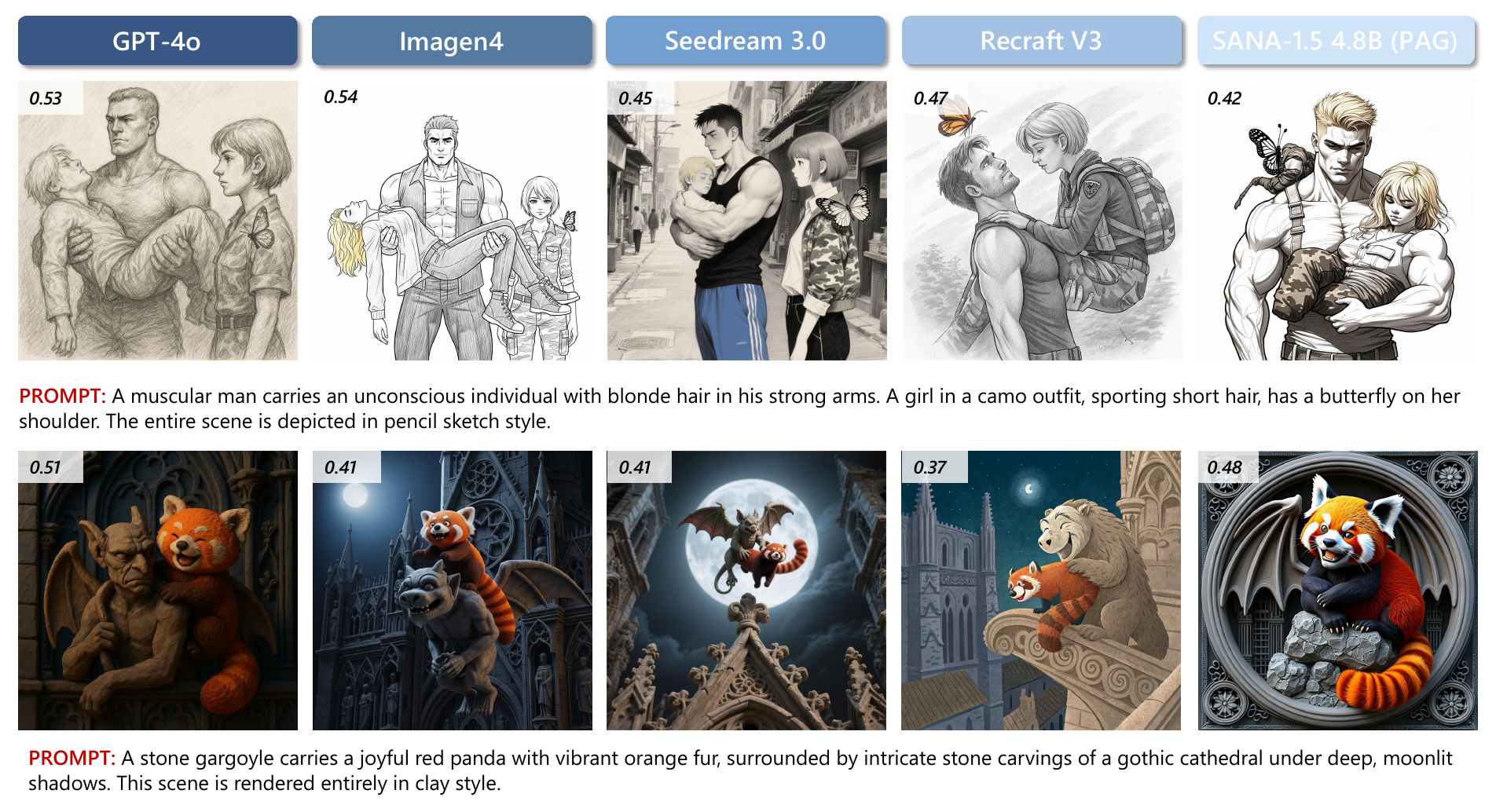}
    \caption{\small \textbf{Visualization results of SOTA methods.} \textbf{Media} styles of GPT-4o~\cite{openai2024gpt4o_image}, Imagen4~\cite{2025Imagen4}, Seedream 3.0~\cite{gao2025seedream}, Recraft V3\cite{2024recraftv3} and SANA-1.5 4.8B (PAG)~\cite{xie2025sana} are evaluated respectively. The styles are  \textbf{pencil sketch} and  \textbf{stone sculpture}.}
    \label{fig:media_style}
    \vspace{-0.35cm}
\end{figure}

\begin{figure}[ht]
    \centering
    \setlength{\abovecaptionskip}{1pt}
    \includegraphics[width=.86\linewidth]{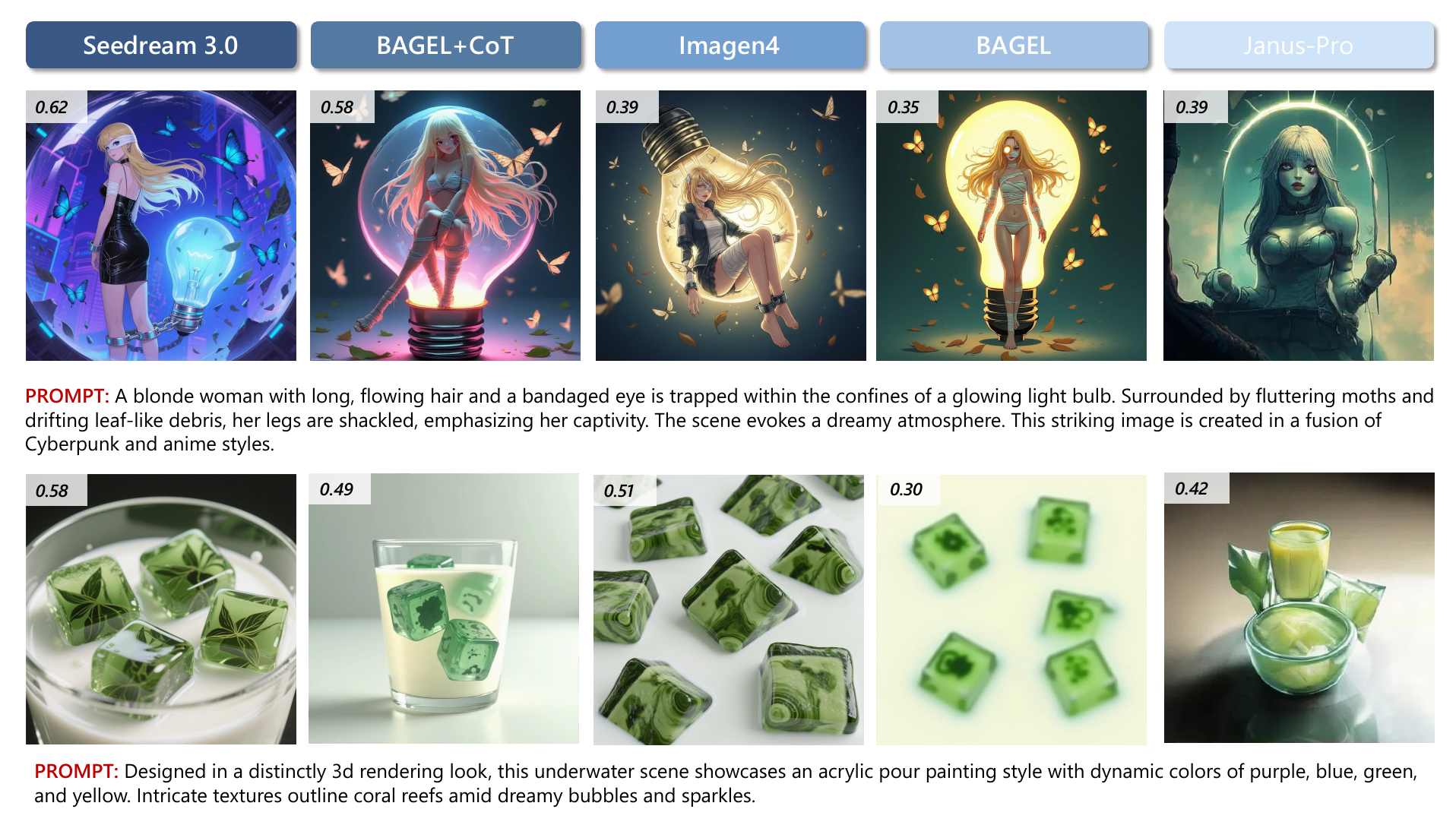}
\caption{\small \textbf{Visualization results of SOTA methods.} \textbf{Anime} styles of Seedream 3.0~\cite{gao2025seedream}, BAGEL+CoT~\cite{deng2025bagel}, Imagen4~\cite{2025Imagen4}, BAGEL~\cite{deng2025bagel} and Janus-Pro~\cite{chen2025janus} are evaluated respectively The styles are \textbf{cyberpunk} and \textbf{3d rendering}.}
    \label{fig:anime_style}
    \vspace{-0.35cm}
\end{figure}

Figures~\ref{fig:traditional_style},~\ref{fig:media_style}, and~\ref{fig:anime_style} show that GPT-4o~\cite{openai2024gpt4o_image} performs well across most styles, though it struggles with specific ones especially some anime styles. Stable Diffusion 1.5~\cite{Rombach_2022_CVPR}, despite its lower visual quality, effectively captures traditional style features. Although Imagen4~\cite{2025Imagen4} does not achieve a high overall score in style, it performs notably well in media and anime styles. It is worth noting that unified multimodal methods, such as BAGEL+CoT~\cite{deng2025bagel}, OmniGen2~\cite{wu2025omnigen2}, BAGEL~\cite{deng2025bagel} and Janus-Pro~\cite{chen2025janus}, exhibit particularly strong performance in the anime style. Overall,  Seedream 3.0\cite{gao2025seedream} and SANA-1.5 4.8B (PAG)\cite{xie2025sana}  also demonstrate strong stylistic consistency, ranking just behind GPT-4o. 

\newpage


\end{document}